\documentclass[5p]{elsarticle}
\usepackage{times}
\usepackage{epsfig}
\usepackage{graphicx}
\usepackage{amsmath}
\usepackage{amssymb}
\usepackage[space]{grffile}
\usepackage{lineno,hyperref}
\usepackage{color}

\journal{Computer Vision and Image Understanding}

\begin{document}

\begin{frontmatter}

\title{Occlusion Coherence: Detecting and Localizing Occluded Faces}

\author{Golnaz Ghiasi} 
\author{Charless C. Fowlkes}
\address{University of California at Irvine, Irvine, CA 92697}

\begin{abstract}
The presence of occluders significantly impacts object recognition accuracy.
However, occlusion is typically treated as an unstructured source of noise and
explicit models for occluders have lagged behind those for object appearance
and shape.  In this paper we describe a hierarchical deformable part model for
face detection and landmark localization that explicitly models part occlusion.
The proposed model structure makes it possible to augment positive training
data with large numbers of synthetically occluded instances.  This allows us to
easily incorporate the statistics of occlusion patterns in a discriminatively
trained model.  We test the model on several benchmarks for landmark 
localization and detection including challenging new data sets featuring
significant occlusion. We find that the addition of an explicit occlusion model
yields a detection system that outperforms existing approaches for occluded
instances while maintaining competitive accuracy in detection and landmark 
localization for unoccluded instances.

\end{abstract}


\end{frontmatter}


\section{Introduction}

\begin{figure}[t!]
	\begin{tabular}{c@{\hskip 0pt}c@{\hskip 0pt}c}
		\includegraphics[width=.3\columnwidth]{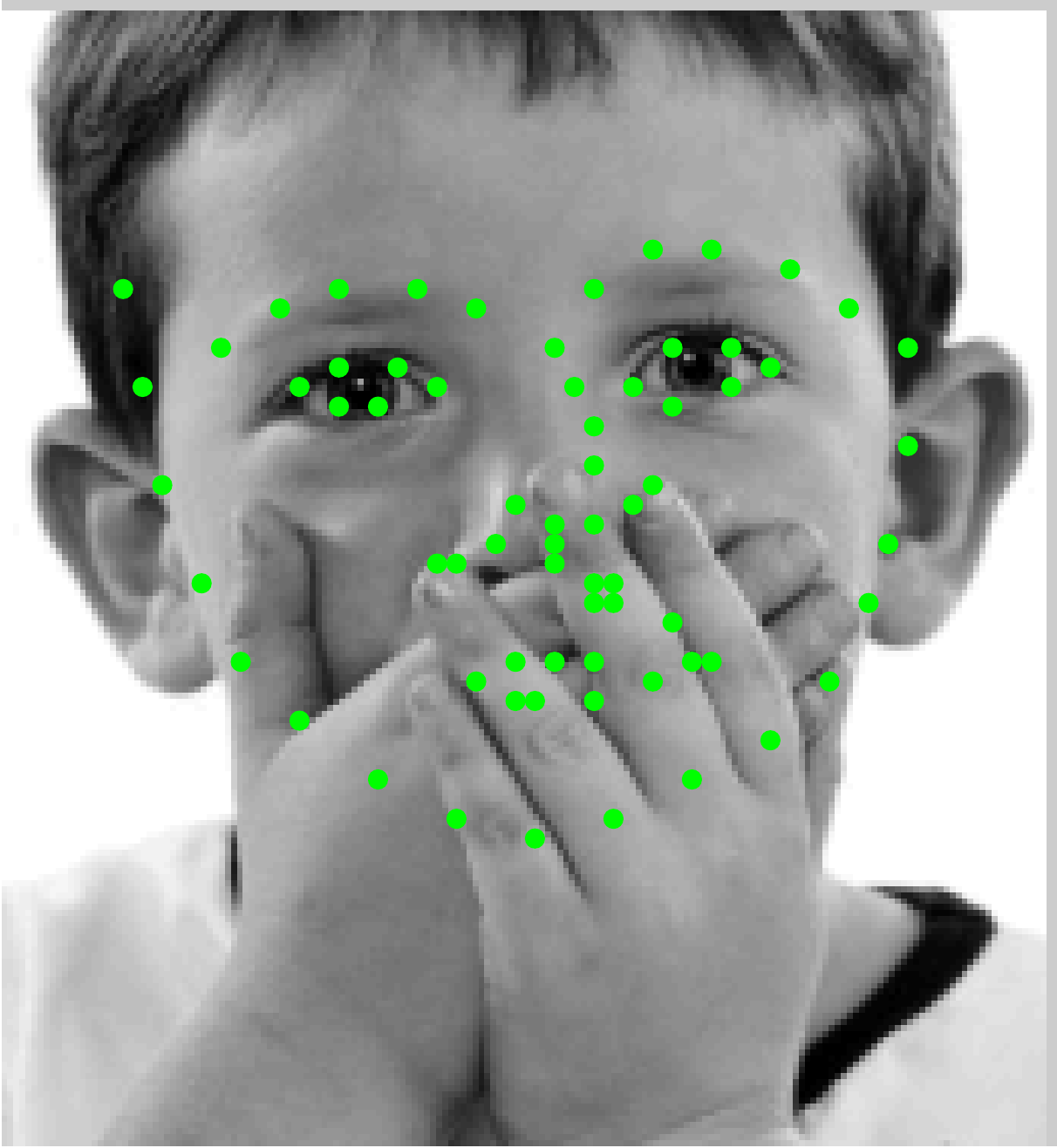}&
		\includegraphics[width=.3\columnwidth]{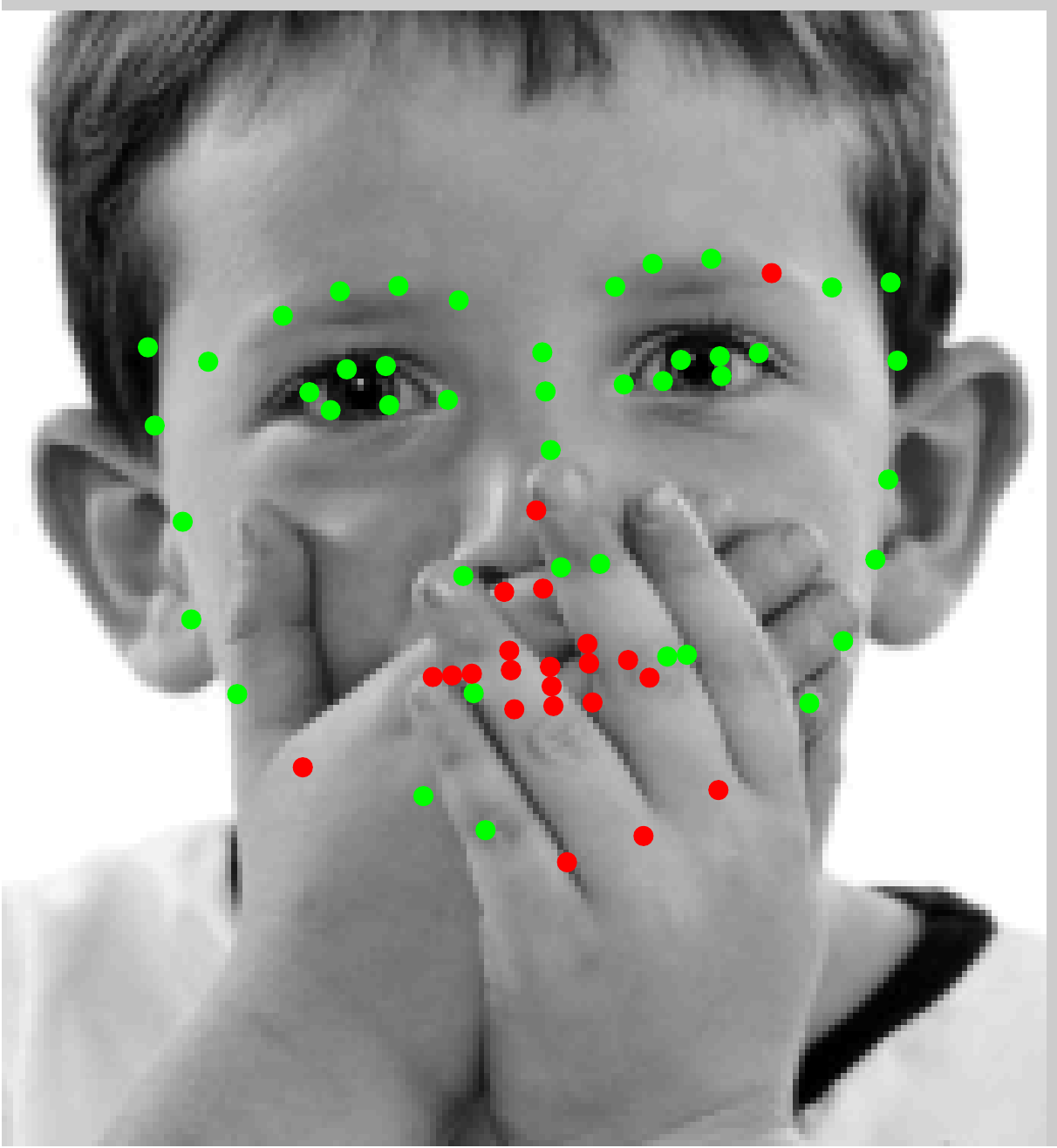}&		
		\includegraphics[width=.3\columnwidth]{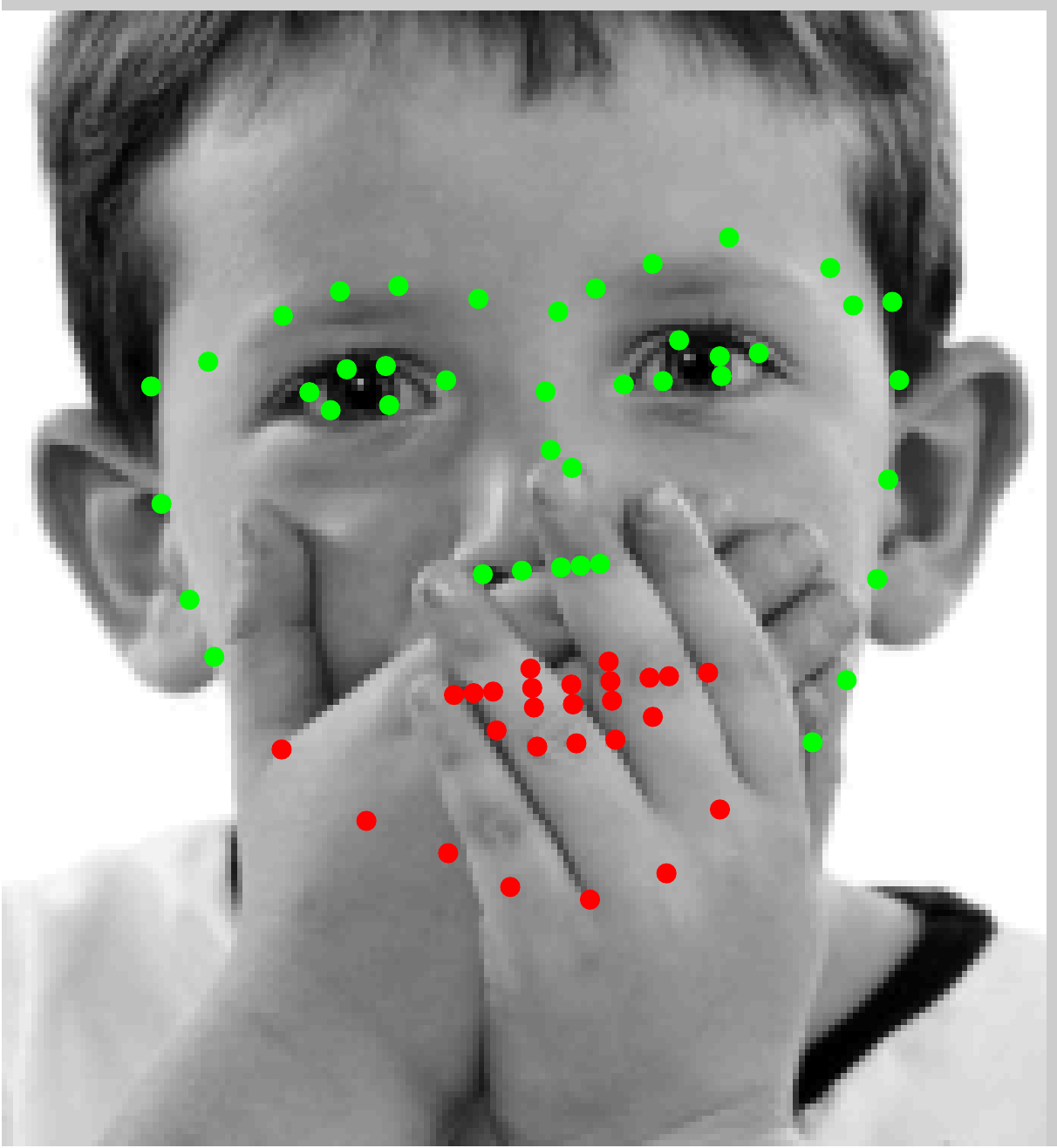}\\
		\includegraphics[width=.3\columnwidth]{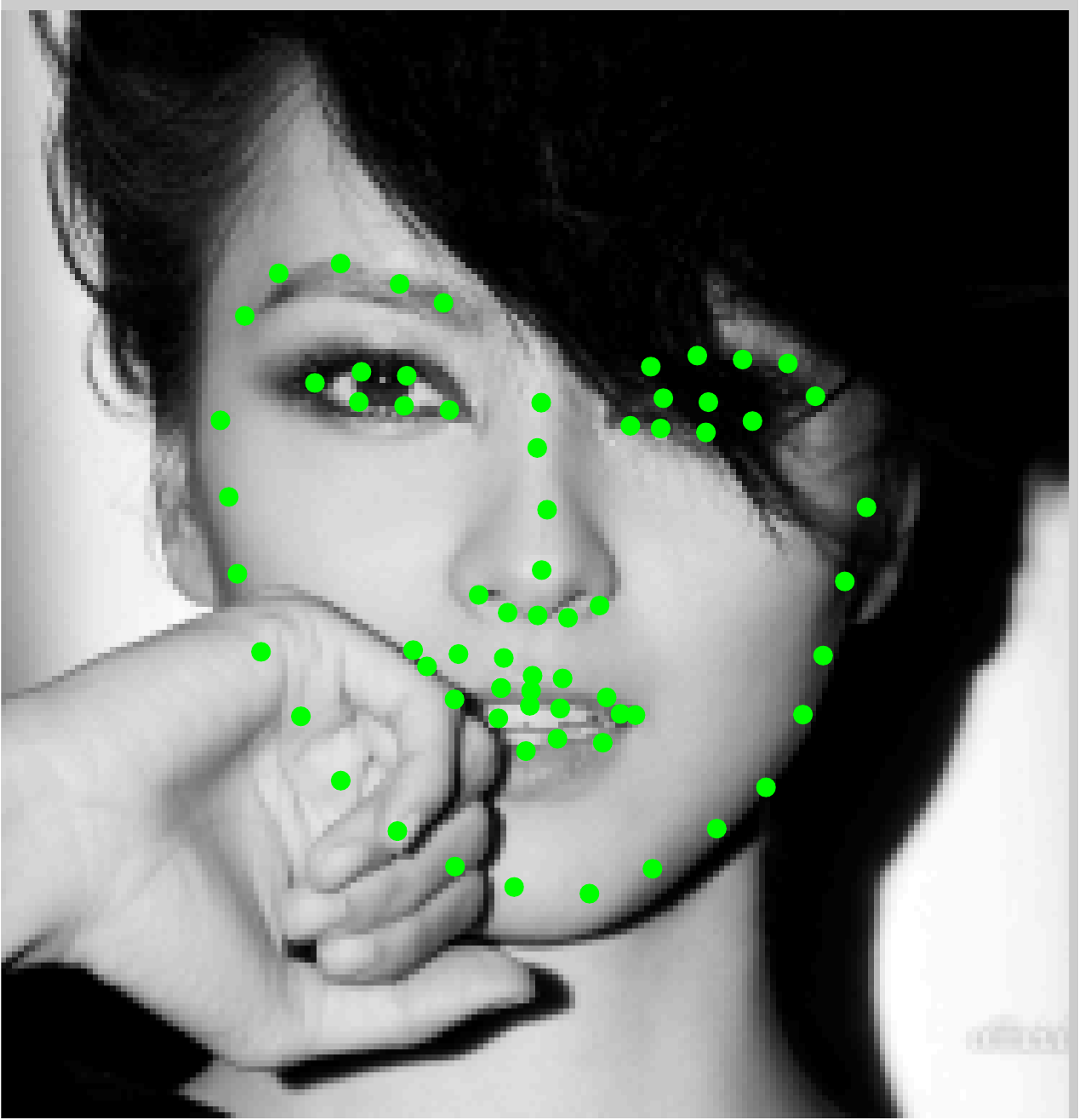}&
		\includegraphics[width=.3\columnwidth]{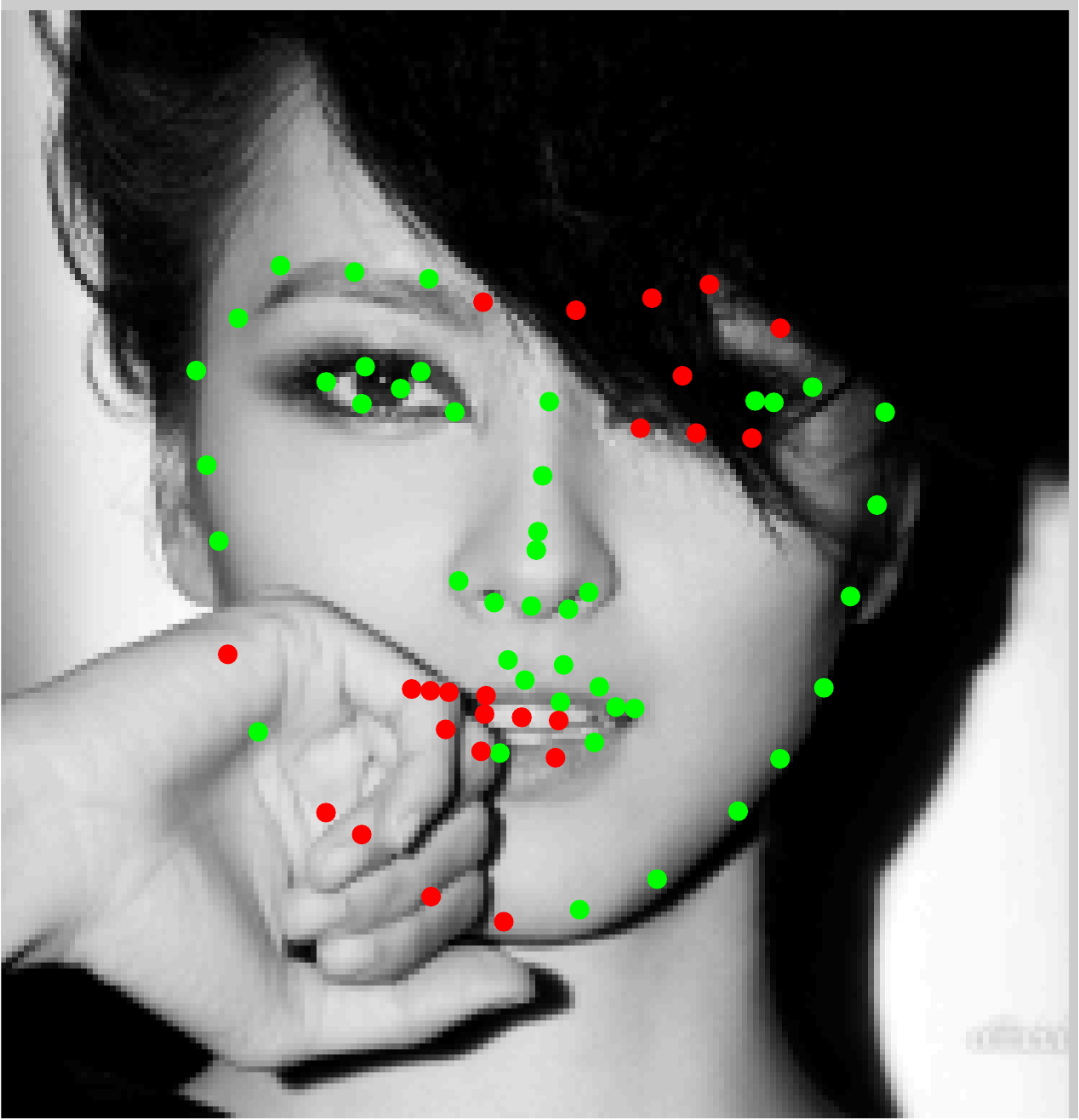}&
		\includegraphics[width=.3\columnwidth]{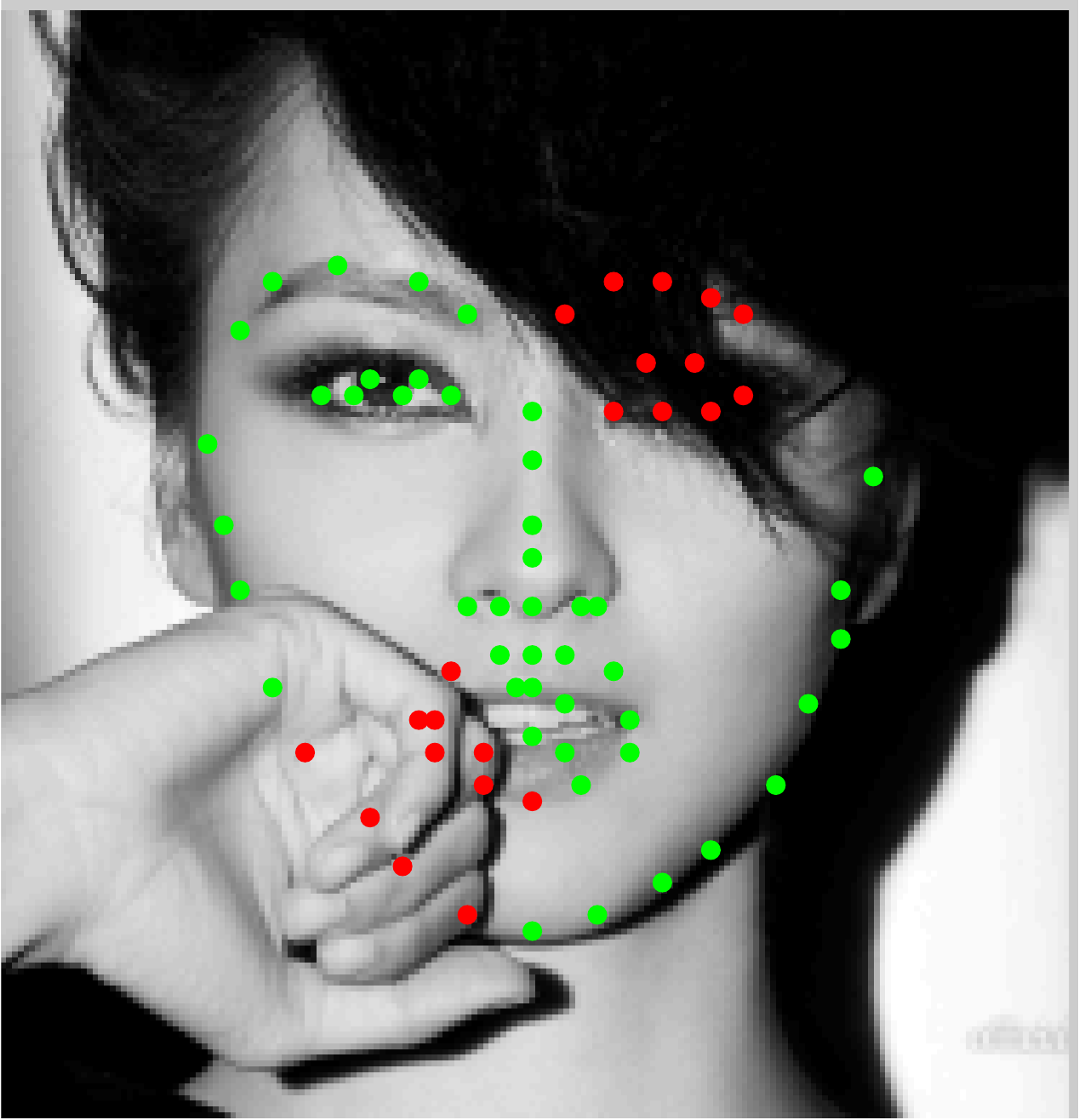}\\	
		\includegraphics[width=.3\columnwidth]{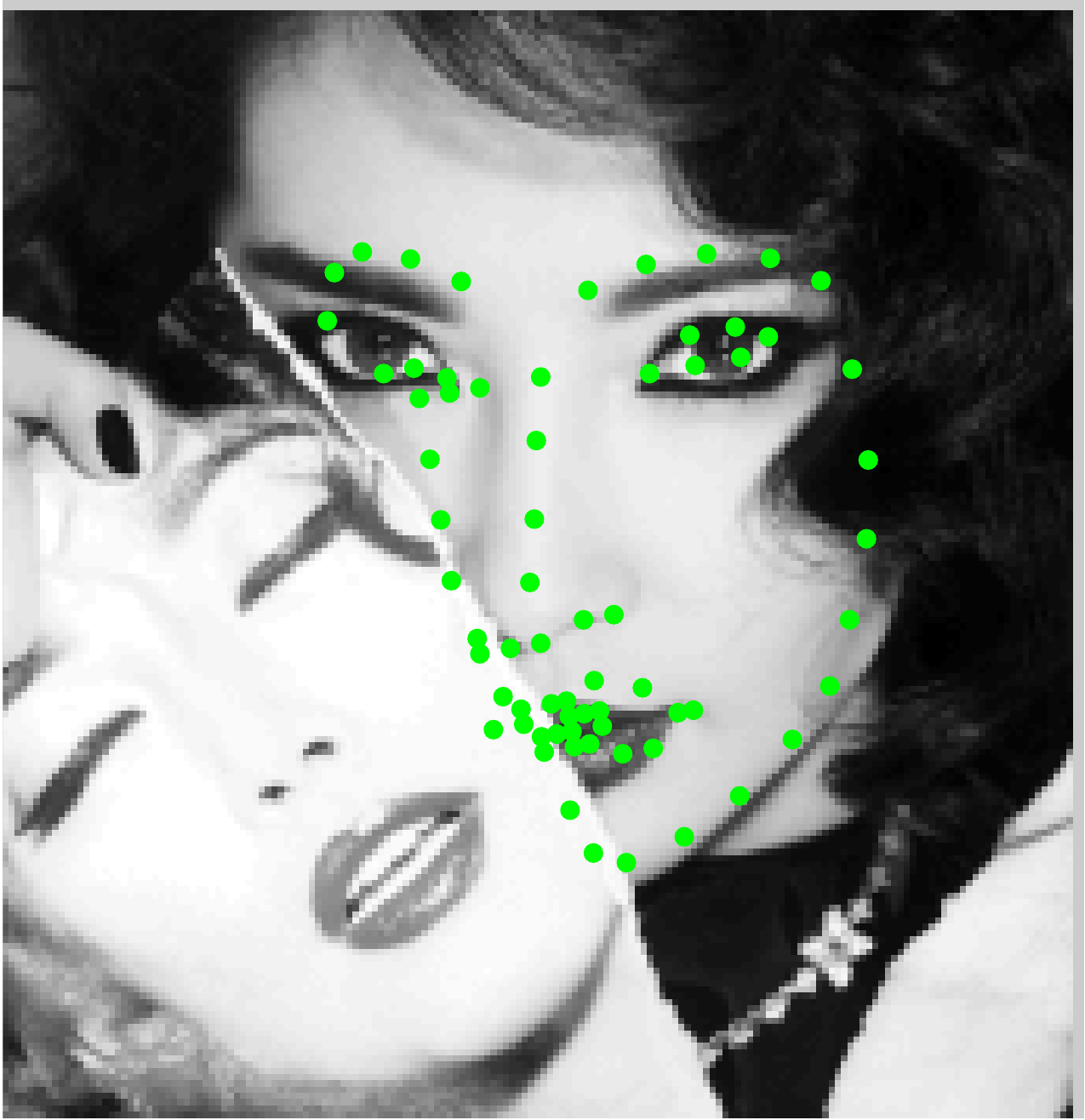}&
		\includegraphics[width=.3\columnwidth]{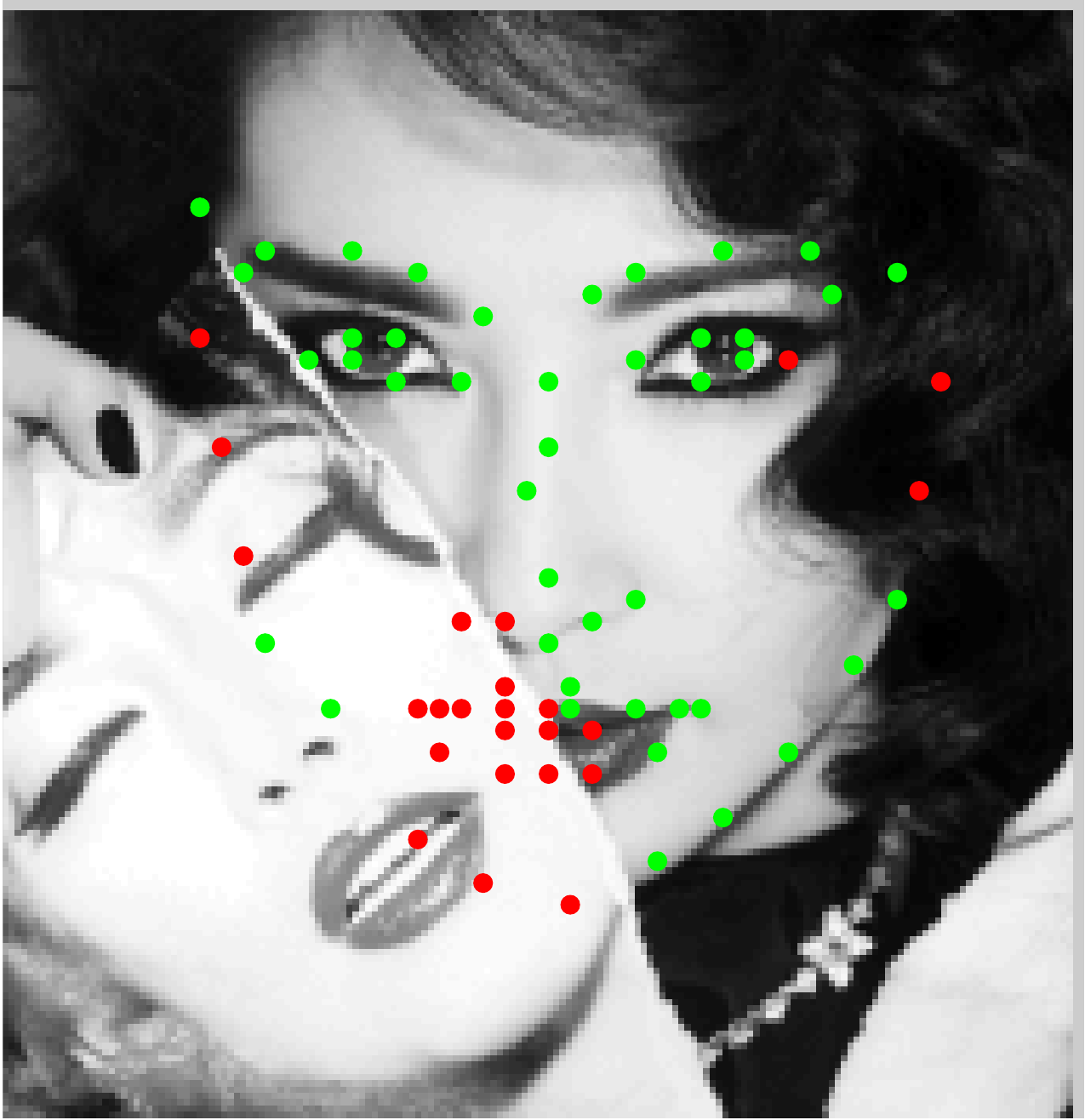}&
		\includegraphics[width=.3\columnwidth]{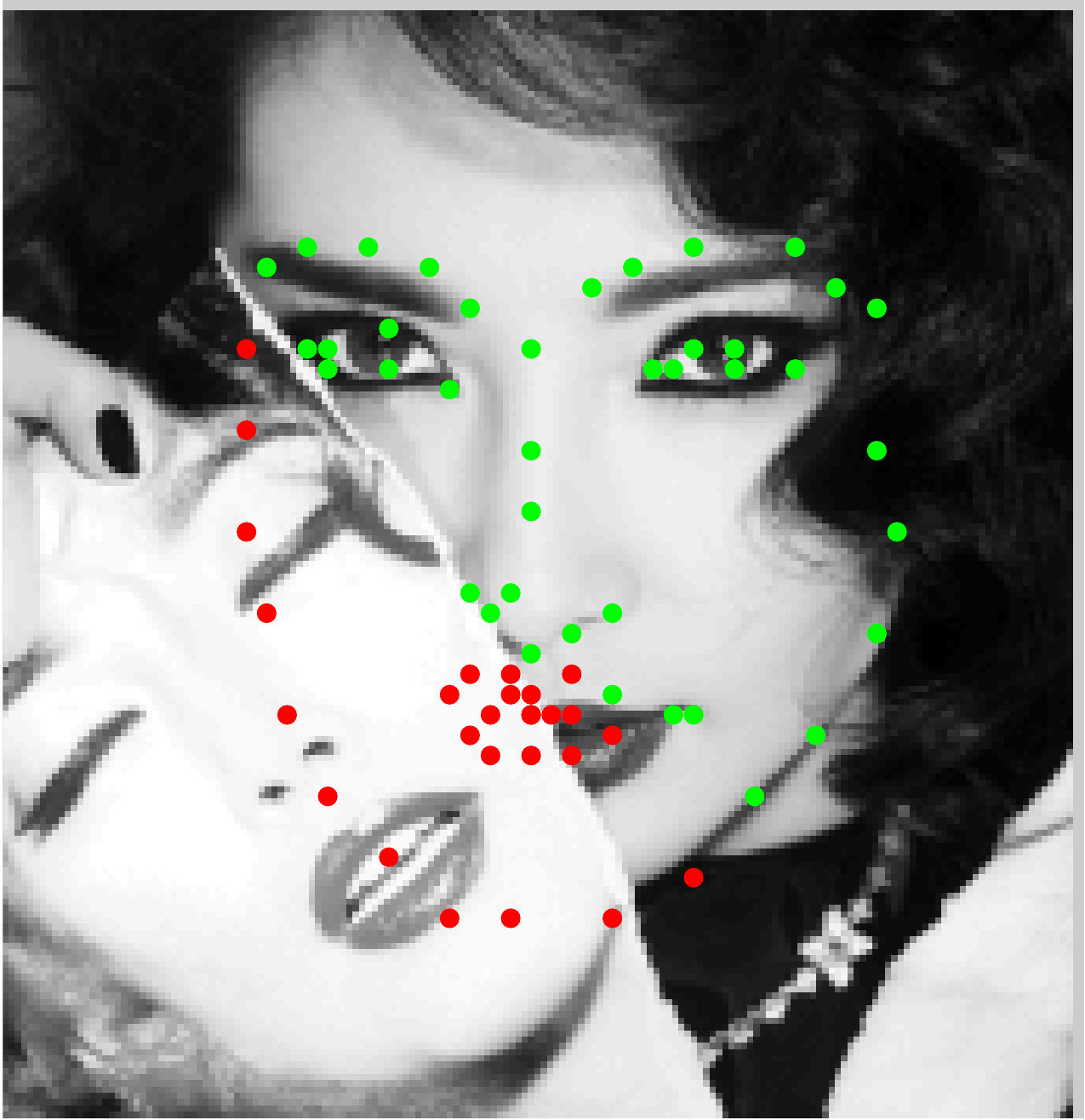}\\	
  	        (a)&(b)&(c)\\
	\end{tabular}
\caption{Occlusion impacts part localization performance.  In panel (a) the output
of a deformable part model~\cite{zhu2012face} is distorted by the presence of
occluders, disrupting localization even for parts that are far from the site of
occlusion. (b) Introducing independent occlusion of each part results in better
alignment but occlusion is treated as an outlier process and prediction of
occlusion state is inaccurate.  (c) The output of our hierarchical part model,
which explicitly models likely patterns of occlusion, shows improved
localization as well as accurate prediction of which landmarks are occluded.
}
\label{fig:splash}
\end{figure}

Accurate localization of facial landmarks provides an
important building block for many applications including
identification~\cite{blanz2003face} and analysis of facial
expressions~\cite{martinez2012model}.  Significant progress has been made in
this task, aided in part by the fact that faces have less intra-category shape
variation and limited articulation compared to other object categories of
interest.  However, feature point localization tends to break down when applied
to faces in real scenes where other objects in the scene (hair, sunglasses,
other people) are likely to occlude parts of the face.
Fig.~\ref{fig:splash}(a) depicts the output of a deformable part 
model~\cite{zhu2012face} where the presence of occluders distorts the final
alignment of the model.

A standard approach to handling occlusion in part-based models is to compete
part feature scores against a generic background model or fixed threshold
(as in Fig. ~\ref{fig:splash}(b)).  However, setting such thresholds is fraught with
difficulty since it is hard to distinguish between parts that are present
but simply hard to detect (e.g., due to unusual lighting) and those which are
genuinely hidden behind another object.  

Treating occlusions as an unstructured source of noise ignores a key aspect of
the problem, namely that occlusions are induced by other objects and surfaces
in the scene and hence should exhibit {\bf occlusion coherence}.  For example,
it would seem very unlikely that every-other landmark along an object
contour would happen to be occluded. Yet many occlusion models make strong
independence assumptions about occlusion, making it difficult to distinguish
{\em a priori} likely from unlikely patterns.  Ultimately, an occluder should
not be inferred simply by the lack of evidence for object features, but rather
by positive evidence for the occluding object that {\bf explains away} the lack
of object features.

The contribution of this paper is an efficient hierarchical deformable part model
that encodes these principles for modeling occlusion and achieves
state-of-the-art performance on benchmarks for occluded face localization and
detection (depicted in Fig.~\ref{fig:splash}(c)).  Building on our previously published
results~\cite{GhiasiFowlkesCVPR2014}, we model the face by an arrangement of
parts, each of which is in turn composed of local landmark features.  This
two-layer model provides a compact, discriminative representation for the
appearance and deformations of parts. It also captures the correlation in
shapes and occlusion patterns of neighboring parts (e.g., if the chin is
occluded it would seem more likely the bottom half of the mouth is also
occluded). In addition to representing the face shape, each part has an
associated occlusion state chosen from a small set of possible occlusion
patterns, enforcing coherence across neighboring landmarks and providing a
sparse representation of the occluder shape where it intersects the part.
We describe the details of this model in Section \ref{sec:model}.

Specifying training data from which to learn feasible occlusion patterns comes
with an additional set of difficulties.  Practically speaking, existing
datasets have focused primarily on fully visible faces.  Moreover, it seems
unlikely that any reasonable sized set of training images would serve to
densely probe the space of possible occlusions.  Beyond certain weak contextual
constraints, the location and identity of the occluder itself are arbitrary and
largely independent of the occluded object.  To overcome this difficulty of
training data, we propose a unique approach for generating synthetically
occluded positive training examples. By exploiting the structural assumptions
built into our model, we are able to include such examples as ``virtual
training data'' without explicitly synthesizing new images. This in turn leads
to an interesting formulation of discriminative training using a loss function
that depends on the latent occlusion state of the parts for negative training
examples which we describe in Section \ref{sec:learning}.

We carry out an extensive analysis of this model performance in terms of
landmark localization, occlusion prediction and detection accuracy.  While our
model is trained as a detector, the internal structure of the model allows it
to perform high-quality landmark localization, comparable in accuracy to pose
regression, while being more robust to initialization and occlusions (Section
\ref{sec:landmarklocalization}). To carry out an empirical comparison to 
recently published models, we provide a new set of 68-landmark annotations for
the Caltech Occluded Faces in the Wild (COFW) benchmark dataset.  We find that
not only the localization but also the prediction of which landmarks are
occluded is improved over simple independent occlusion models (Section
\ref{sec:occlusionprediction}). Unlike landmark regression methods, our model
does not require initialization and achieves good performance on standard face
detection benchmarks such as FDDB \cite{fddbTech}. Finally, to illustrate the
impact of occlusion on existing detection models, we evaluate performance on a
new face detection dataset that contains significant numbers of partially
occluded faces (Section \ref{sec:facedetection}).

\begin{figure*}[t!]
\begin{tabular}{ccccccc}
\includegraphics[width=.3\columnwidth]{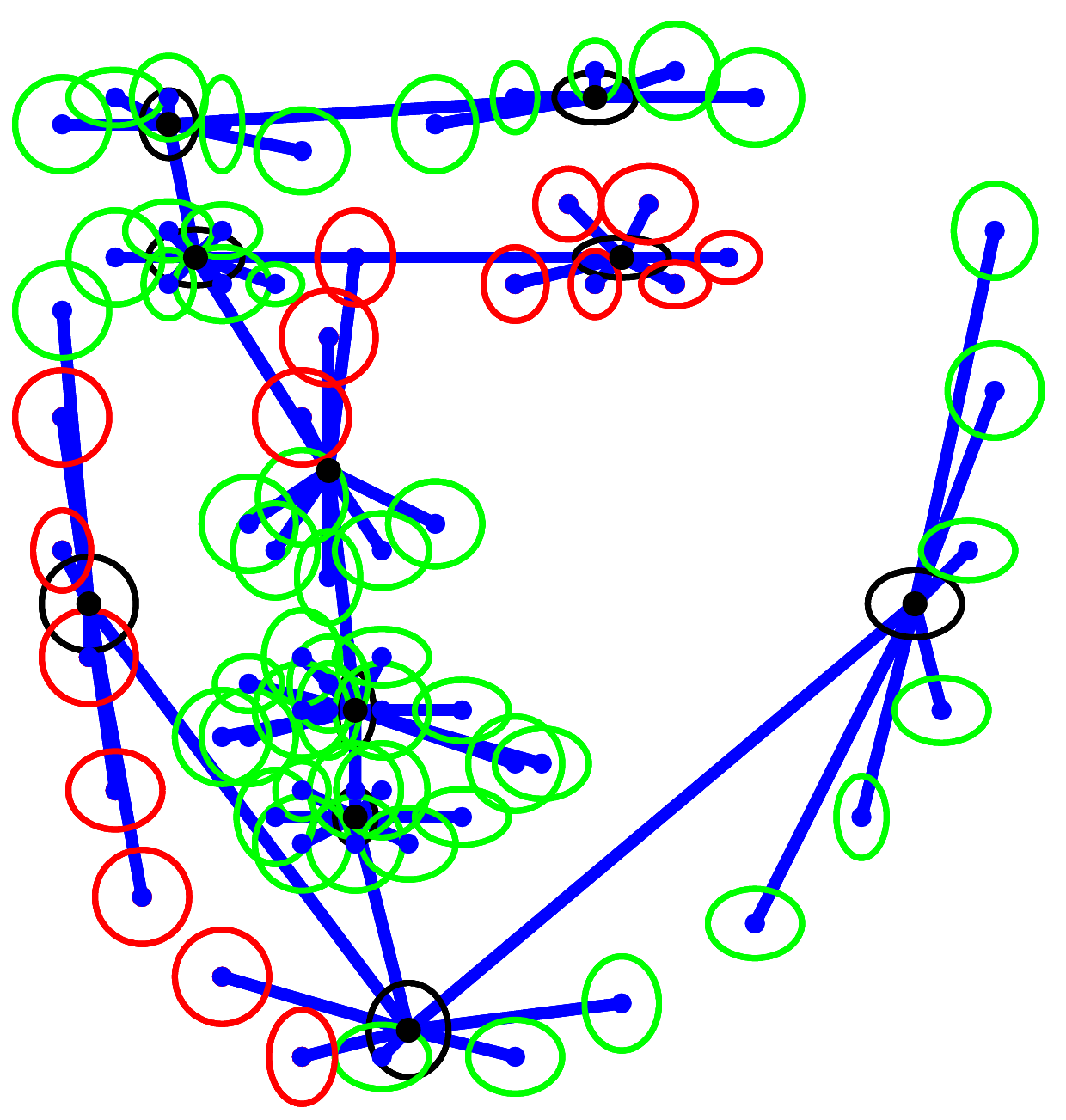}&
\includegraphics[width=.3\columnwidth]{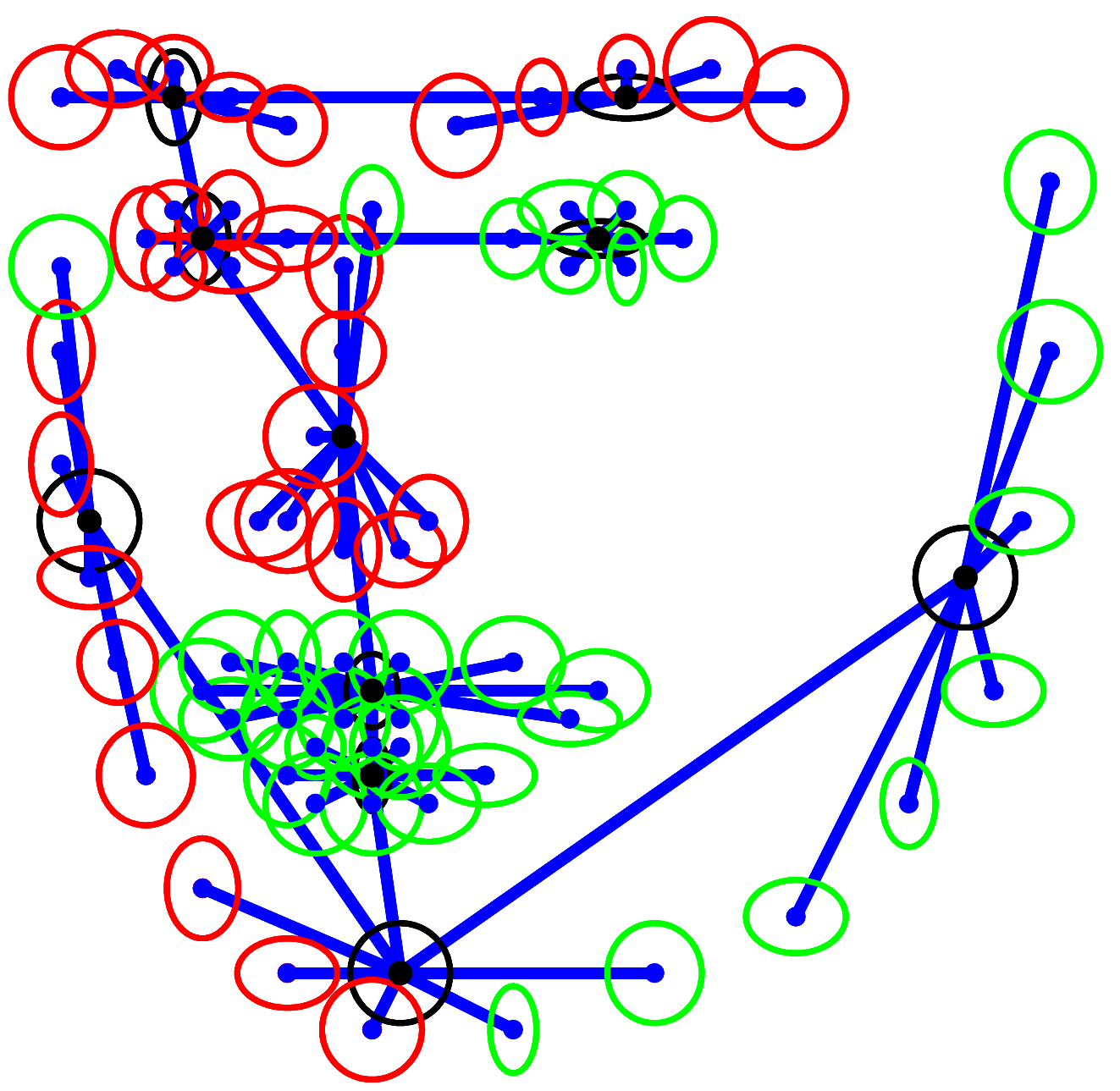}&
\includegraphics[width=.3\columnwidth]{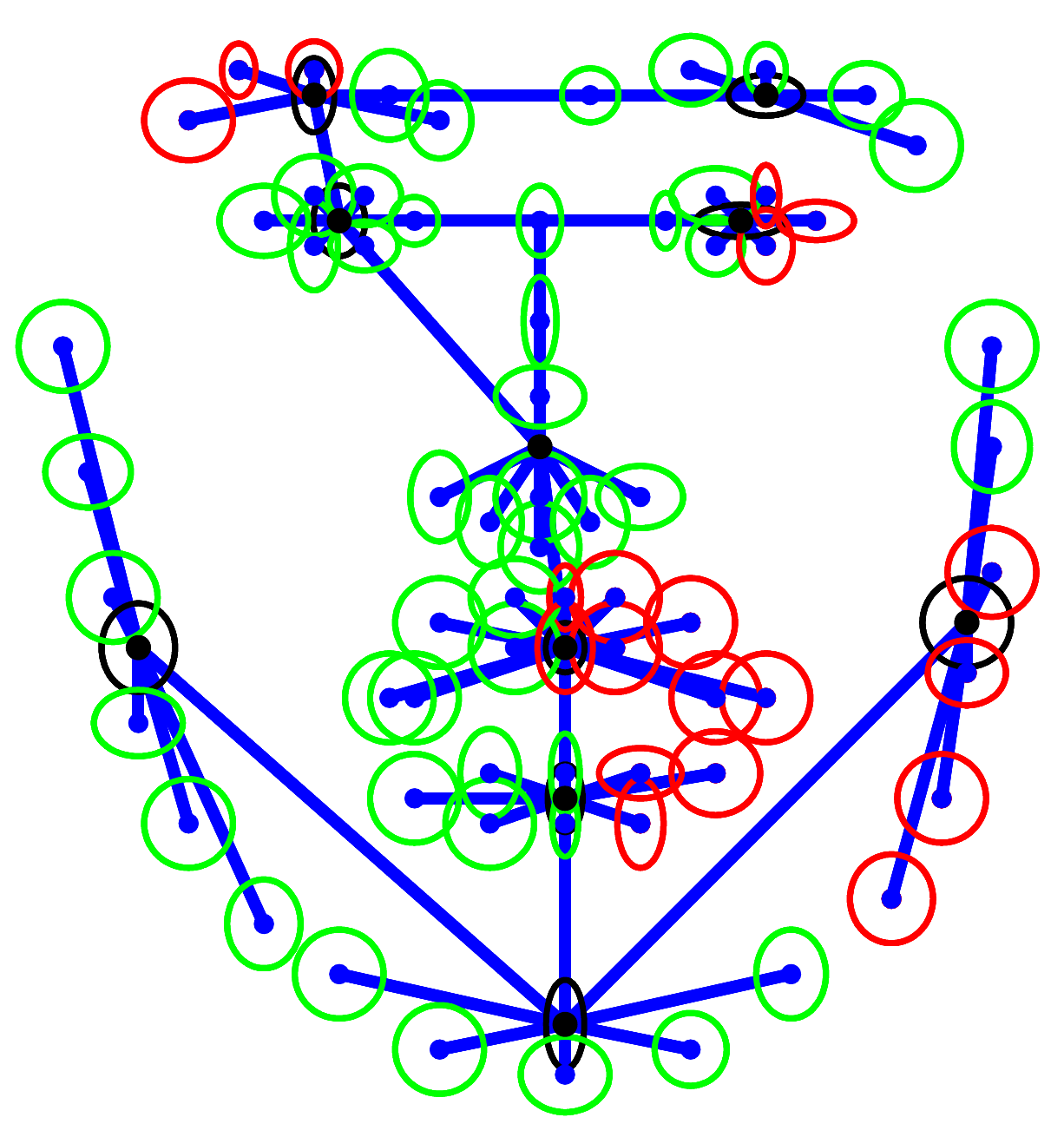}&
\includegraphics[width=.3\columnwidth]{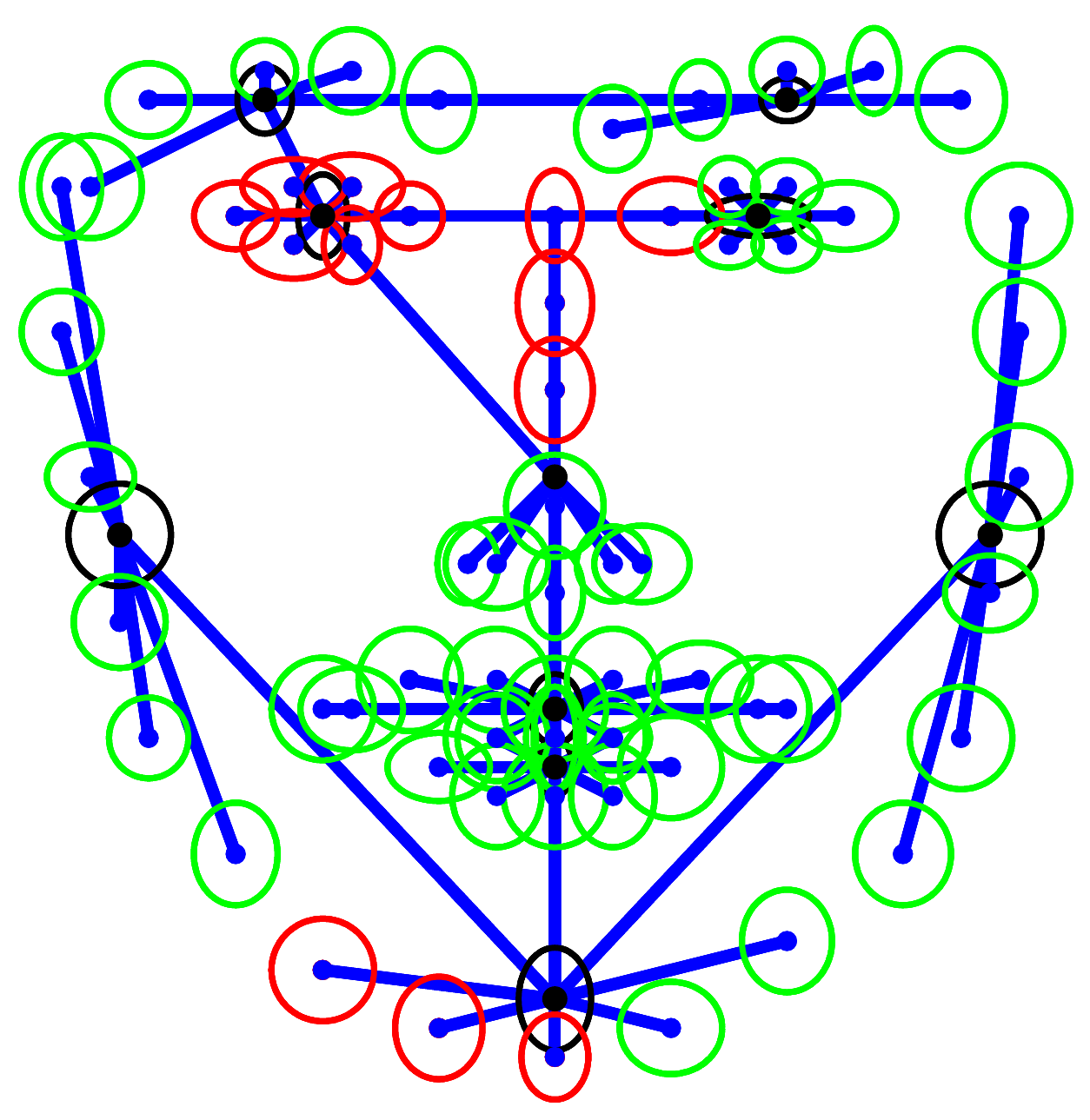}&
\includegraphics[width=.3\columnwidth]{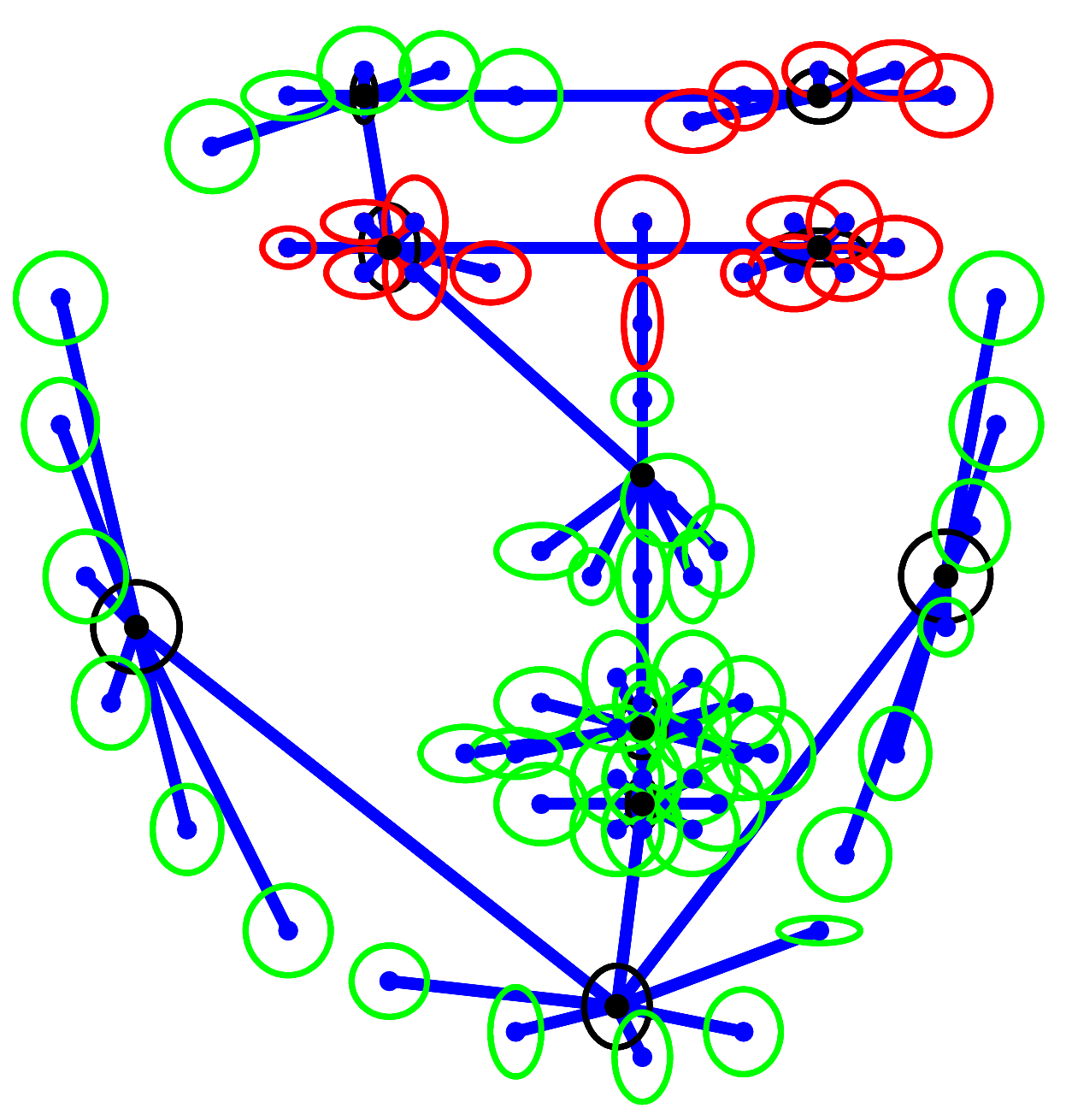}&
\includegraphics[width=.3\columnwidth]{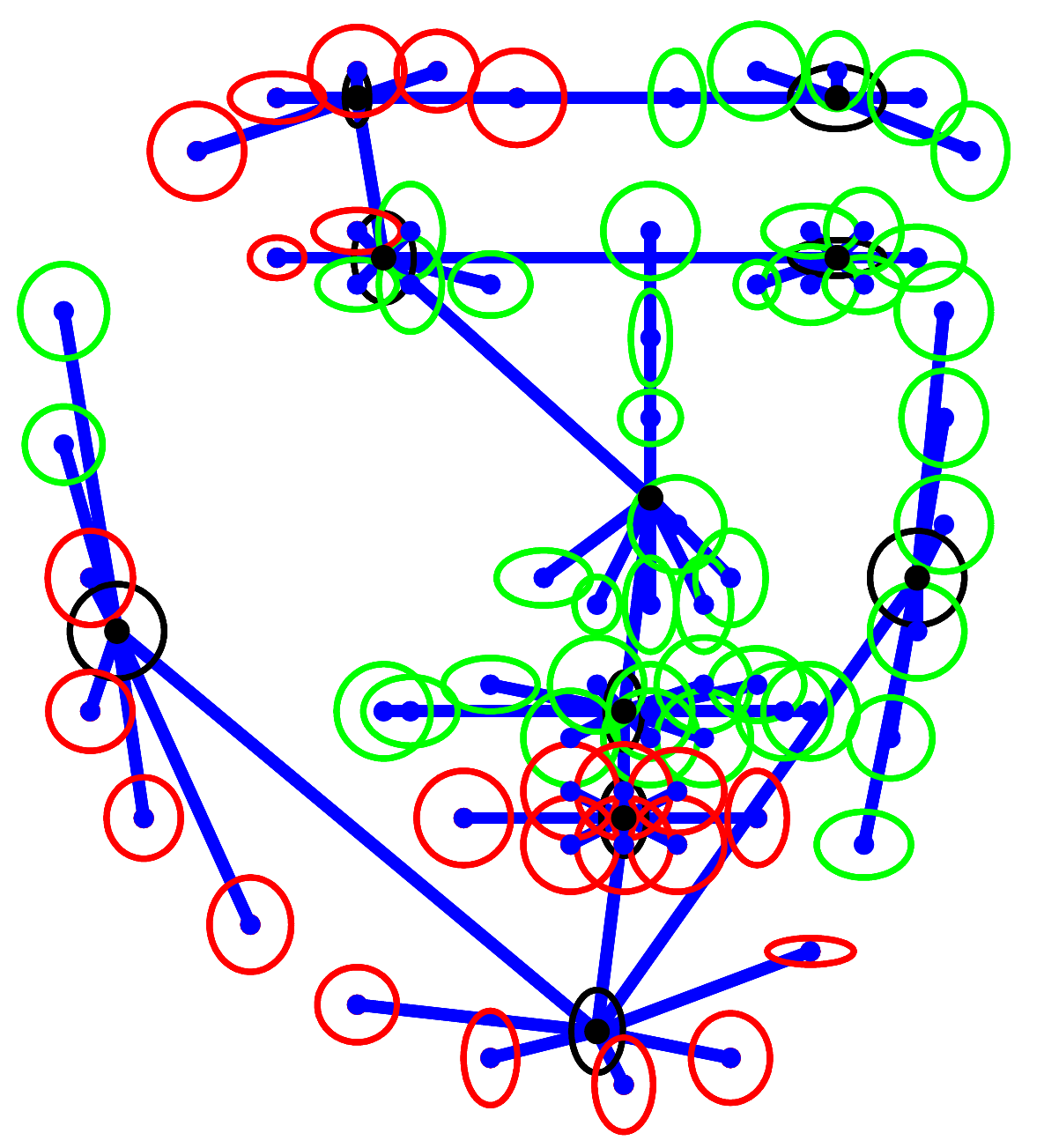}\\
\includegraphics[width=.3\columnwidth]{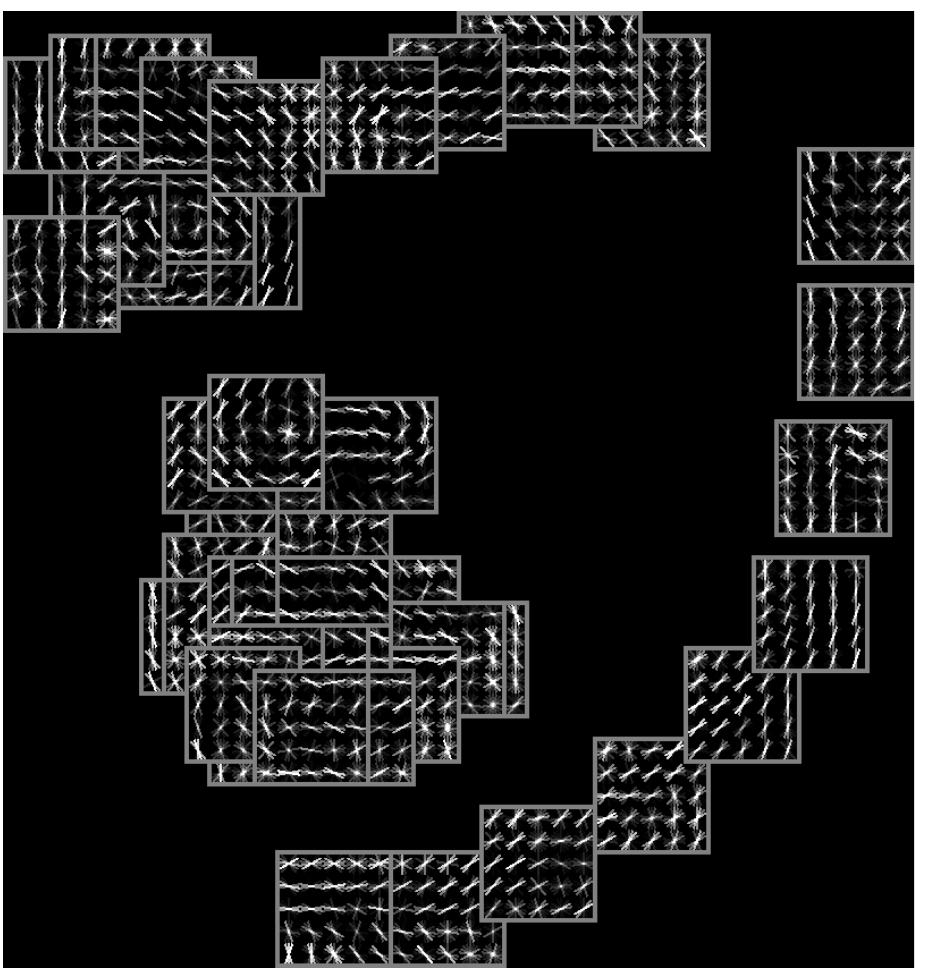}&
\includegraphics[width=.3\columnwidth]{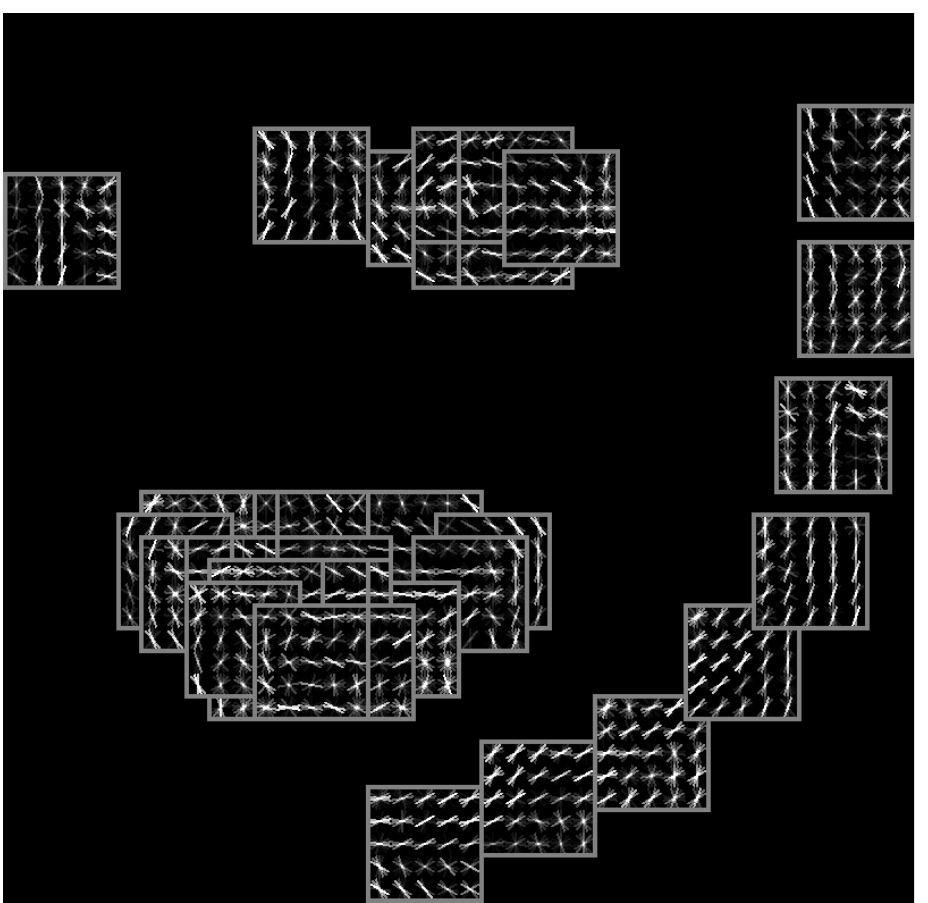}&
\includegraphics[width=.3\columnwidth]{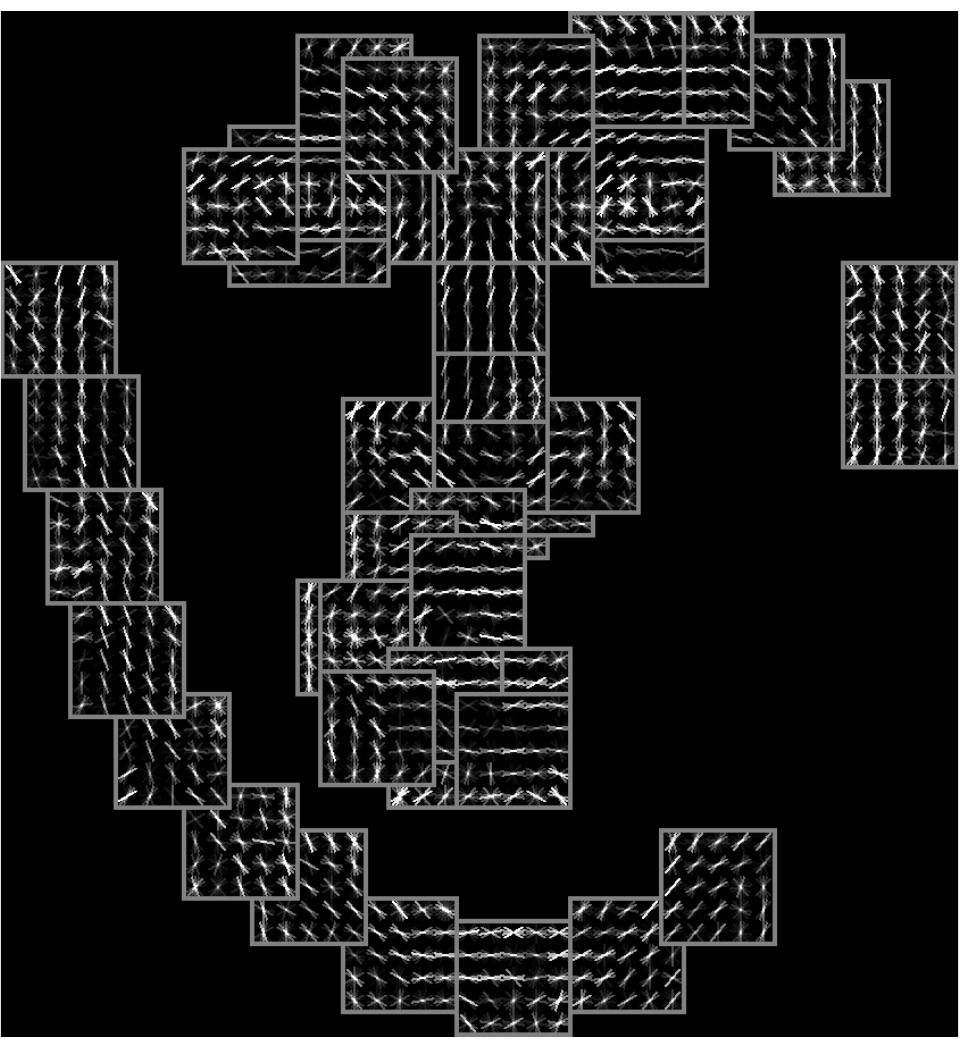}&
\includegraphics[width=.3\columnwidth]{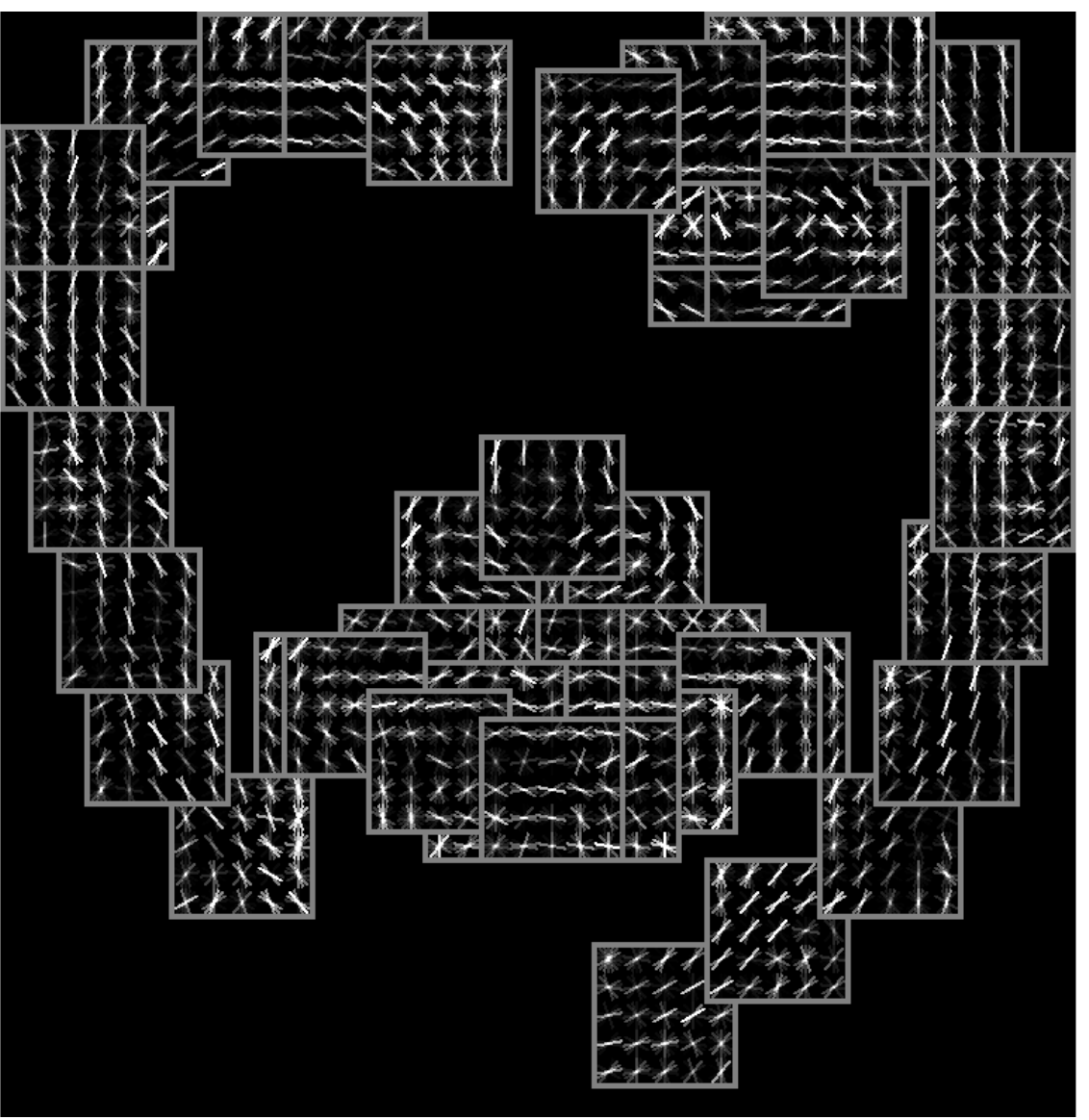}&
\includegraphics[width=.3\columnwidth]{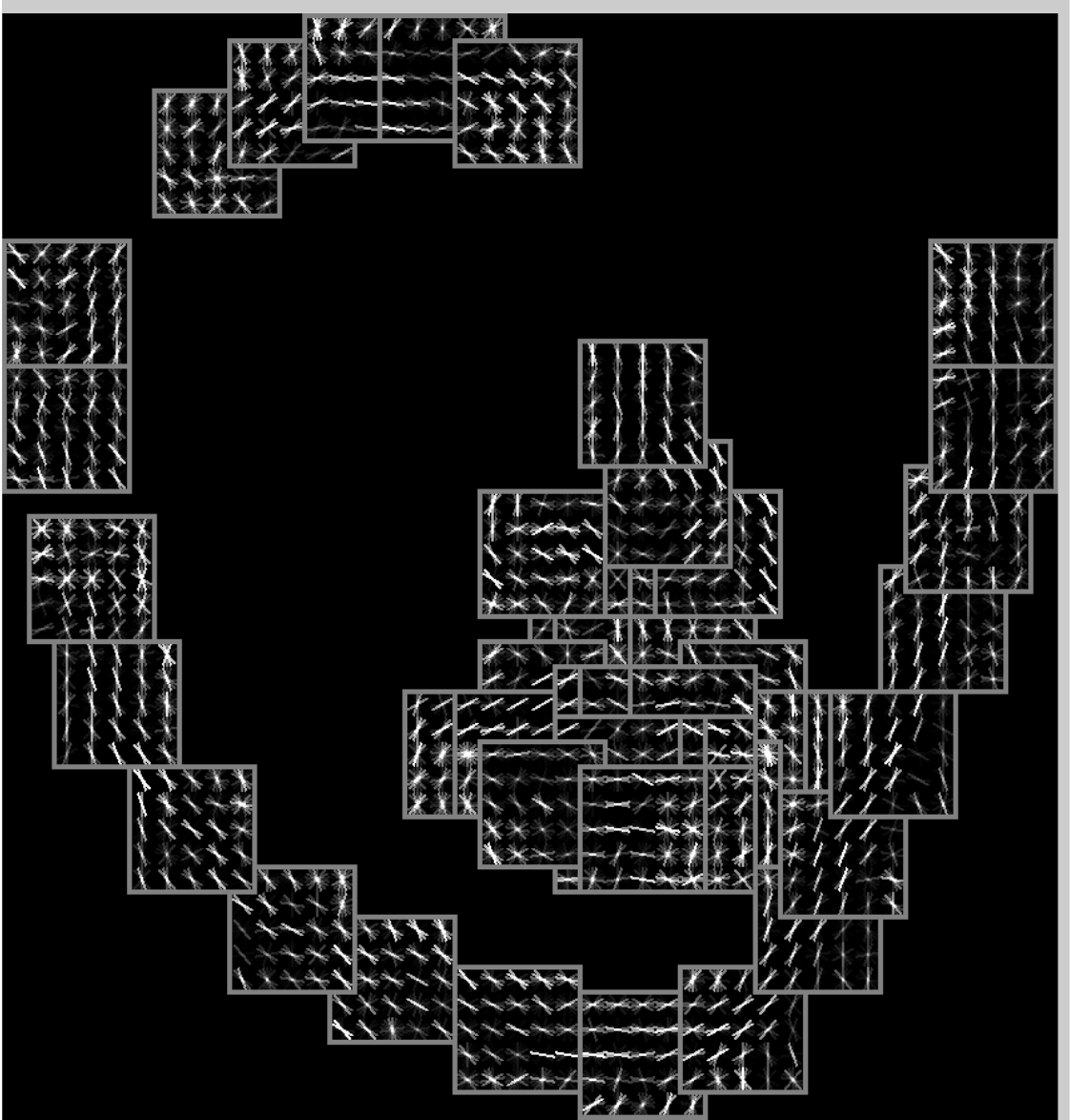}&
\includegraphics[width=.3\columnwidth]{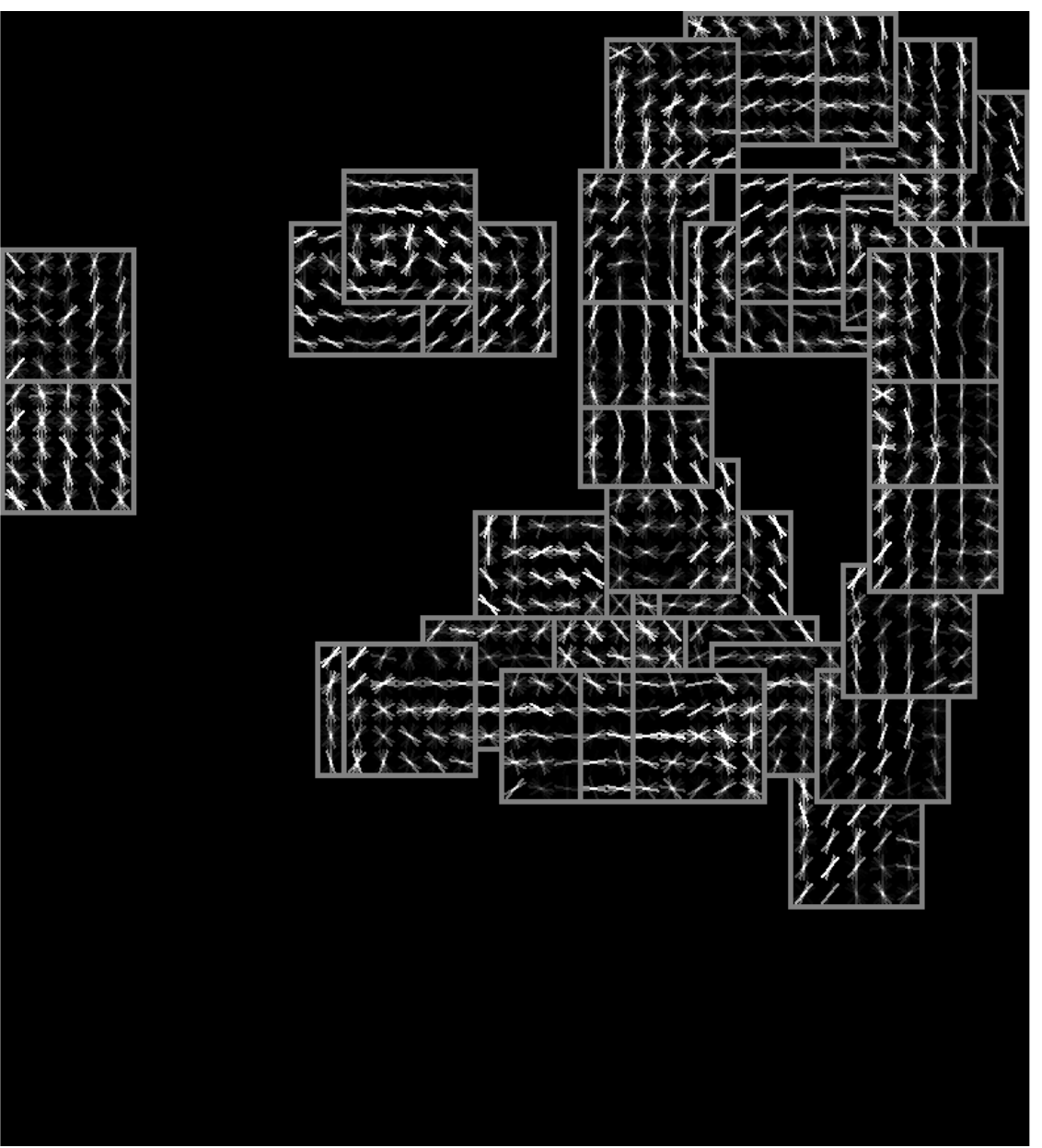}\\
\end{tabular}
\caption{Our model consists of a tree of parts (black circles) each of which is
connected to a set of landmarks (green or red) in a star topology.  The
examples here show templates corresponding to different choices of part shape
and occlusion patterns. Red indicate occluded landmarks. Shape parameters are
independent of occlusion state.  Landmark appearance is modeled with a small 
HOG template (2nd row) and occluded landmarks are constrained to have an
appearance template fixed to 0.  Note how the model produces a wide range of
plausible shape configurations and occlusion patterns.}
\label{fig:shapes}
\end{figure*}

\section{Related Work}

\paragraph{Face Detection and Localization}
There is a long history of face detection in the computer vision literature.  A
classic approach treats detection as problem aligning a model to a test image
using techniques such as Deformable Templates \cite{yuille1992feature}, Active
Appearance Models (AAMs)
\cite{cootes2001active,matthews2004active,milborrow2008locating} and elastic
graph matching~\cite{wiskott1997face}.  Alignment with full 3D models provides
even richer information at the cost of additional
computation~\cite{gu20063d,blanz2003face}.  A key difficulty in many of these
approaches is the dependence on iterative and local search techniques for
optimizing model alignment with a query image.  This typically results in high
computational cost and the concern that local minima may undermine system
performance.

Recently, approaches based on {\em pose regression}, which train
regressors that predict landmark locations from both appearance and spatial
context provided by other detector responses, has also shown impressive
performance \cite{valstar2010facial,efraty2011facial,belhumeur2011localizing,
burgos2013robust, cao2012face,dantone2012real, xiong2013supervised,ren2014face,zhu2015face}.
While these approaches lack an explicit model of face shape, stage-wise
pose-regression models can be trained efficiently in a discriminative fashion
and thus sidestep the optimization problems of global model alignment while
providing fast, feed-forward performance at test time.

Pose-regression is flexible in the choice of features and regressors used.
Supervised Descent Method (SDM) \cite{xiong2013supervised} employs linear
regression on SIFT features to compute shape increments. ESR \cite{cao2012face}
and RCPR \cite{burgos2013robust} predict shape increments using simple
pixel-difference features and boosted ferns.  LBF \cite{ren2014face} learns a
set of binary features and a regression function using random forest
regression.  Zhu et al. proposed a Coarse-to-Fine Shape Searching method (CFSS)
\cite{zhu2015face} in which at each stage a cascade of linear regressors are
used to calculate a finer sub-space (represented as a center and scope).
The incorporation of Deep Convolutional Neural Network features has allowed
further improvements by using raw image pixels as input instead of hand-designed
features and allows end-to-end training.  Zhang et al. proposed successive
auto-encoder networks (CFAN) to perform coarse-to-fine alignment
\cite{zhang2014coarse}.  TCDCN \cite{zhang2016learning} train a multi-task DCNN
jointly for landmark localization along with prediction of other facial
attributes.  They show that facial attributes such as gender and expression can
help in learning a robust landmark detector.

Our model is most closely related to the work of \cite{zhu2012face}, which
applies discriminatively trained deformable part models (DPM)
\cite{felzenszwalb2010object} to face analysis.  This
offers an intermediate between the extremes of model alignment and landmark  
regression by utilizing mixtures of simplified shape models that make
efficient global optimization of part placements feasible while exploiting
discriminative training criteria. Similar to \cite{yang2013pose}, we use local
part and landmark mixtures to encode richer multi-modal shape distributions.
We extend this line of work by adding hierarchical structure and explicit
occlusion to the model. We introduce intermediate part nodes that do not have
an associated ``root template'' but instead serve to encode an intermediate
representation of occlusion and shape state.  The notion of hierarchical part
models has been explored extensively as a tool for compositional representation
and parameter sharing (see e.g., \cite{zhu2010learning,girshick2011object}).
While the intermediate state represented in such models can often be formally
encoded in by non-hierarchical models with expanded state spaces and tied
parameters, our experiments show that the particular choice of model structure
proves essential for efficient representation and inference. 

\paragraph{Occlusion Modeling}
Modeling occlusion is a natural fit for recognition systems with an explicit
representation of parts. Work on generative constellation
models~\cite{weber2000towards,fergus2003object} learned parameters of a full
joint distribution over the probability of part occlusion and relied on brute
force enumeration for inference, a strategy that doesn't scale to large numbers
of landmarks.  More commonly, part occlusions are treated independently which
makes computation and representation more efficient.  For example, the
supervised detection model of~\cite{azizpour2012object} associates with each
part a binary variable indicating occlusion and learns a corresponding
appearance template for the occluded state.

The authors of \cite{girshick2011object} impose a more structured distribution
on the possible occlusion patterns by specifying grammar that generates a
person detector as a variable length vertical chain of parts terminated by an
occluder template, while \cite{Chen_CVPR15} allows ``flexible compositions''
which correspond to occlusion patterns that leave visible a connected subgraph
of the original tree-structure part model.  Our approach provides a stronger
model than full independence, capturing correlations between occlusions of
non-neighboring landmarks.  Unlike the grammar-based approach, occlusion
patterns are not specified structurally but instead learned from data and
encoded in the model weights.

Pose regression approaches have also been adapted to incorporate explicit
occlusion modeling.  For example, the face model of
\cite{saragih2011deformable} uses a robust m-estimator which serves to truncate
part responses that fall below a certain threshold.  In our experiments, we
compare our results to the recent work of \cite{burgos2013robust} which uses
occlusion annotations when training a cascade of regressors where each layer
predicts both part locations and occlusion states.

\section{Hierarchical Part Model}
\label{sec:model}

In this section we develop a hierarchical part model that simultaneously
captures face appearance, shape and occlusion.  Fig.~\ref{fig:shapes} shows a
graphical depiction of the model structure.  The model has two layers: the face
consists of a collection of parts (nose, eyes, lips) each of which is in turn
composed of a number of landmarks that specify local edge features making up
the part.  Landmarks are connected to their parent part nodes with a star
topology while the connections between parts forms a tree.
In addition to location, each part takes one of a discrete set of shape states
(corresponding to different facial shapes or expressions) and occlusion states
(corresponding to different patterns of visibility).  The model topology which
groups facial features into parts was specified by hand while the shape and
occlusion patterns are learned automatically from training data (see Section
\ref{sec:learning}).  This model, which we term a hierarchical part model (HPM)
is a close cousin of the deformable part model (DPM) of
\cite{felzenszwalb2010object} and the flexible part model (FMP) of
\cite{zhu2012face}.  It differs in the addition of part nodes that model shape
but don't include any ``root filter'' appearance term, and by the use of
mixtures to model occlusion patterns for each part.  In this section we
introduce some formal notation to describe the model and some important
algorithmic details for performing efficient message passing during inference. 

\subsection{Model Structure}
Let $l,s,o$ denote the hypothesized locations, shape and occlusion of $N_p$ parts
and $N_l$ landmarks describing the face.  Locations $l \in \mathbb{R}^{2N}$ range
over the whole image domain and $o \in \mathcal{O}_1 \times \mathcal{O}_2 \ldots
\times \mathcal{O}_N$ indicates the occlusion states of parts and landmarks and
$N = N_p + N_l$.
The shape $s \in \mathcal{S}_1 \times \mathcal{S}_2 \ldots
\times \mathcal{S}_N$ selects one of a discrete set of shape mixture components
for each part.  We define a tree-structured scoring function by: 
\begin{align}
  &Q(l,s,o|I) = \sum\limits_{i} \phi_{i}(l_i,s_i,o_i|I) 
  \label{eqn:score}
  \\
  & + \sum\limits_i \sum\limits_{j \in child(i)} \psi_{ij}(l_i,l_j,s_i,s_j) + b_{ij}(s_i,s_j,o_i,o_j) 
  \nonumber
\end{align}
where the potential $\phi$ scores the consistency of the local image appearance
around location $l_i$, $\psi$ is a quadratic shape deformation penalty, and $b$
is a co-occurrence bias.

The first (unary) term scores the appearance evidence.  We linearly
parameterize the unary appearance term with filter weights $w_i^{s_i}$ 
that depend on the discrete shape mixture selected
\[
\phi_{i}(l_i,s_i,o_i|I) = w_i^{s_i} \cdot \phi(l_i,o_i|I)
\]
Appearance templates are only associated with the leaves (landmarks) in the
model so the unary term only sums over those leaf nodes.  The occlusion
variables $o_i$ for the landmarks are binary, corresponding to either occluded
or visible.  If the $i$th landmark is unoccluded, the appearance feature $\phi$
is given by a HOG~\cite{dalal2005histograms} feature extracted at location
$l_i$, otherwise the feature is set to $0$. This is natural on theoretical
grounds since the appearance of the occluder is arbitrary and hence
indistinguishable from background based on its local appearance.  Empirically
we have found that unconstrained occluder templates learned with sufficiently
varied data do in fact have very small norms, further justifying this choice
\cite{ghiasiYRF2014parsing}. 

The second (pairwise) term in Eq. 1 scores the placement part $j$
based on its location relative to its parent $i$ and the shape 
mixtures of the child and parent.  We model this with a linearly parameterized
function:
\[
\psi_{ij}(l_i,l_j,s_i,s_j) = w_{ij}^{s_i,s_j} \cdot \psi(l_i - l_j)
\]
where the feature $\psi$ includes the $x$ and $y$ displacements and their
cross-terms, allowing the weights $w_{ij}$ to encode a standard quadratic
``spring''.  We assume that the shape of the parts is independent of any
occluder so the spring weights do not depend on the occlusion states.
\footnote{In practice we find it is sufficient for the deformation cost to only
depend on the child shape mixture, i.e. 
$\psi_{ij}(l_i,l_j,s_i,s_j) = w_{ij}^{s_j} \cdot \psi(l_i - l_j)$
which gives a factor $S$ speedup with little decrease in performance.}
The pairwise parameter $b_{ij}$ encodes a bias of particular occlusion patterns
and shapes to co-occur.  Formally, each landmark has the same number of
occlusion states and shape mixtures as its parent part, but we fix the bias
parameters between the part and its constituent landmarks to impose a hard
constraint that the mixture assignments are compatible.

\begin{figure}[t!]
\includegraphics[width=0.9\columnwidth]{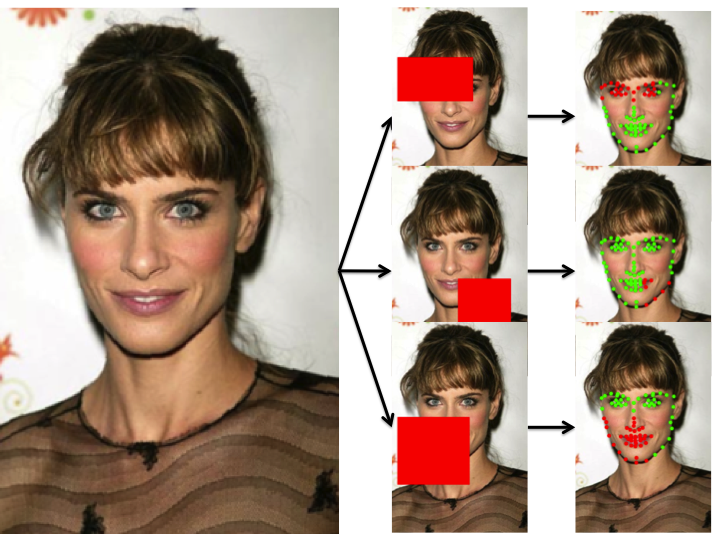}
\caption{
Virtual positive examples are generated synthetically by starting with a fully
visible training example and sampling random coherent occlusion patterns.}
\label{fig:occgen}
\end{figure}

\subsection{Efficient Message Passing}
The model above can be made formally equivalent to the FMP model used in
\cite{yang2013pose} by introducing local mixture variables that live in
the cross-product space of $o_i$ and $s_i$.  However, this reduction fails to
exploit the structure of the occlusion model.  This is particularly important due
to the large size of the model.  Naive inference is quite slow due to the large
number of landmarks and parts (N=68+10), and huge state space for each node which
includes location, occlusion pattern and shape mixtures.  Consider the message
passed from one part to another where each part has $L$ possible locations, $S$
shape mixtures and $O$ occlusion patterns.  In general this requires minimizing
over functions of size $(LSO)^2$ or $L(SO)^2$ when using the distance
transform. In the models we test, $SO=12$ which poses a substantial computation
and memory cost, particularly for high-resolution images where $L$ is large.

\paragraph{Part-Part messages} 
While the factorization of shape and occlusion doesn't change the asymptotic
complexity, we can reduce the runtime in practice by exploiting distributivity
of the distance transform over $\max$ to share computations.  Standard message
passing from part $j$ to part $i$ requires that we compute:
\begin{align*}
&\mu_{j \rightarrow i}(l_i,s_i,o_i) = \max_{l_j,s_j,o_j} \biggl[ \psi_{ij}(l_i,l_j,s_i,s_j)  \\
&  + \sum_{k \in child(j)} \mu_{k \rightarrow j}(l_j,s_j,o_j) + b_{ij}(s_i,s_j,o_i,o_j)\biggr] 
\end{align*}
where we have dropped the unary term $\phi_j$ which is $0$ for parts. Since
the bias doesn't depend on the location of parts we can carry out the computation in two
steps:
\begin{align*}
&\nu_{ij}(l_j,s_i,s_j,o_j) =\\
&  \quad \quad \max_{l_j} \left[\psi_{ij}(l_i,l_j,s_i,s_j) + \sum_{k \in child(j)} \mu_{k \rightarrow j}(l_j,s_j,o_j) \right]\\
&\\
&\mu_{j \rightarrow i}(l_i,s_i,o_i) = \max_{s_j,o_j} \left[\nu_{ij}(l_j,s_i,s_j,o_j) + b_{ij}(s_i,s_j,o_i,o_j) \right] 
\end{align*}
which only requires computing $S^2O$ distance transforms.

\paragraph{Landmark-Part messages} 
In our model the occlusion and shape variables for a landmark are determined
completely by the parent part state.  Since the score is known for an occluded
landmark in advance, it is not necessary to compute distance transforms for
those components. We write this computation as:
\begin{align*}
&\nu_{jk}(l_j,s_j,o_j) = \left\{
\begin{array}{ll}
0 \quad \quad \text{ if k occluded in $o_j$} \\
\max_{l_k} \psi_{jk}(l_j,l_k,s_j,s_j) + \phi_k(l_k,s_j,o_j|I) \\
\end{array}
\right.\\
&\mu_{k \rightarrow j}(l_j,s_j,o_j) = \nu_{jk}(l_k,s_j,o_j) + b_{jk}(s_j,o_j,s_j,o_j) 
\end{align*}
Where we have used the notation to explicitly capture the constraint that
landmark shape and occlusion mixtures $(s_k,o_k)$ must match those of the
parent part $(s_j,o_j)$.  In our models, this reduces the memory and inference
time by roughly a factor of 2, a savings that becomes increasingly significant
as the number of occlusion mixtures grows.

\subsection{Global Mixtures for Viewpoint and Resolution} 
Viewpoint and image resolution are the largest sources of variability in the
appearance and relative location of landmarks. To capture this, we use a
mixture over head poses. 
These ``global'' mixtures can be represented with the same notation as above by expanding
the state-space of the shape variables to be the cross product of the set of
local shapes for part $i$ and the global viewpoint for the model (i.e., $s_i
\in \mathcal{S}_i \times \mathcal{V}$) and fixing some entries of the bias
$b_{ij}$ to be $-\infty$ to prevent mixing of local shapes from different
viewpoints. In our implementation we tie parameters to enforce the left- and
right-facing models to be mirror symmetric.

The HPM model we have described includes a large number of landmarks.  While
this is appropriate for high resolution imagery, it does not perform well in
detecting and modeling low resolution faces ($<\!150$ pixels tall).  To address
this we introduce an additional global mixture component for each viewpoint
that corresponds to low-resolution HPM model consisting of a single
half-resolution template for each part and no landmark templates. This 
mixture is trained jointly with the full resolution model using the strategy
described in \cite{park2010multiresolution}.

\section{Model Training and Inference}
\label{sec:learning}

The potentials in our shape model are linearly parameterized, allowing
efficient training using an SVM solver \cite{felzenszwalb2010object}.  Face
viewpoint, landmark locations, shape and occlusion mixtures are completely
specified by pre-clustering the training data so that parameter learning is
fully supervised. We first describe how these supervised labels are derived 
from training data and how we synthesize ``virtual'' positive training examples
that include additional occlusion. We then discuss the details of the 
parameter learning and test-time prediction.

\subsection{Training Data}

We assume that a training data set of face images has been annotated with
landmark locations for each face. From such data we automatically generate
additional mixture labels specifying viewpoint, shape, and occlusion. We also
generate additional virtual training examples by synthesizing plausible
coherent occlusion patterns.

\paragraph{Viewpoint and Resolution Mixtures} 
To cluster training examples into a set of discrete viewpoints, we make use of
the MultiPIE dataset \cite{gross2010multi} which provides ground-truth
viewpoint annotations for a limited set of faces.  We perform Procrustes
alignment between each training example and examples in the MultiPIE database
and then transfer the viewpoint label from nearest MultiPIE example to the
training example.  In our experiments we used either 3 or 7 viewpoint clusters
(each viewpoint spans 15 degrees).
In addition to viewpoint, alignment to MultiPIE also provides a standard scale
normalization and removes in-plane rotations from the training set.  To train
the low-resolution mixture components, we use the same training data but
down-sample the input image by a factor of 2.

\begin{figure}[t!]
\begin{center}
	\begin{tabular}{|c|c|c|}
\hline
\includegraphics[width=0.19\columnwidth]{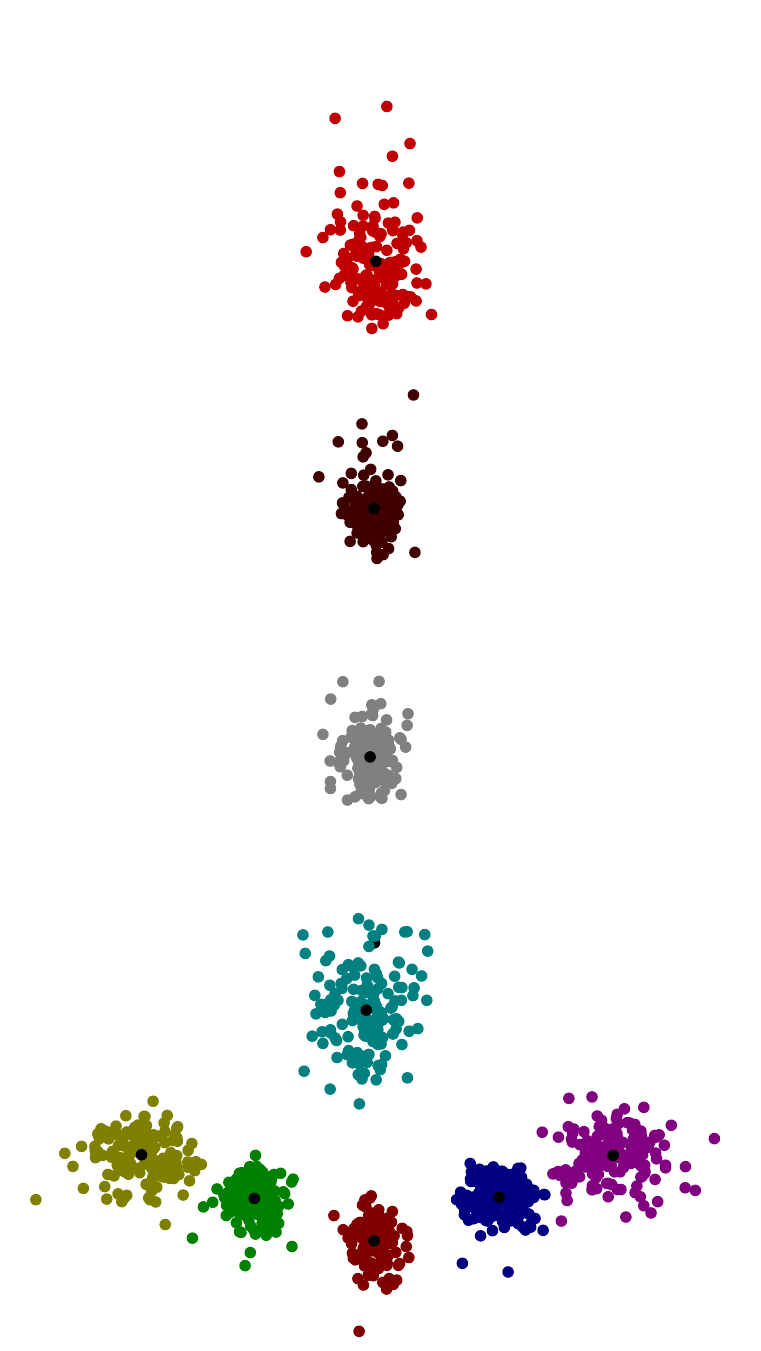}&
\includegraphics[width=0.19\columnwidth]{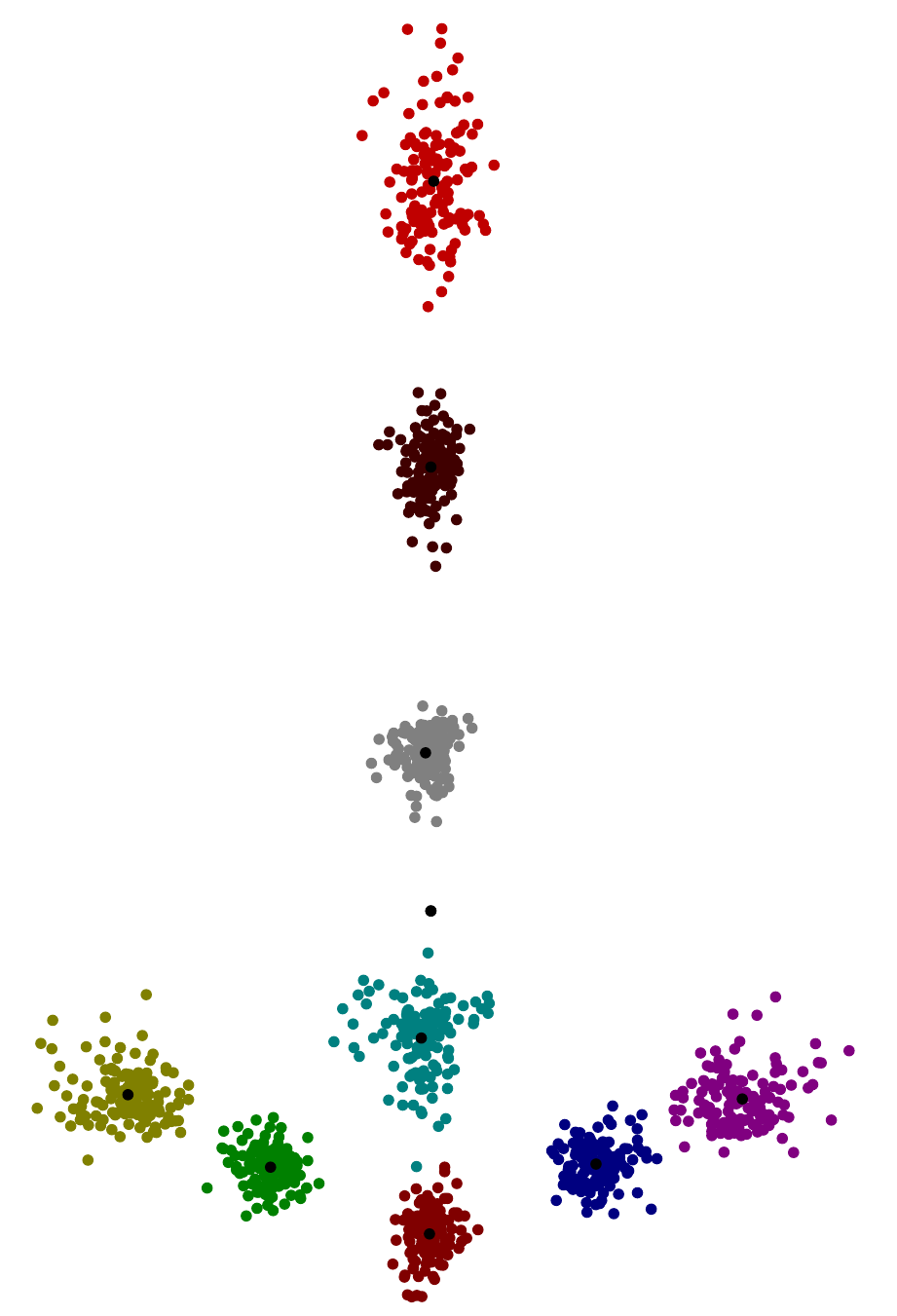}&
\includegraphics[width=0.19\columnwidth]{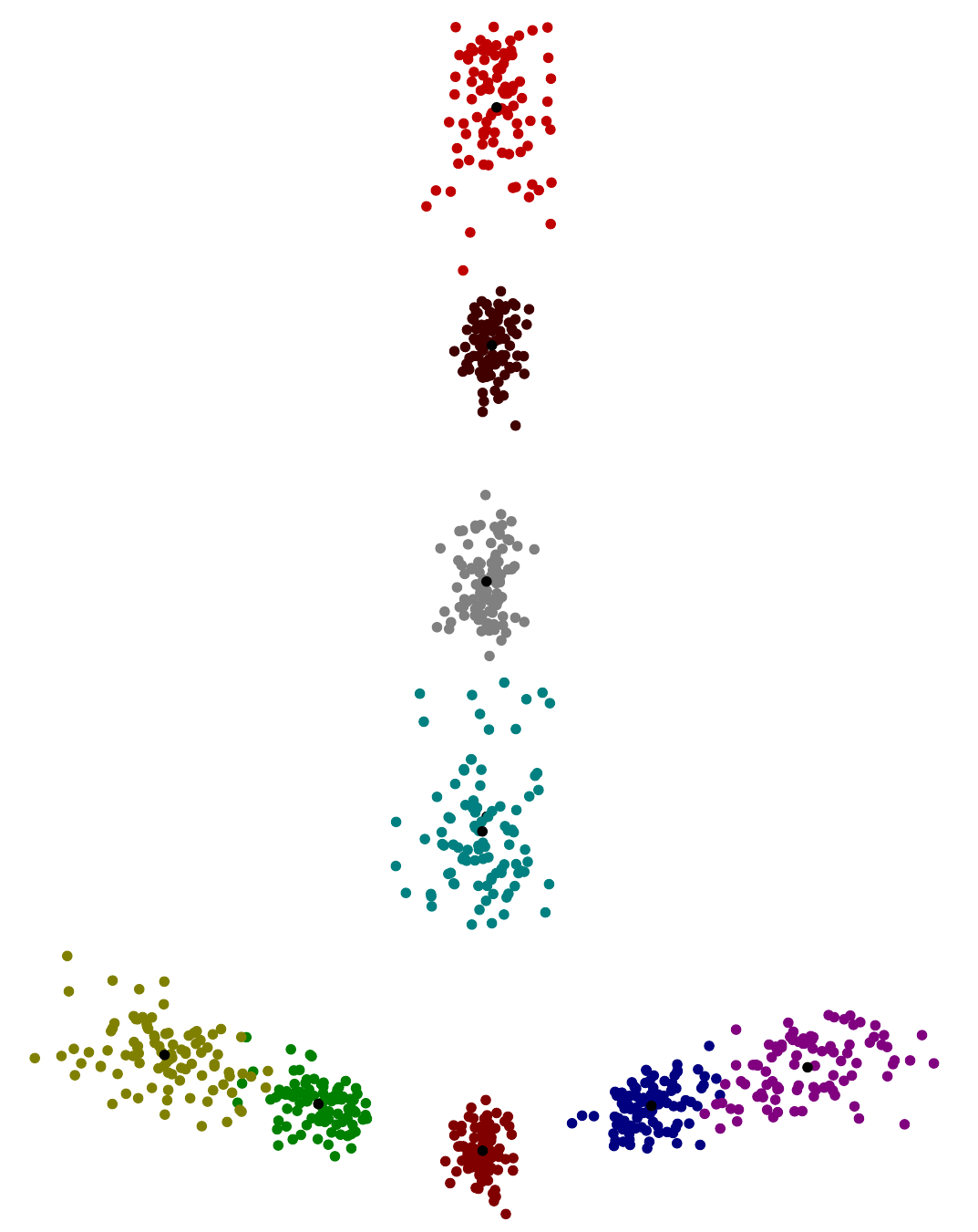}\\ \hline
\includegraphics[width=0.19\columnwidth]{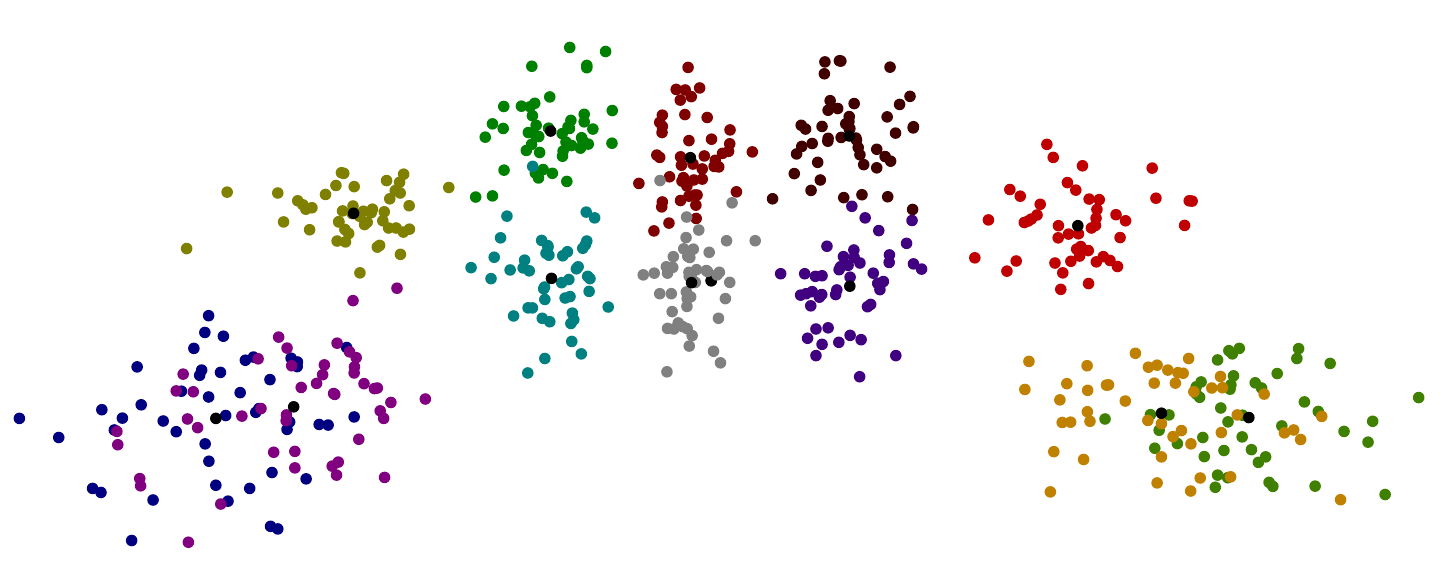}&
\includegraphics[width=0.19\columnwidth]{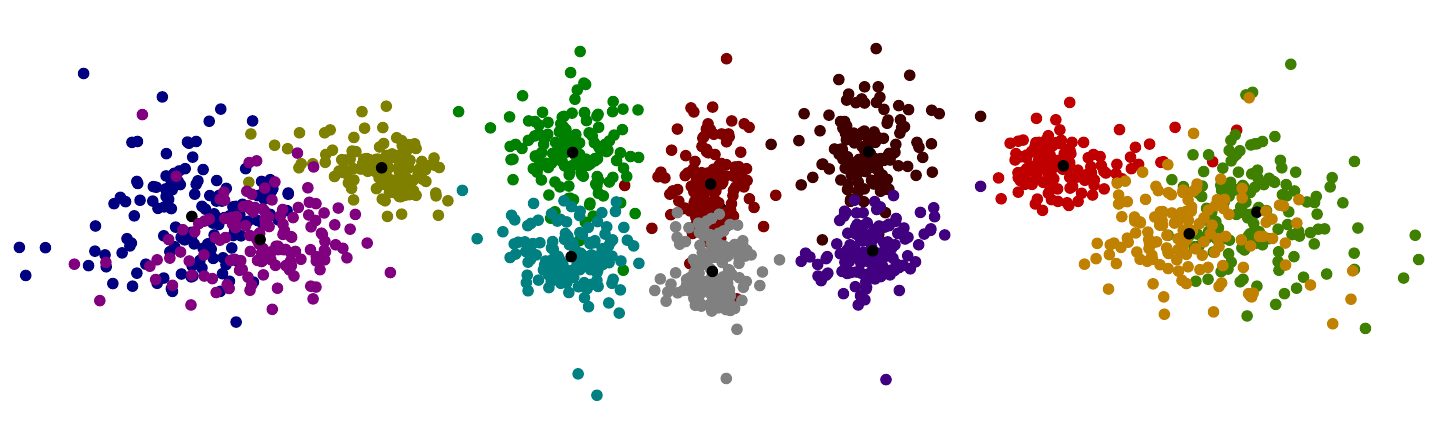}&
\includegraphics[width=0.19\columnwidth]{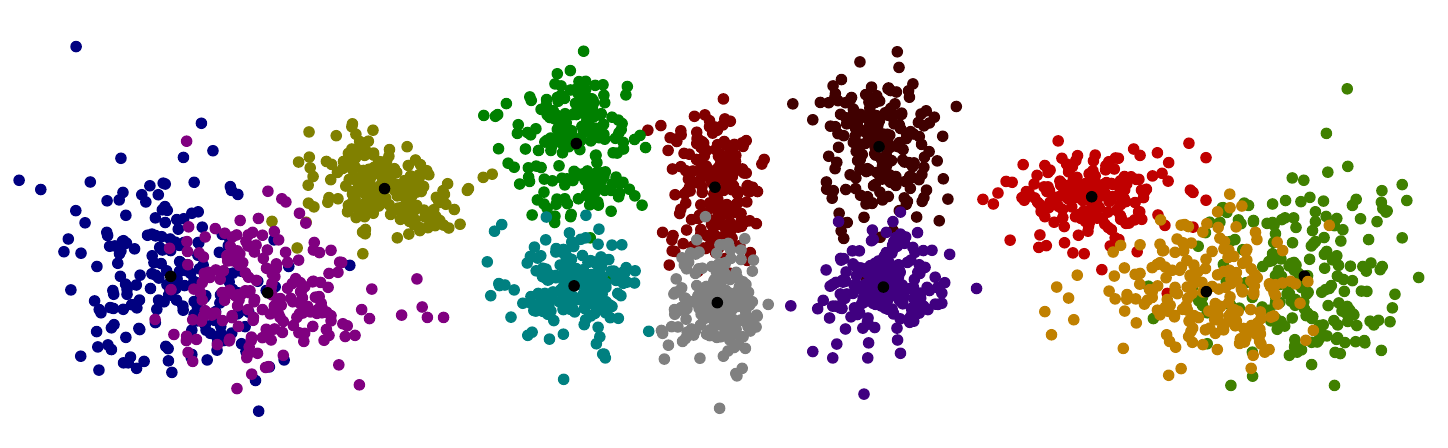}\\ \hline
\includegraphics[width=0.19\columnwidth]{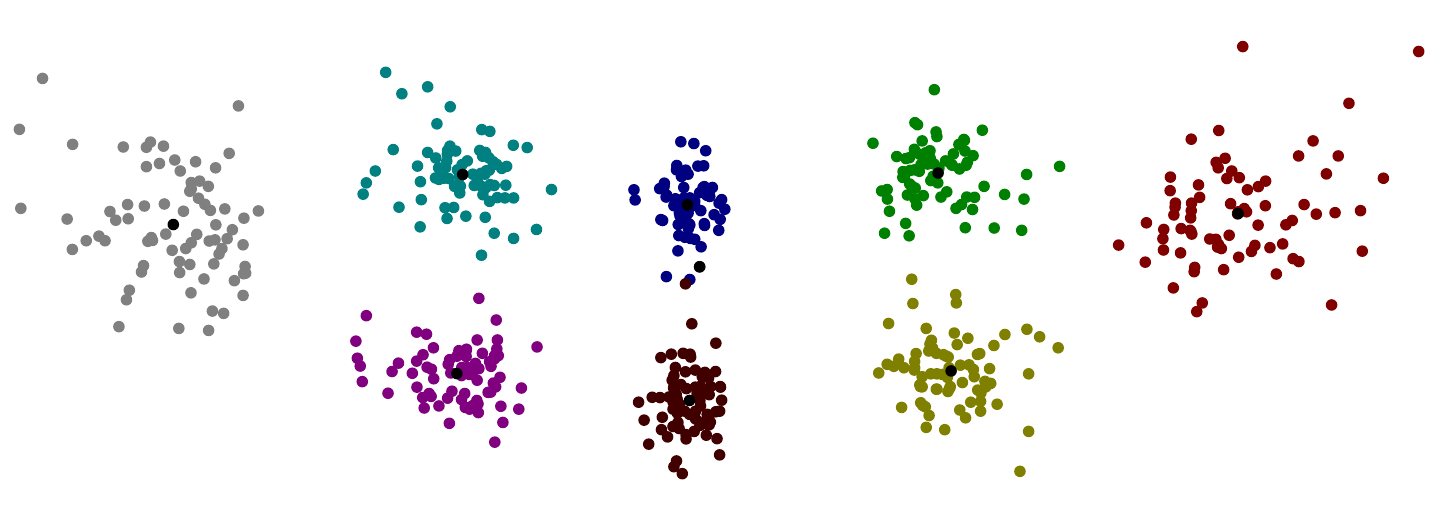}&
\includegraphics[width=0.19\columnwidth]{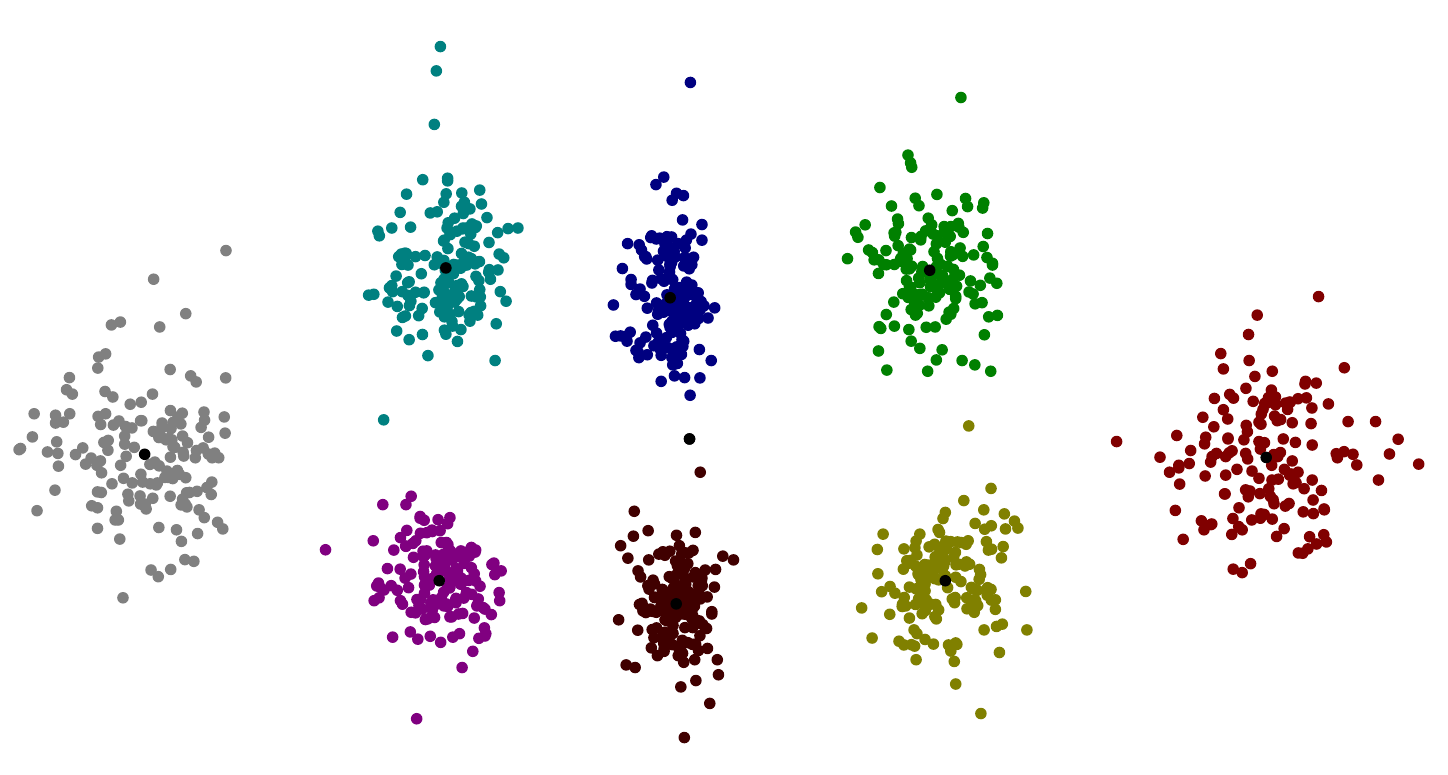}&
\includegraphics[width=0.19\columnwidth]{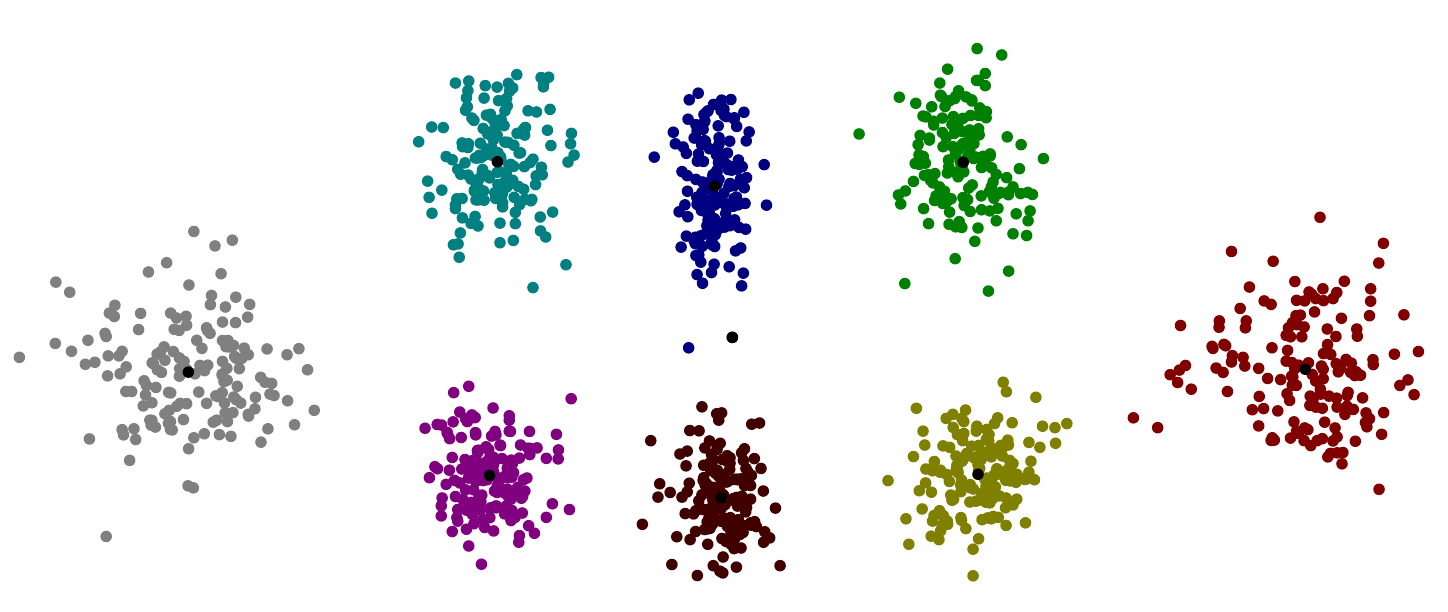}\\ \hline
  \end{tabular}
\end{center}
  \caption{Example shape clusters for face parts (nose, upper lip, lower lip).
  Co-occurrence biases for combinations of part shapes are learned
  automatically from training data. Different colored points correspond to
  location of each landmark relative to the part (centroid).}
  \label{fig:keyclusters}
\end{figure}

\paragraph{Part Shape and Occlusion Mixtures}
For each part and each viewpoint, we cluster the set of landmark configurations
in the training data in order to come up with a small number of shape mixtures
for that part.  The part shapes in the final model are represented by
displacements relative to a parent node so we subtract off the centroid of the
part landmarks from each training example prior to clustering.  The vectors
containing the coordinates of the centered landmarks are clustered using
k-means.  We imagine it would be efficient to allocate more mixtures to parts
and viewpoints that show greater variation in shape, but in the final model
tested here we use fixed allocation of $S=3$ shape mixtures per part per 
viewpoint.  Fig.~\ref{fig:keyclusters} shows example clusterings of part shapes
for the center view.

\paragraph{Synthetic Occlusion Patterns}
In the model each landmark is fully occluded or fully visible.  The occlusion
state of a part describes the occlusion of its constituent landmarks.  If there
are $N_l$ landmarks then there are $2^{N_l}$ possible occlusion patterns.  However,
many of these occlusions are quite unlikely (e.g., every other landmark
occluded) since occlusion is typically generated by an occluder object with a
regular, compact shape.

To model spatial coherence among the landmark occlusions, we
synthetically generate ``valid'' occlusions patterns by first sampling 
mean part and landmark locations from the model and then randomly sampling
a quarter-plane shaped occluder and setting as occluded those landmarks 
that fall behind the occluder.  Let $a,b$ be uniformly sampled from a tight box
surrounding the face. This selected origin point induces a partition of the
image into quadrants (i.e., $(x < a) \land (y < b)$, $(x \geq a) \land (y <
b)$, etc.).  We choose a quadrant at random and mark all landmarks falling in
that landmark as occluded.  While our occluder is somewhat ``boring'', it is
straightforward to incorporate more interesting shapes, e.g., by sampling from
a database of segmented objects.  Fig.~\ref{fig:occgen} shows example
occlusions generated for a training example.

In our experiments we generate $8$ synthetically occluded
examples for each original training example.  For each part in the model we
cluster the set of resulting binary vectors in order to generate a list of
valid  part occlusion patterns. The occlusion state for each landmark in a
training example is then set to be consistent with the assigned part occlusion
pattern.  In our experiments we utilized only $O=4$ occlusion
mixtures per part, typically corresponding to unoccluded, fully occluded and
two half occluded states whose structure depended on the part shape and 
location within the face.
\begin{figure*}[ht!]
		\includegraphics[height=0.16\textwidth]{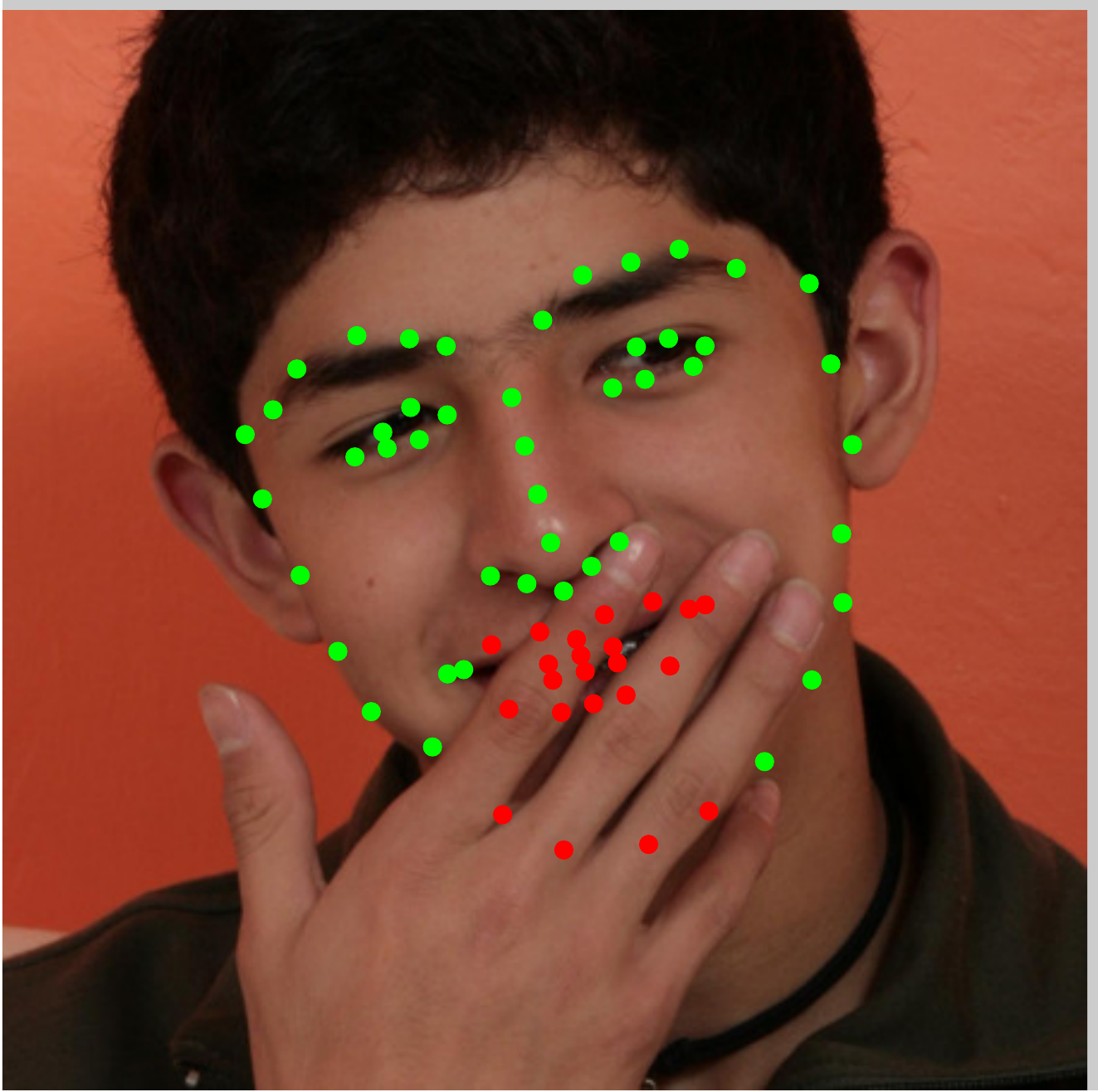} \hspace{-5pt}
		\includegraphics[height=0.16\textwidth]{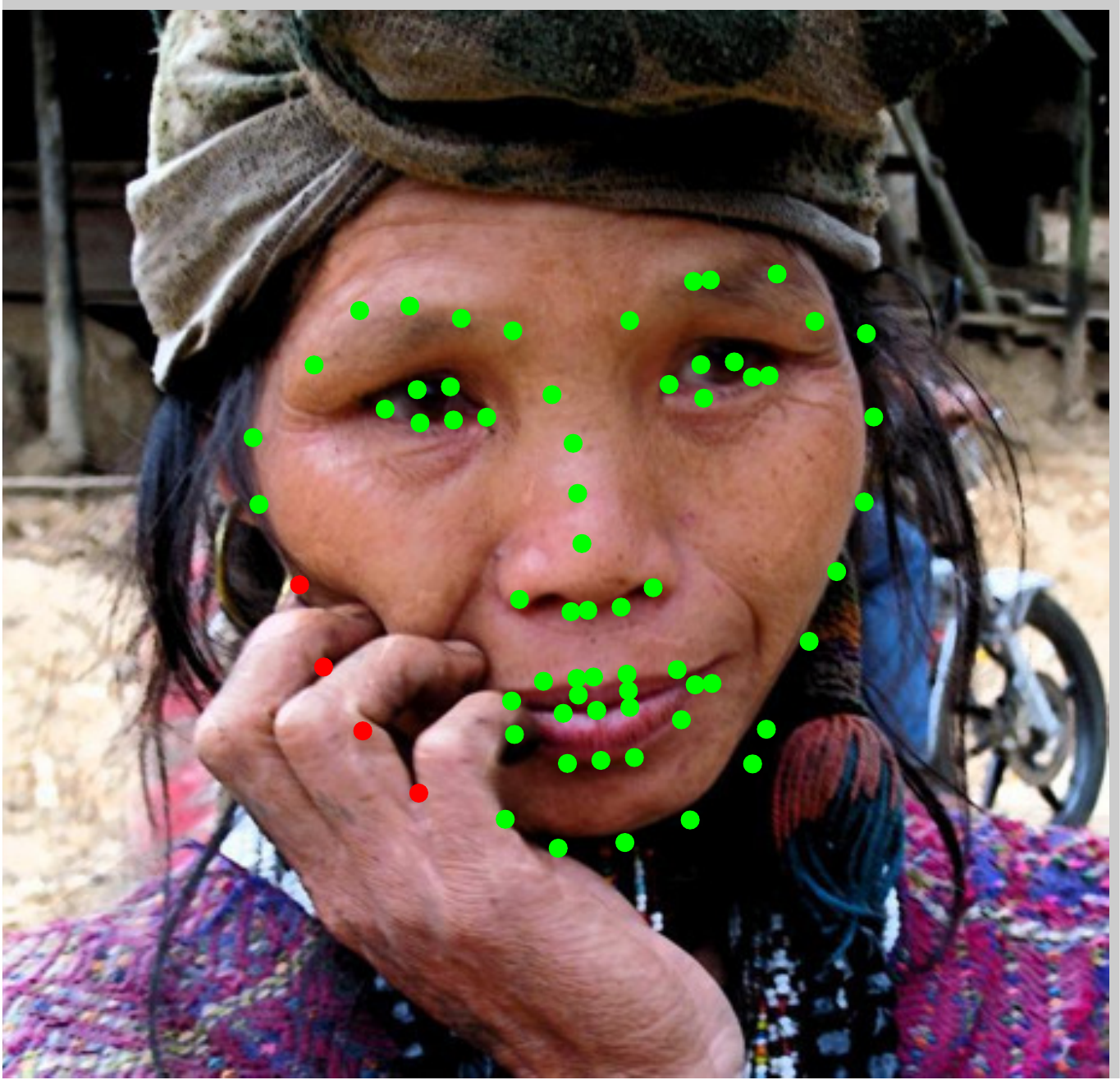} \hspace{-5pt}
		\includegraphics[height=0.16\textwidth]{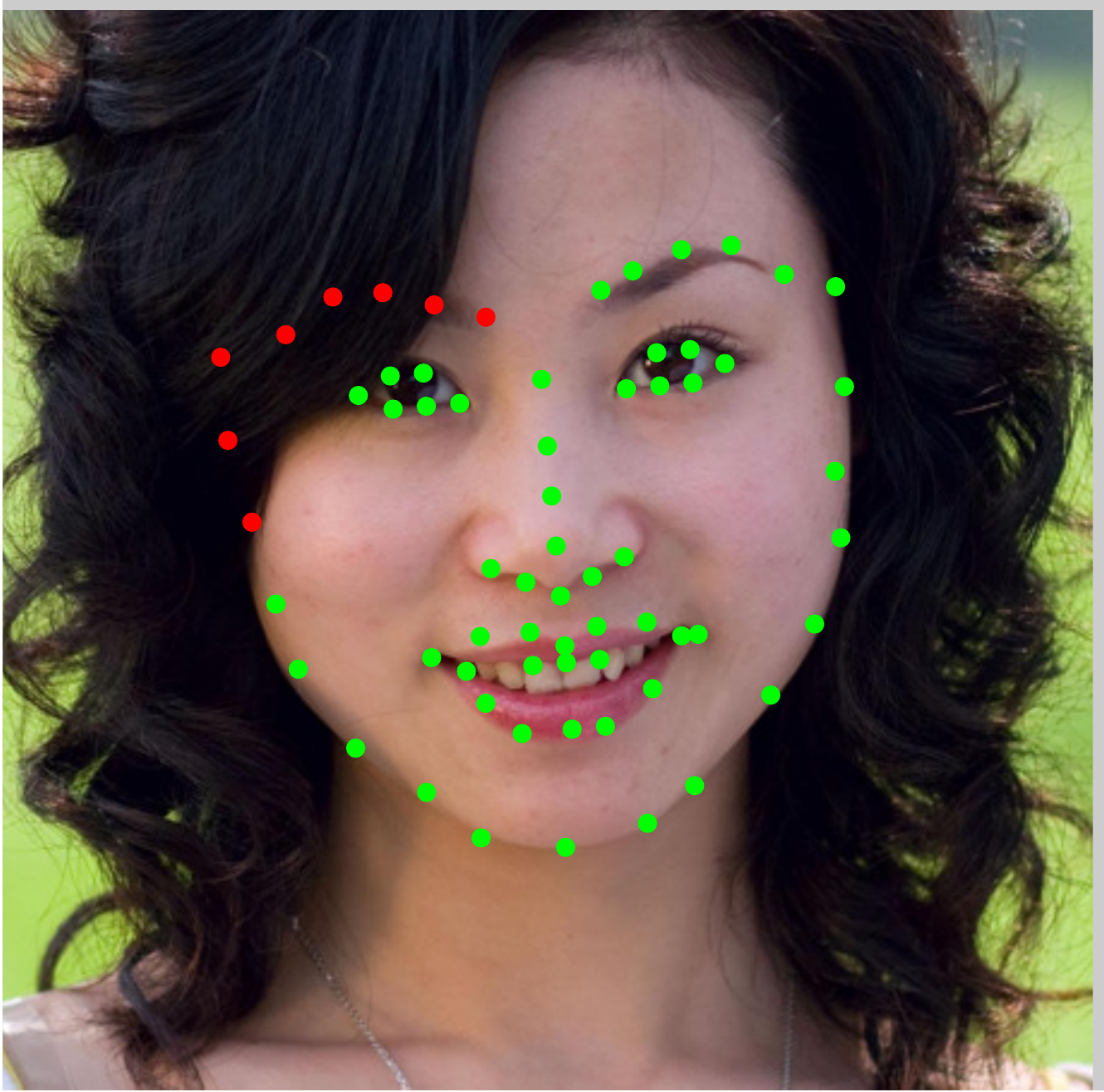} \hspace{-5pt}
		\includegraphics[height=0.16\textwidth]{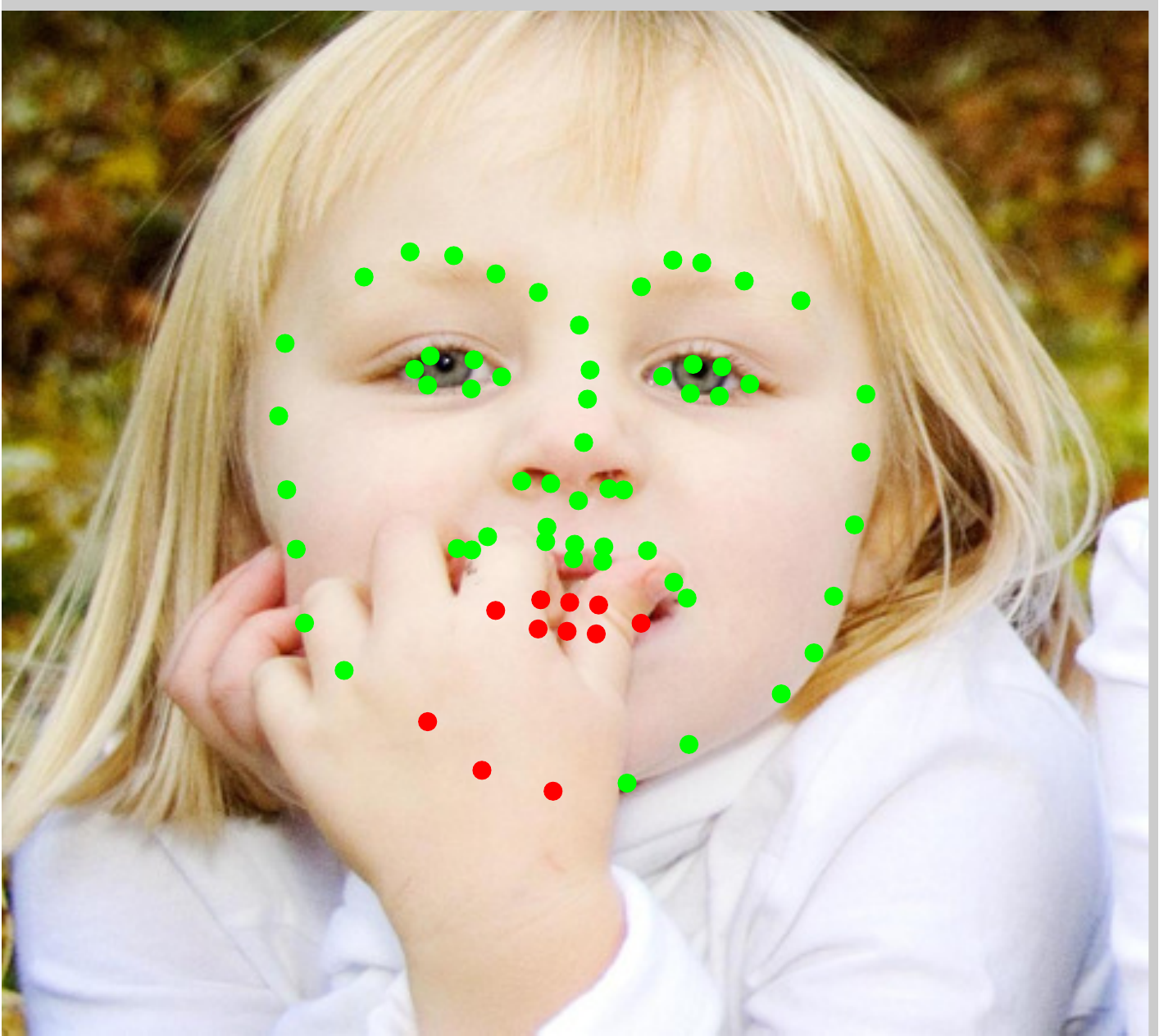} \hspace{-5pt}
		\includegraphics[height=0.16\textwidth]{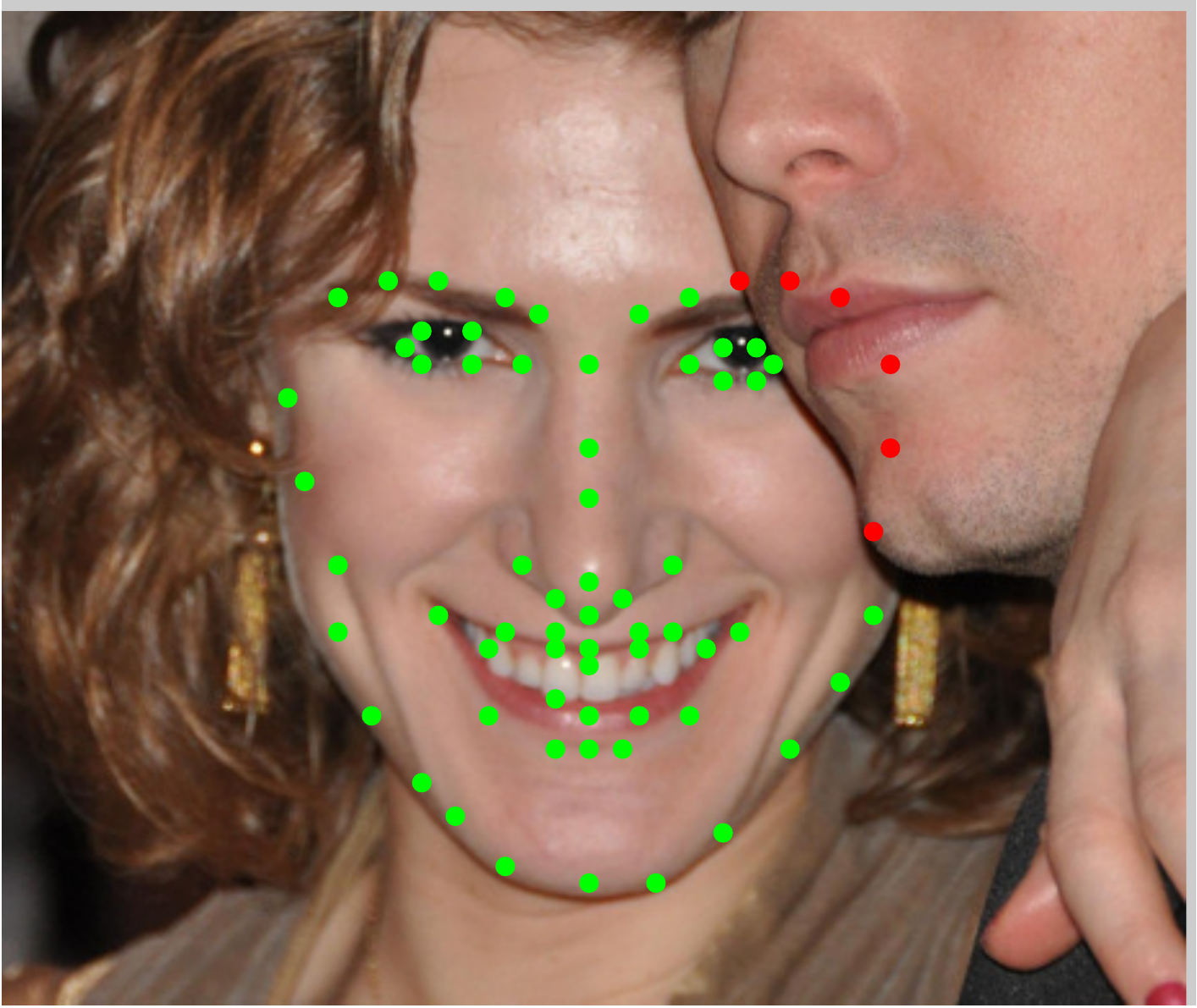} \hspace{-5pt}
		\includegraphics[height=0.16\textwidth]{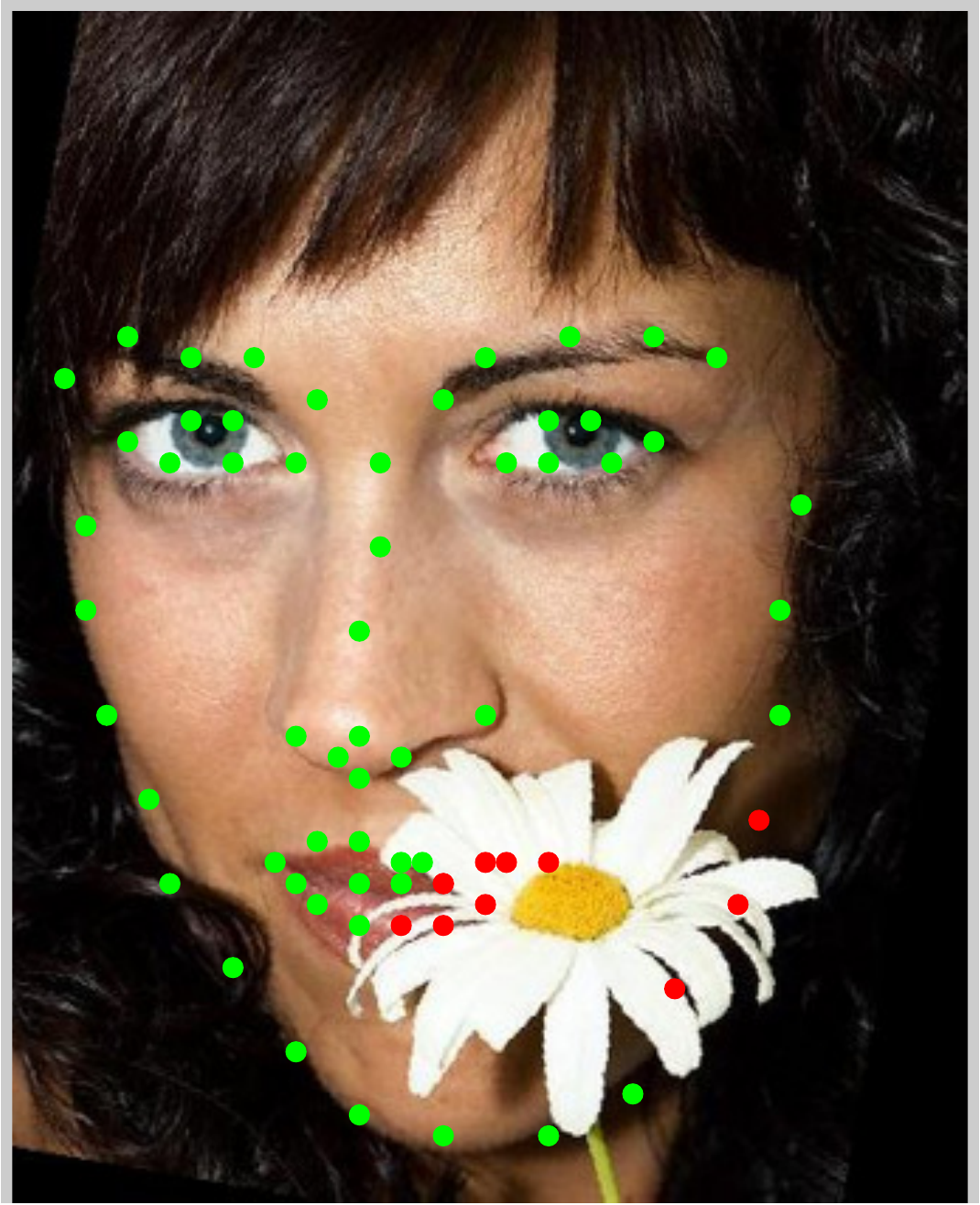}\\
		\includegraphics[height=0.16\textwidth]{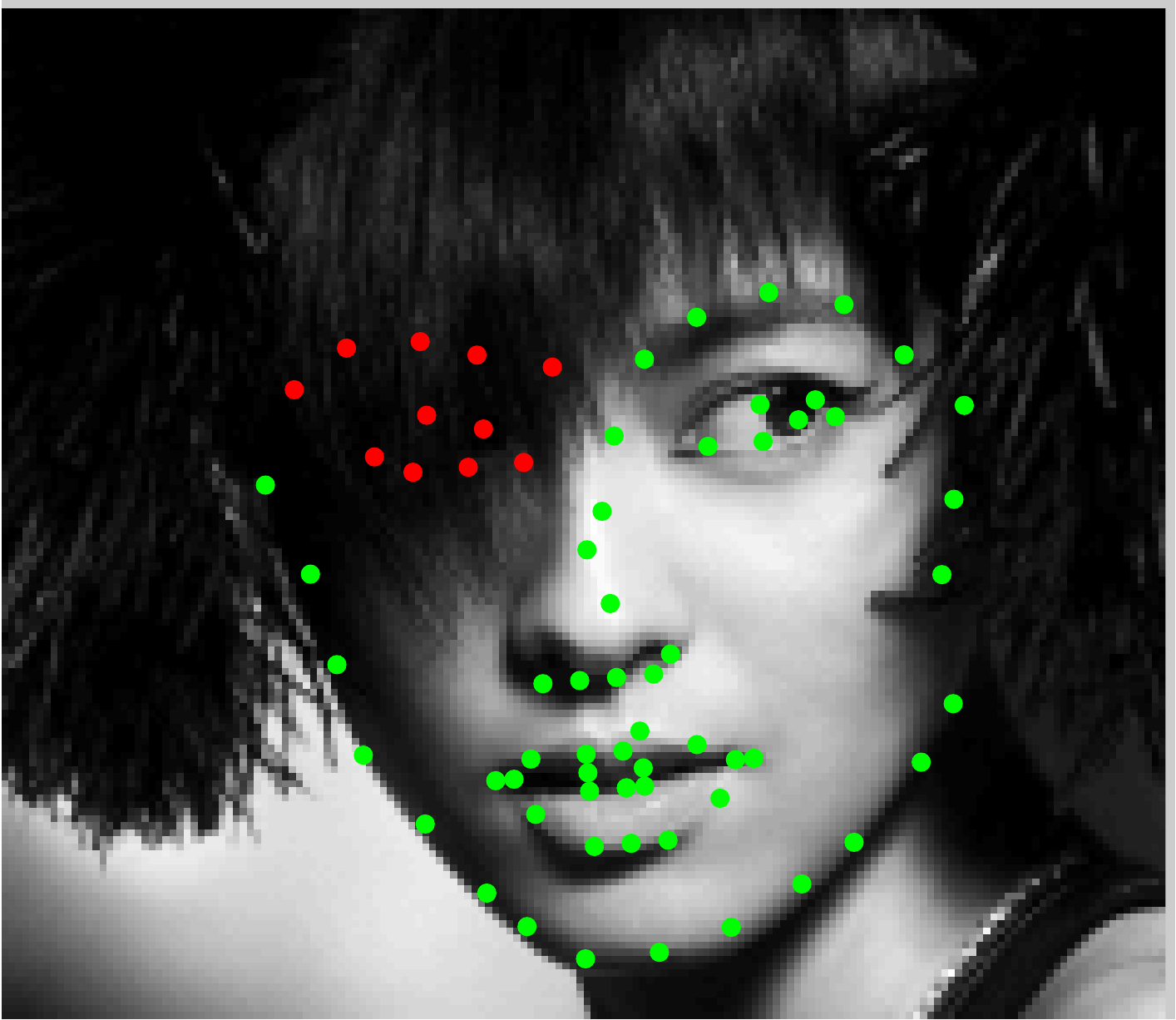} \hspace{-2pt}
		\includegraphics[height=0.16\textwidth]{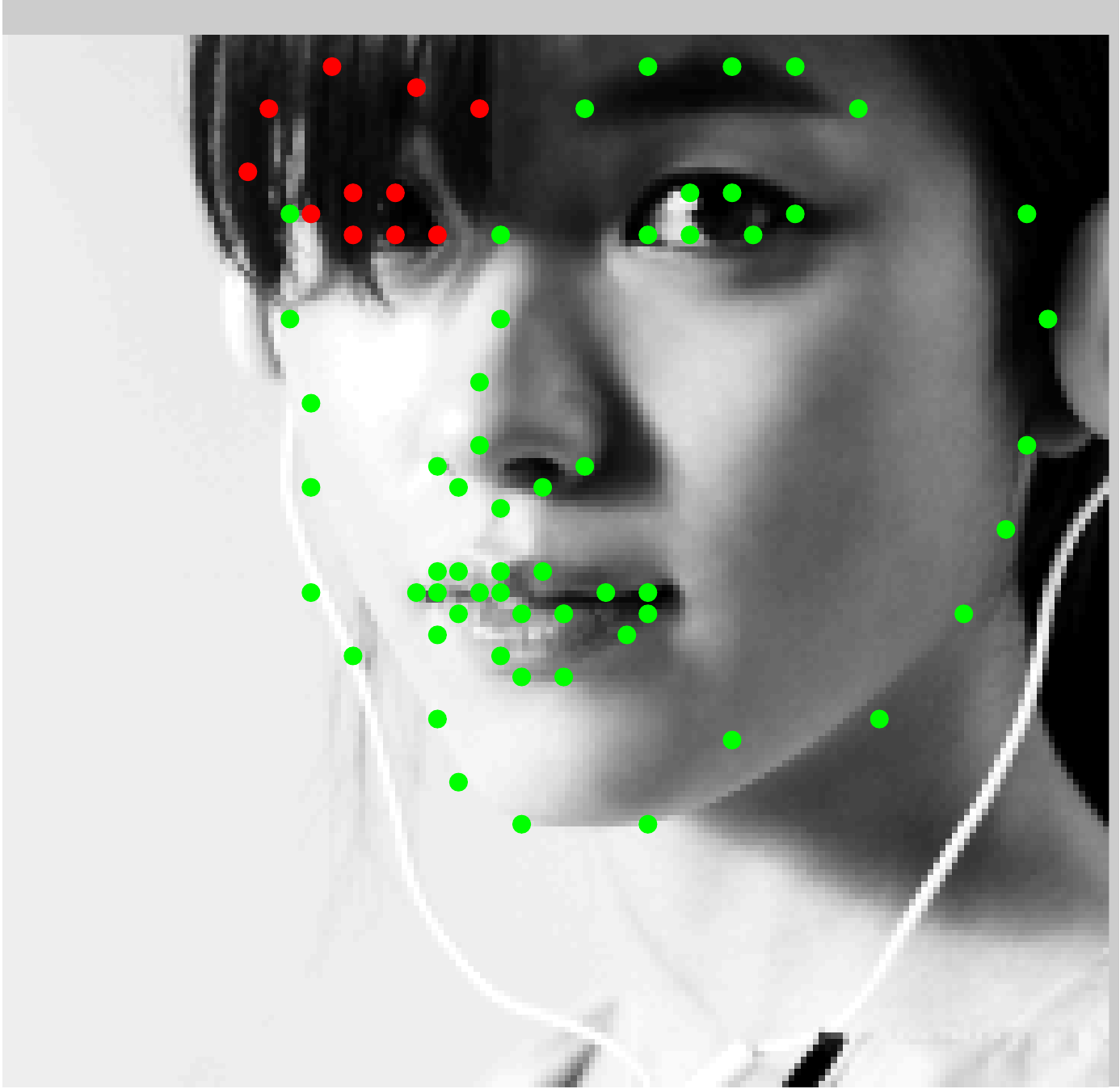} \hspace{-2pt}
		\includegraphics[height=0.16\textwidth]{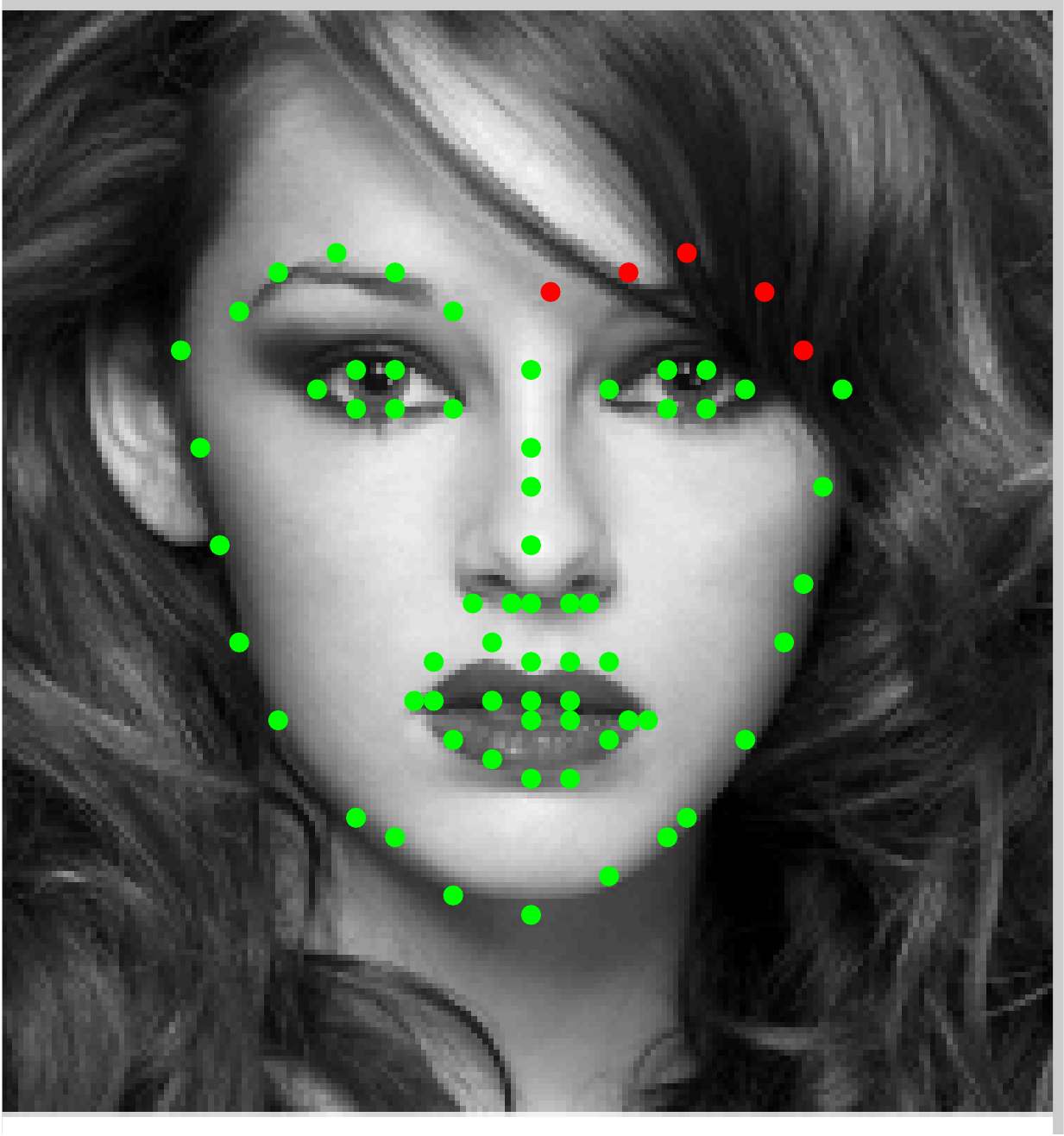} \hspace{-2pt}
		\includegraphics[height=0.16\textwidth]{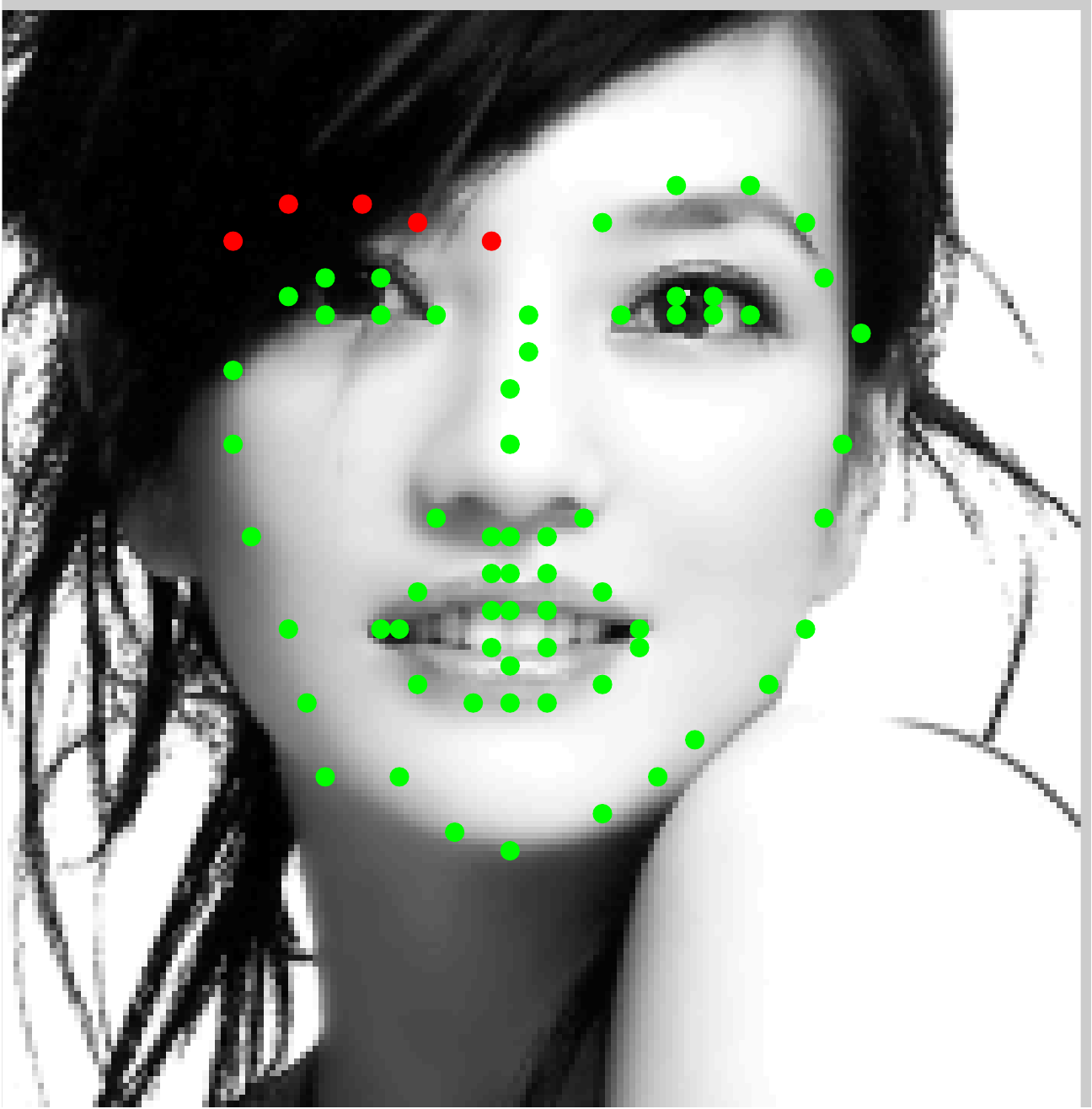} \hspace{-2pt}
		\includegraphics[height=0.16\textwidth]{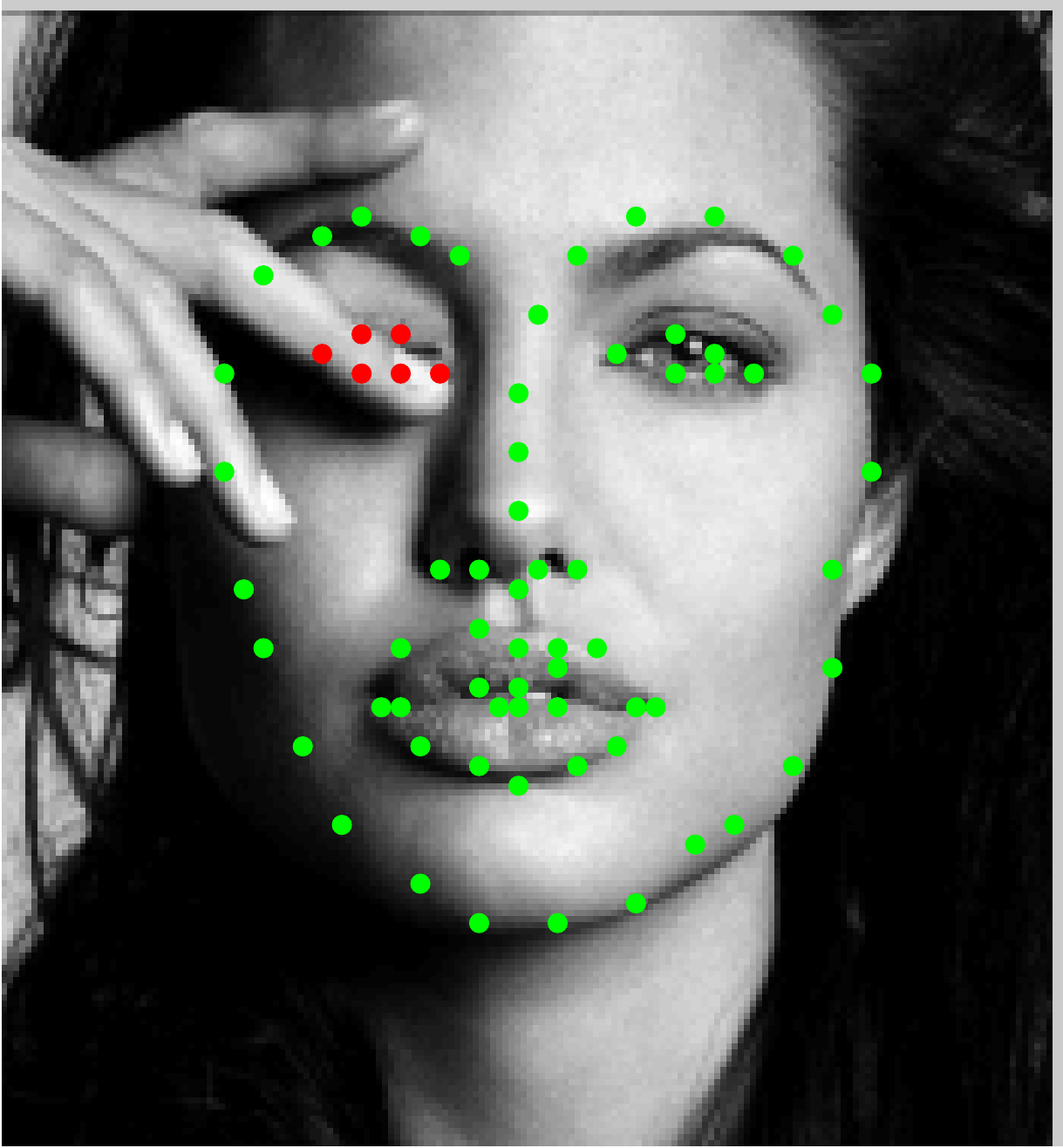} \hspace{-2pt}
		\includegraphics[height=0.16\textwidth]{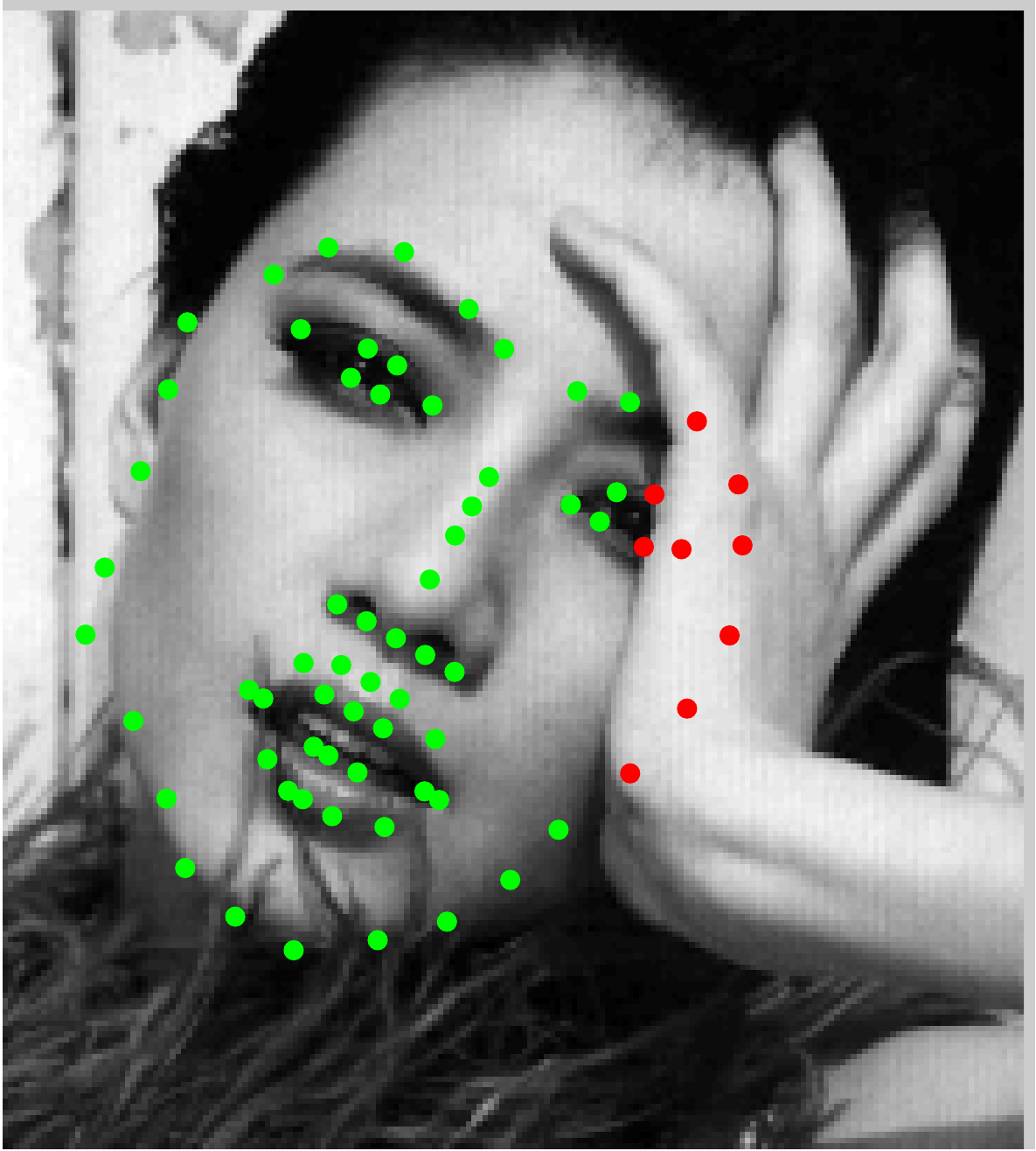}\\
		\includegraphics[height=0.16\textwidth]{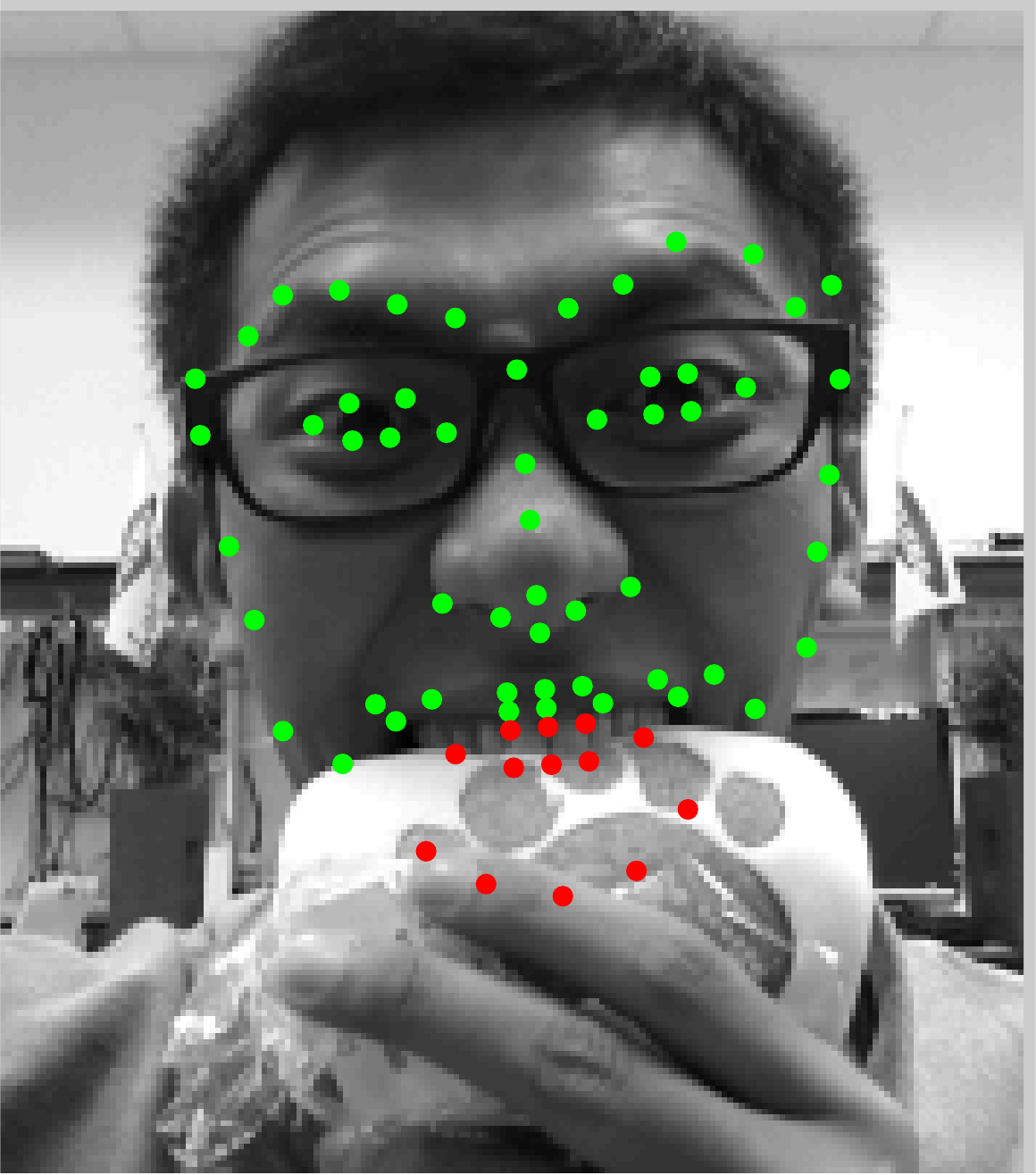} \hspace{-5pt}
		\includegraphics[height=0.16\textwidth]{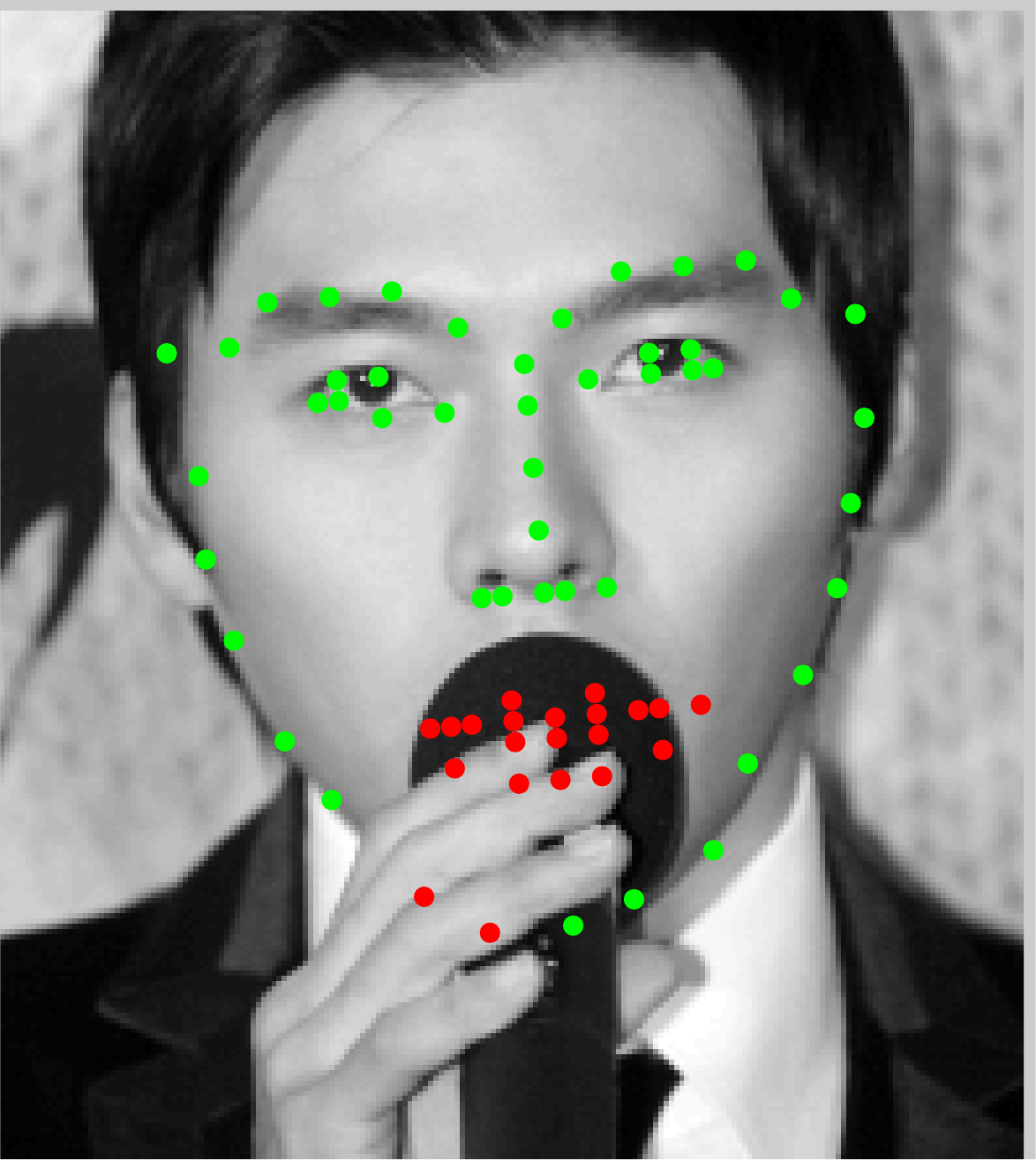} \hspace{-4pt}
		\includegraphics[height=0.16\textwidth]{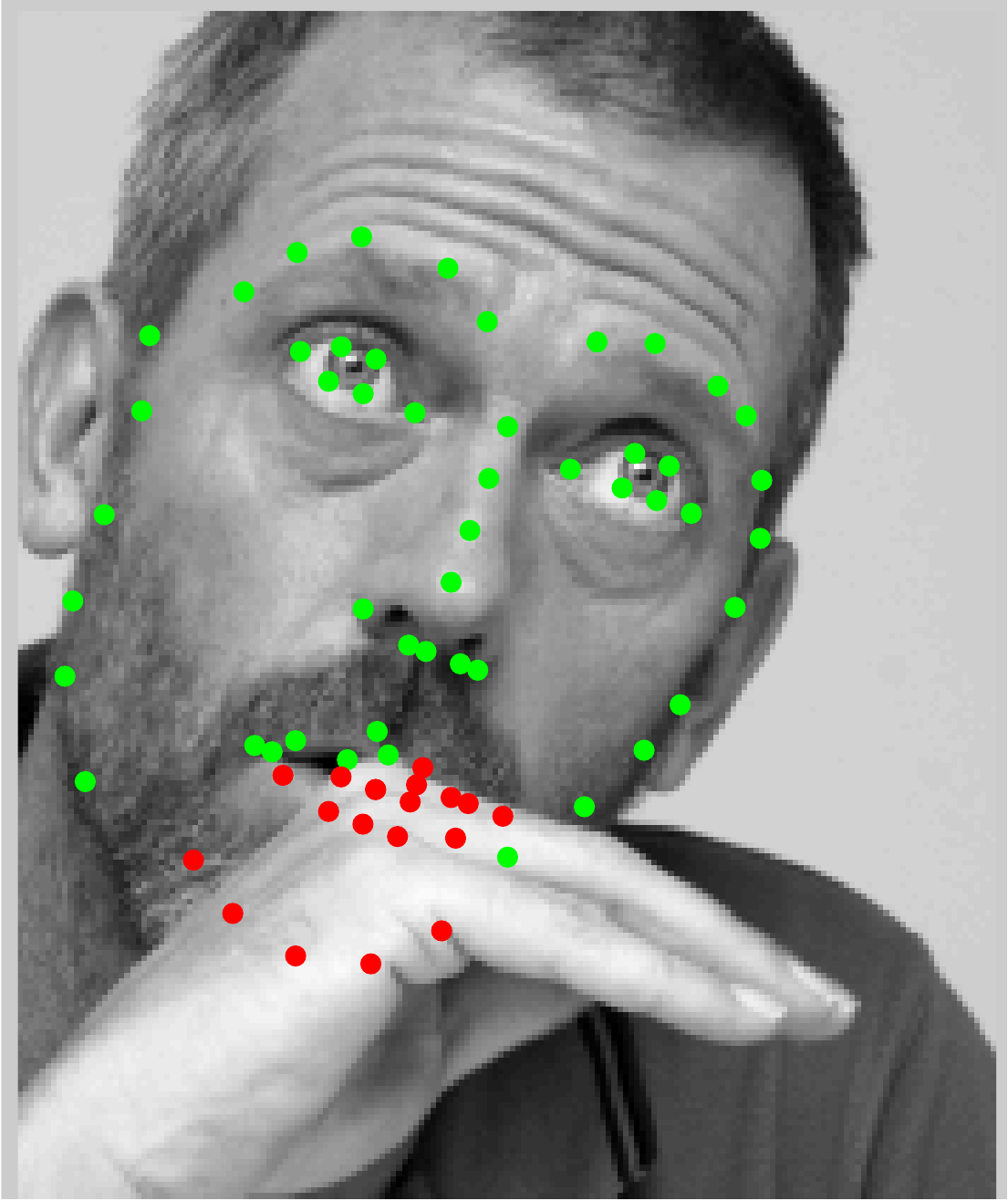} \hspace{-5pt}
		\includegraphics[height=0.16\textwidth]{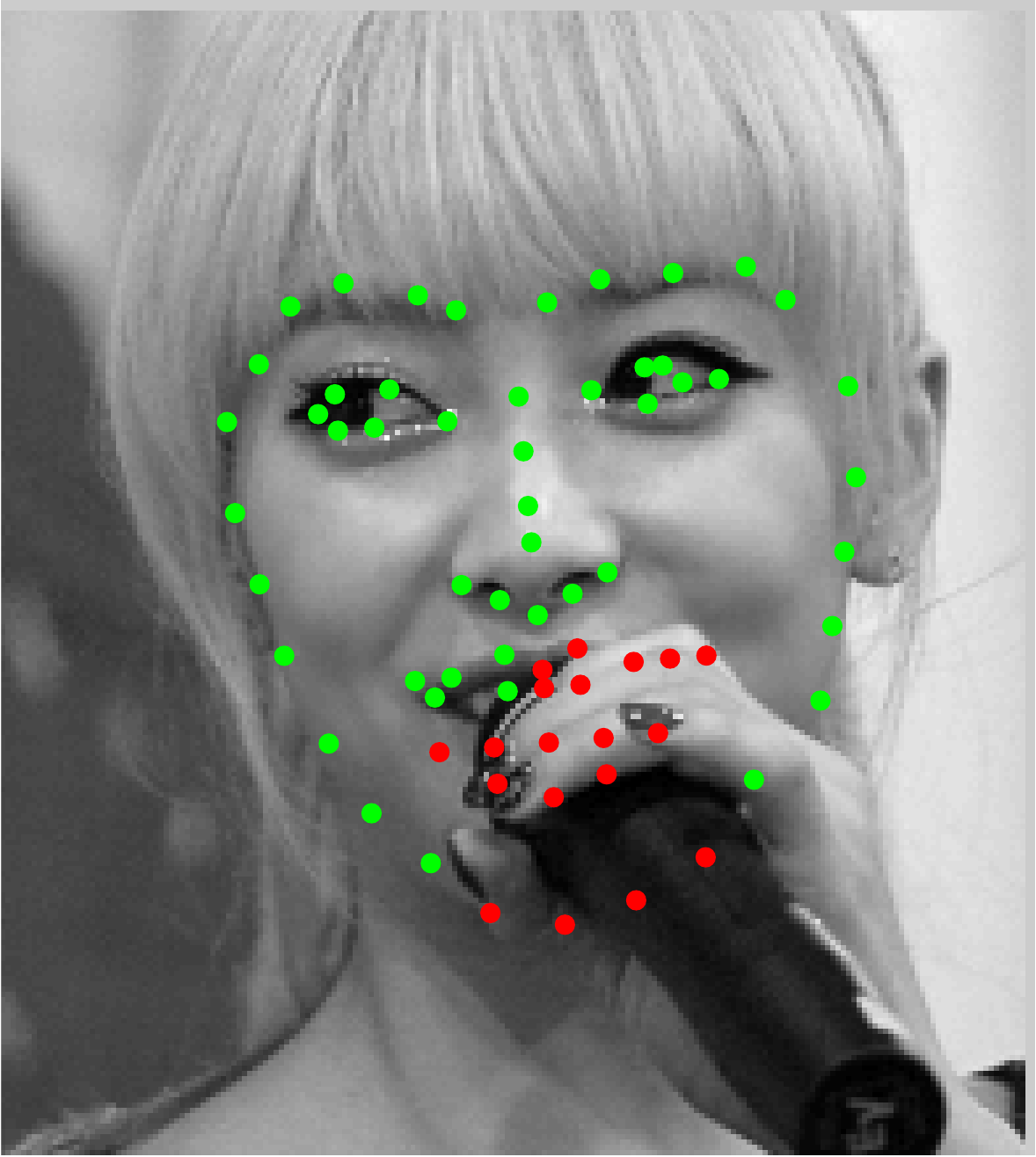} \hspace{-4pt}
		\includegraphics[height=0.16\textwidth]{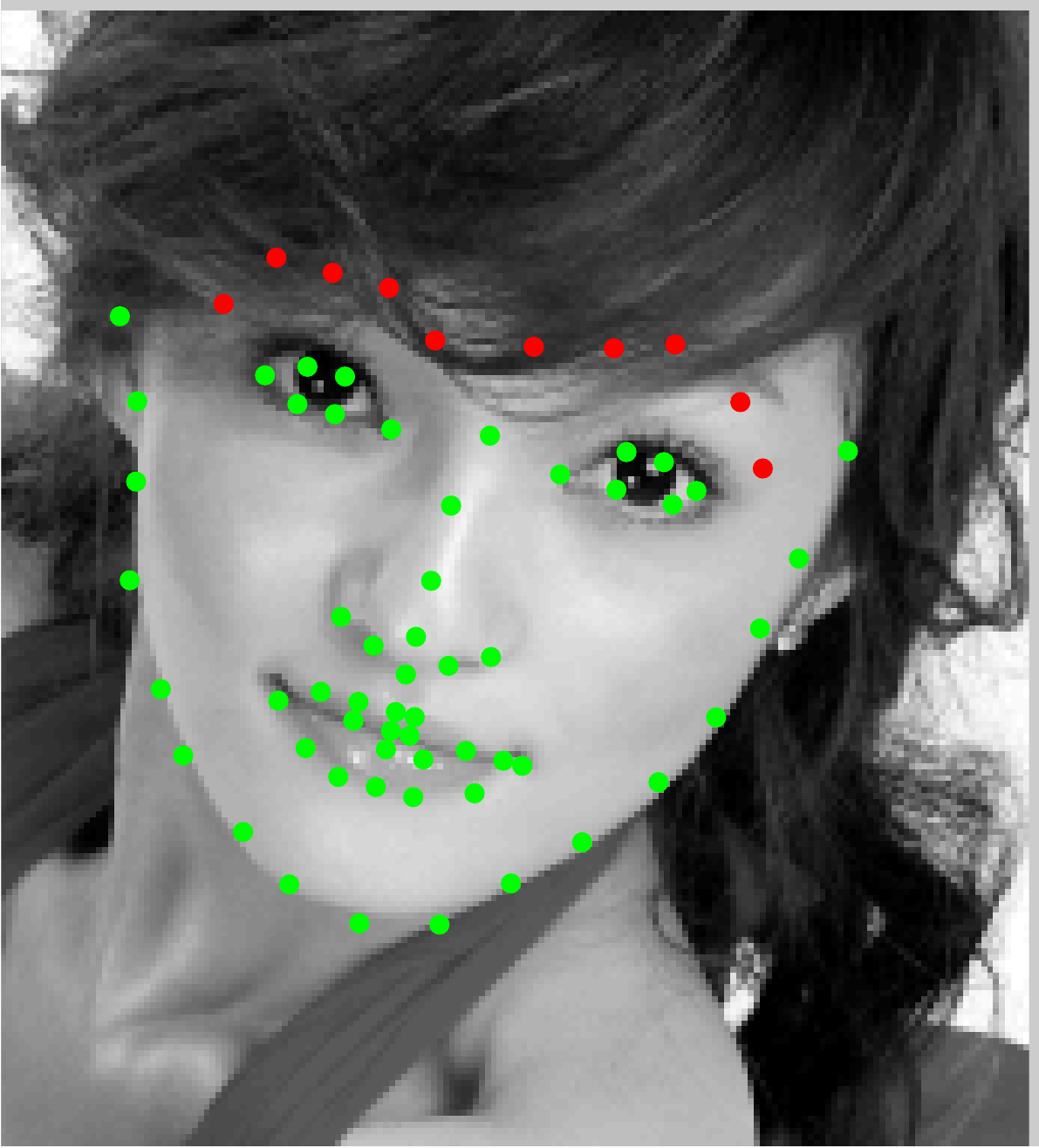} \hspace{-5pt}
		\includegraphics[height=0.16\textwidth]{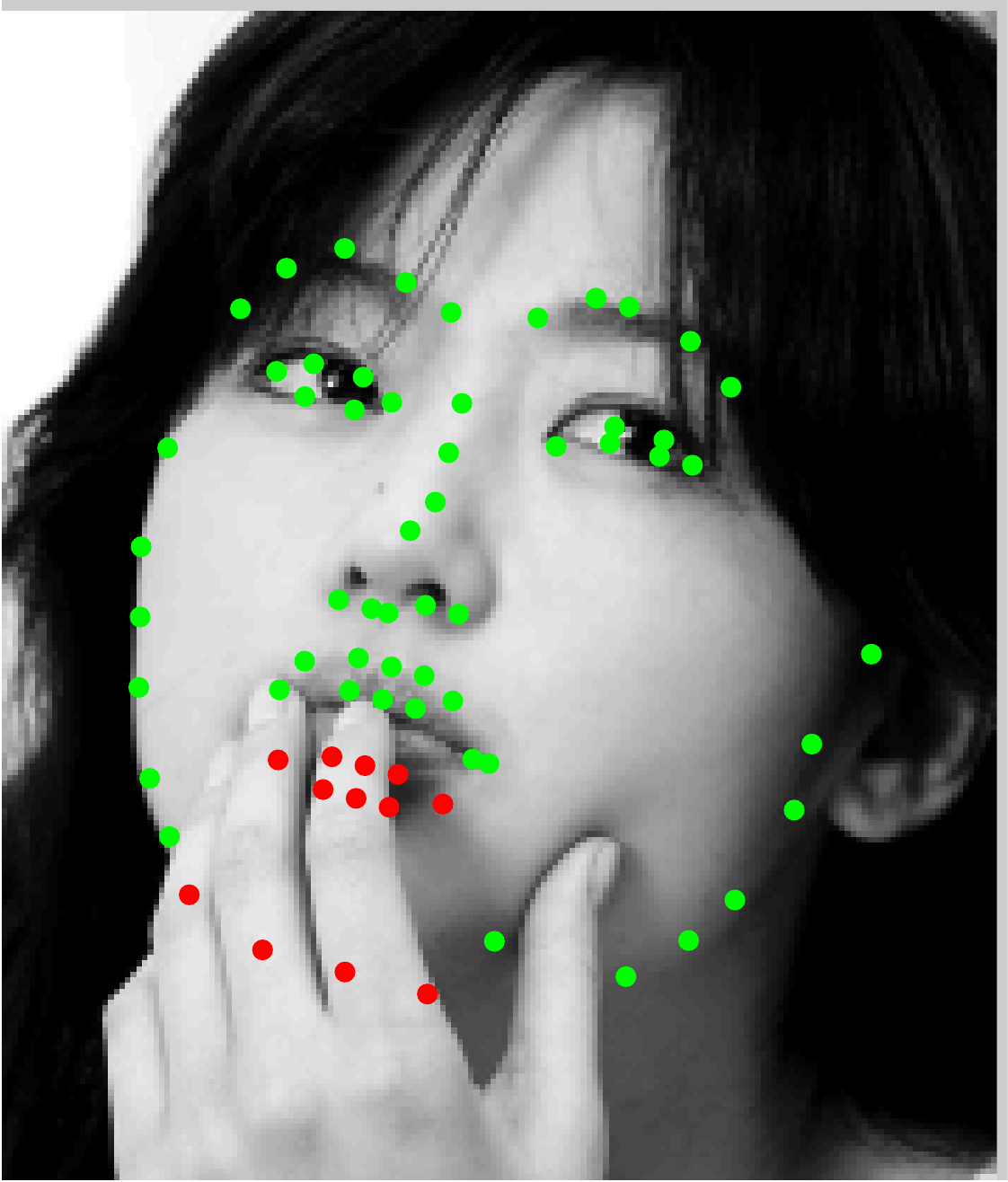} \hspace{-4pt}
		\includegraphics[height=0.16\textwidth]{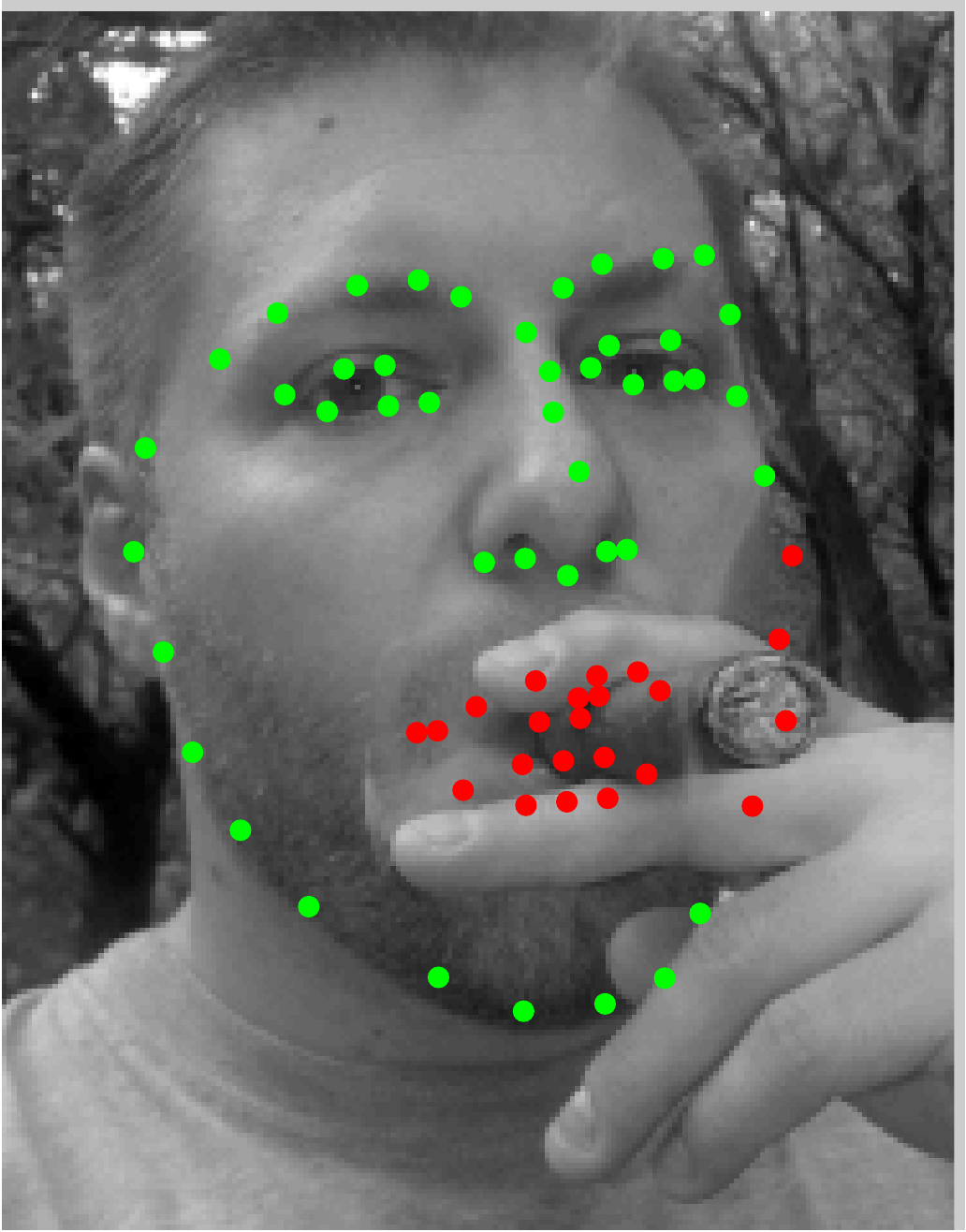}\\
\caption{Examples of landmark localization and occlusion estimation for
images from the HELEN (row 1) and COFW (rows 2-3) test datasets. Red
indicates those landmarks which are predicted as being occluded by the HPM.}
\label{fig:results1}
\end{figure*}

\subsection{Parameter learning}
\label{sec:param_learning}
Recall that our model (Eqn. \ref{eqn:score}) is parameterized by a set of
weights and biases, which we collect into a parameter vector $w$.  Each weight
is multiplied by some corresponding feature that depends on the hypothesized
model configuration $(l,s,o)$ and input image $I$.  Collecting these features a
feature vector $\Psi(l,s,o|I)$, we write the scoring function as an inner
product with the model weights $Q(l,s,o) = w \cdot \Psi(l,s,o|I)$.  We learn
the model weights using a regularized SVM objective:
\begin{align*}
\min_w &\frac{1}{2} \|w\|^2 + C \sum_t \eta_t & &\\
&  w \cdot \Psi(l^t,s^t,o^t|I^t) \geq 1 - \eta_t  & &\forall t \in \mathcal{P} \\
& w \cdot \Psi(l,s,o|I^t) \leq -(1 - m \delta(o) - \eta_t) & \forall l,s,o \; \; & \forall t \not\in \mathcal{P} 
\end{align*}
where $(l^t,s^t,o^t)$ denotes the supervised model configuration for a positive
training example, $\delta(o)$ is a margin scaling function that measures the
fraction of occluded landmarks and $C$ and $m$ are hyper-parameters (described
below).  The constraint on positive images $t \in \mathcal{P}$ encourages that
the score of the correct model configuration be larger than $1$ and penalizes
violations using slack variable $\eta_t$. The second constraint encourages the
score to be low on all negative training images $t \not\in \mathcal{P}$ for all
configurations of the latent variables.

\paragraph{Margin scaling for occlusion}
This formulation differs from standard supervised DPM training in the treatment
of negative training examples.  Since landmarks can be occluded in our model,
fully or partially occluded faces can be detected by our model in the negative
training images. These images do not contain any faces and we would like our model
generates low scores for these detections. However, a landmark which is detected as
occluded in a negative image is in some sense correct.  {\em There is no real
distinction between a negative image and a positive image of a fully occluded
face!}  Thus we penalize negative detections (false positives) with significant
amounts of occlusion less than fully-visible false positives.

For this purpose, we scale the margin for negative examples in proportion to
the number of occluded landmarks. We specify the margin for a negative example
as $1-m \delta(o)$, where the function $\delta(o)$ measures the fraction of
occluded landmarks and $m$ is a hyper-parameter. As the number of occluded
landmarks increases the margin decreases and the model score for that example
can be larger without violating the constraint. The margin for a fully occluded
example is equal to $1-m$. Setting $m=0$ corresponds to standard classification
where all the negatives have the same margin of $1$.  In this case the biases
learned for occluded landmarks tend to be low (otherwise many fully or
partially occluded negative examples will violate the constraint). As a result,
models trained with $m=0$ tend not to predict occlusion.  As we increase $m$,
the scores of fully or partially occluded negative examples can be larger
without violating the constraint and the training procedure is thus free to
learn larger bias parameters associated with occluded landmarks.  As we show in
our experimental evaluation, this results in higher recall of occluded
landmarks and improved test-time performance.

We use a standard hard-negative mining or cutting-plane approach to find a
small set of active constraints for each negative image.  Given a current
estimate of the model parameters $w$, we find the model configuration $(s,l,o)$
that maximizes $w \cdot \Psi(l,s,o|I^t)-m \delta(o)$ on a negative window
$I^t$.  Since the loss $m \delta(o)$ can be decomposed over individual
landmarks, this loss-augmented inference can be easily performed using the same
inference procedure introduced in section \ref{sec:model}.  We simply subtract
$\frac{m}{N_l}$ from the messages sent by occluded landmarks where $N_l=68$ is
the number of landmarks.  During training we make multiple passes through the
negative training set and maintain a pool of hard negatives for each image.  We
share the slack variable $\eta_t$ for all such negatives found over a single
window $I^t$.

\subsection{Test-time Prediction}

\paragraph{Scale and In-plane Rotation}
We use a standard sliding window approach to search over a range of locations
and scales in each test image.  In our experiments, we observed that part
models with standard quadratic spring costs are surprisingly sensitive to
in-plane rotation.  Models that performed well on images with controlled
acquisition (such as MultiPIE) fared poorly ``in the wild'' when faces were
tilted.  The alignment procedure described above explicitly removes scale and
in-plane rotations from the set of training examples.  At test time, we perform
an explicit search over in-plane rotations (-30 to 30 degrees with an increment
of 6 degrees).

\paragraph{Landmark Prediction}
The number of landmarks in our model was chosen based on the availability of
68-landmark ground-truth annotations.  In cases where it was useful to
benchmark landmark localization of our model on datasets using different
landmark annotation standards (e.g., COFW 29-landmark data), we used additional
held-out training data to fit a simple linear map from the part locations
returned by our hierarchical part model to the desired output space.  This
provided a more stable procedure than simpler heuristics such as hand
selecting a subset of landmarks.

Let $l^i \in {\mathbb R}^{2N_l}$ be the vector of landmark locations returned
at the top scoring detection when running the model on a training example $i$.
Let ${\hat l^i} \in {\mathbb R}^{2M}$ a vector of ground-truth landmark
locations for that image based on some other annotation standard (i.e., $M
\not= N_l$).  We train a linear regressor
\[
\min_{\beta} \sum_i \|{\hat l}^i - \beta^T l^i\|^2 + \lambda \|\beta\|^2
\]
where $\beta \in {\mathbb R}^{2N_l\times2M}$ is the matrix of learned
coefficients and $\lambda$ is a regularization parameter.  To prevent
overfitting, we restrict $\beta_{pq}$ to be zero unless the landmark $p$
belongs to the same part as $q$.

To predict landmark occlusion, we carried out a similar mapping procedure using
regularized logistic regression.  However, in this case we found that simply
specifying a fixed correspondence between the two sets of landmarks based on
their average locations and transferring the occlusion flag from the model to
benchmark landmark space achieved the same accuracy. 

\section{Experimental Evaluation}

Figure \ref{fig:results1} shows example outputs of the HPM model run on example
face images. The model produces both a detection score and estimates of
landmark locations and occlusion states.  While the possible occlusion patterns
are quite limited (4 occlusions patterns per part shape), the final predicted
occlusions (marked in red) are quite satisfying in highlighting the support of
the occluder for many instances.  We evaluate the performance of the model on
three different tasks: landmark localization, landmark occlusion prediction,
and face detection.  In our experiments we focus on test datasets that have
significant amounts of occlusion and emphasize the ability of the model to
generalize well across datasets.

\subsection{Landmark Localization}
\label{sec:landmarklocalization}

\begin{figure*}[ht!]
\begin{center}
\begin{tabular}{ccc}
\includegraphics[width=0.65\columnwidth]{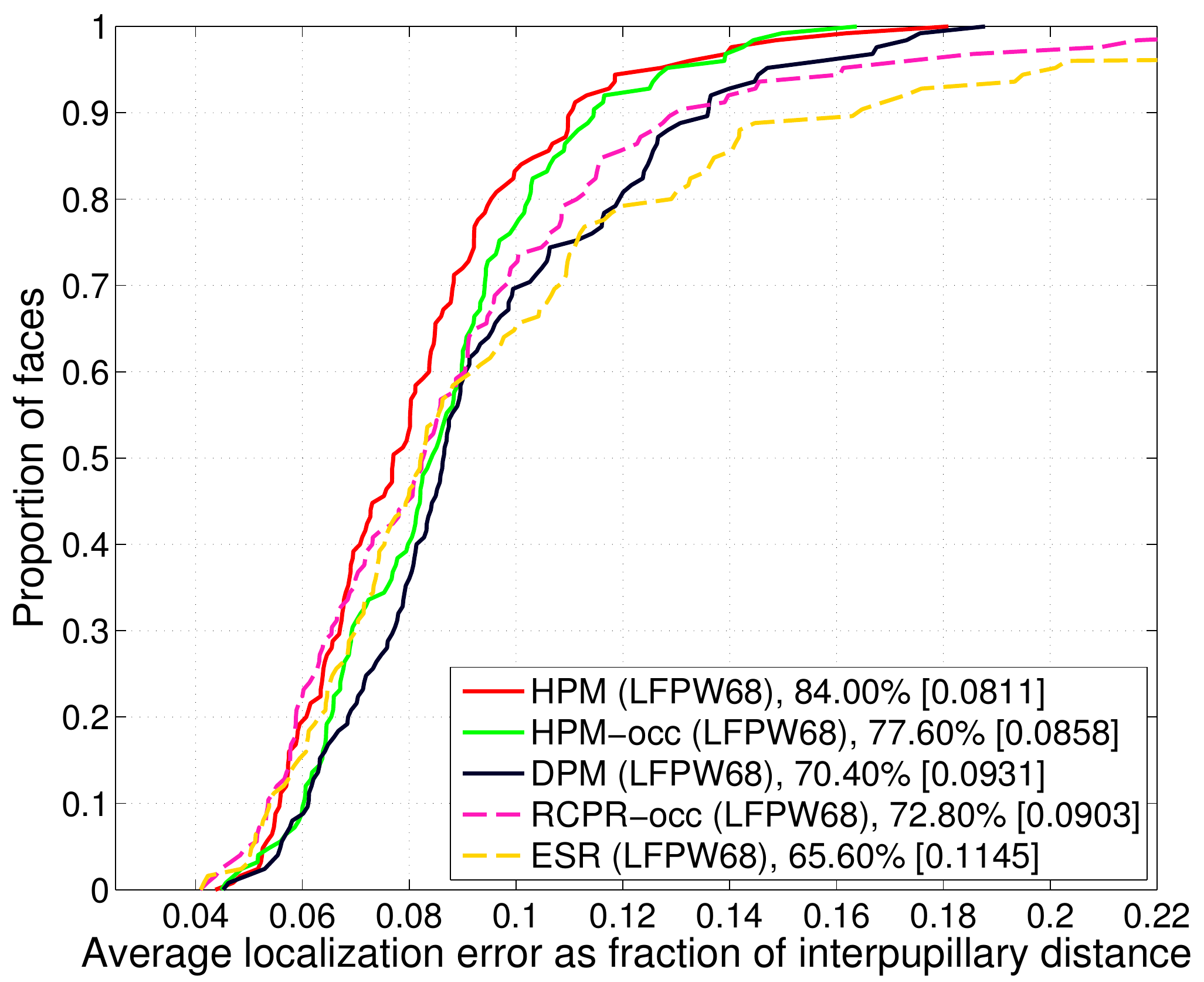}&
\includegraphics[width=0.65\columnwidth]{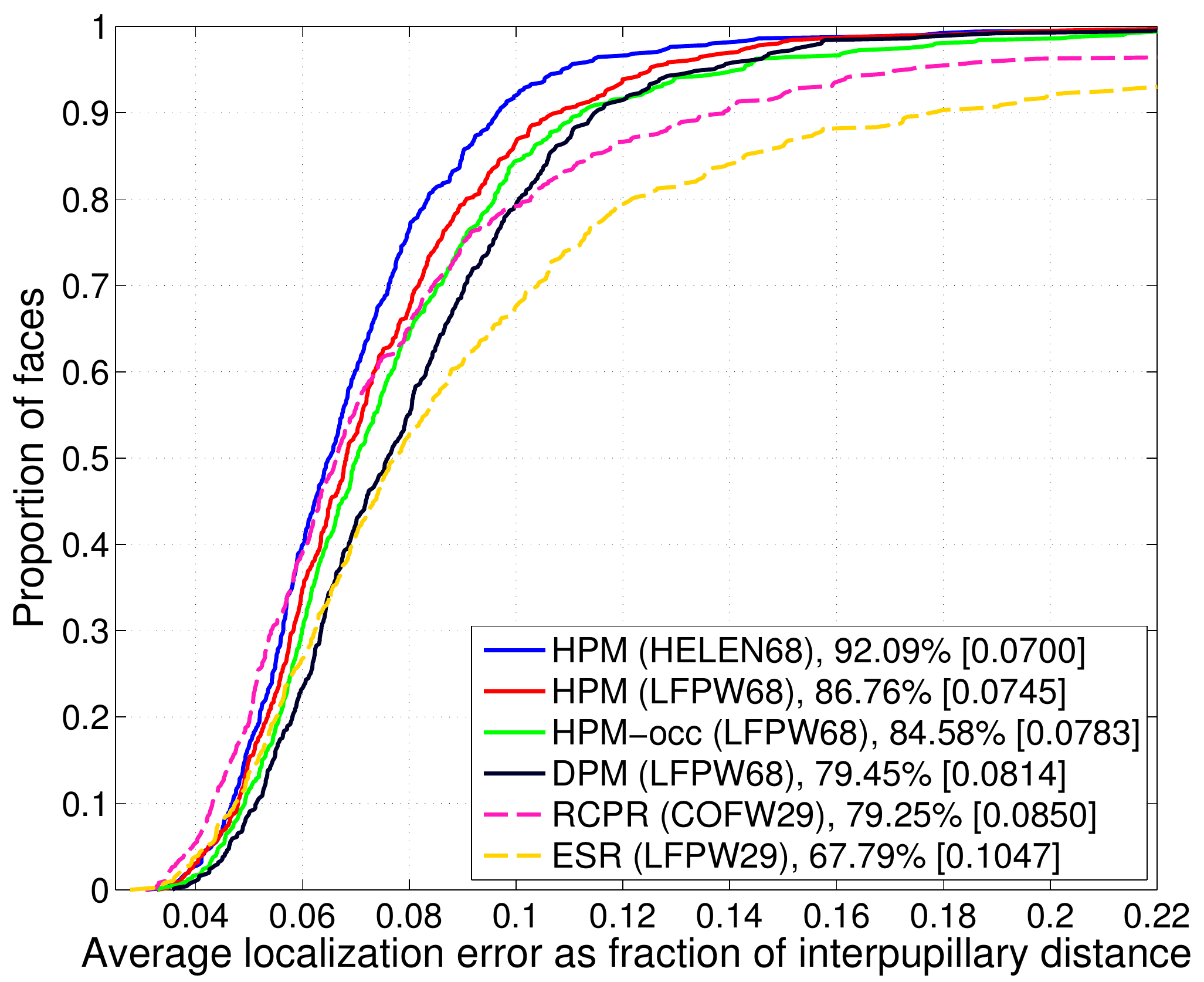}&
\includegraphics[width=0.32\linewidth]{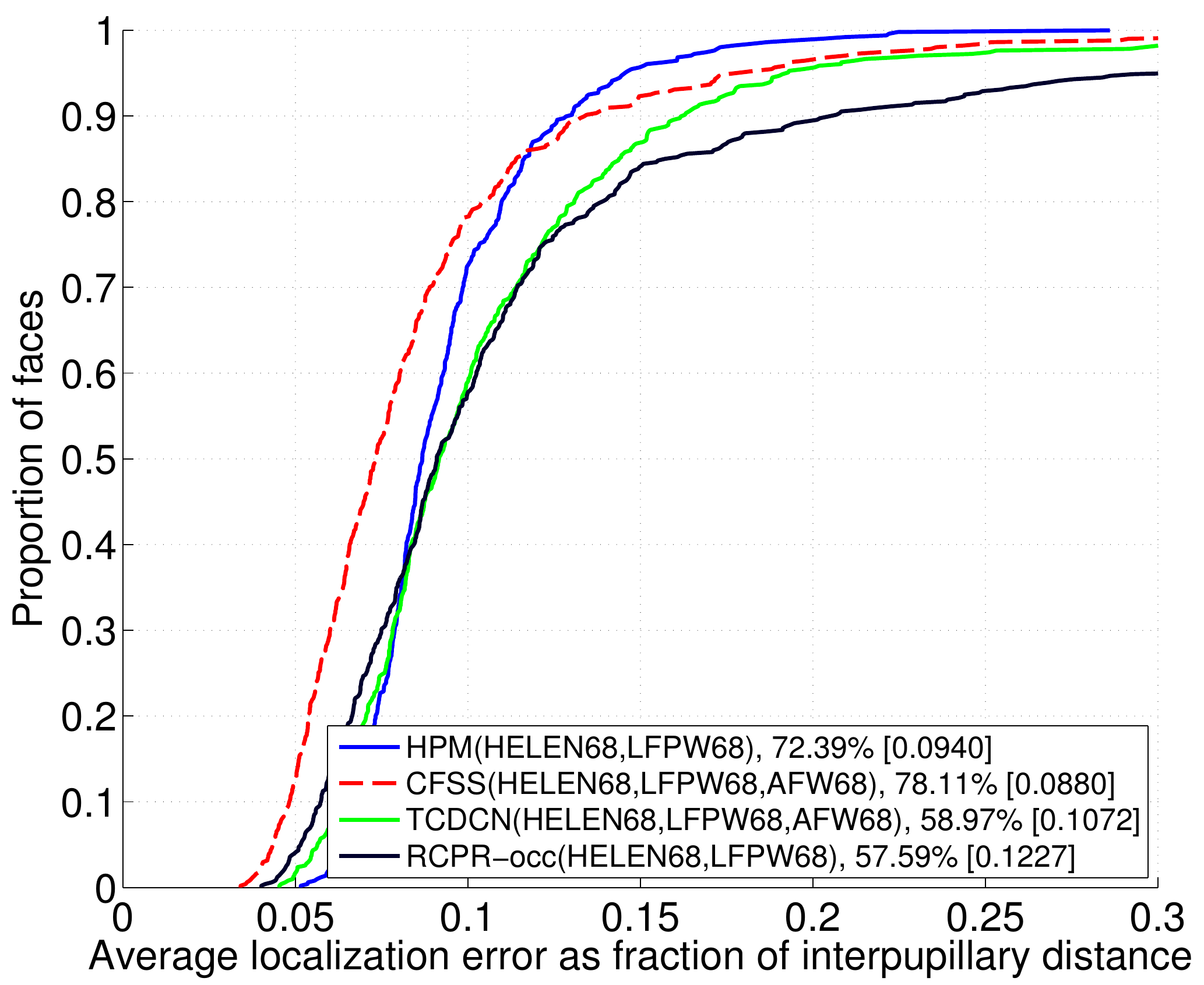}\\
(a) Occluded HELEN68 & (b) COFW29 & (c) COFW68\\
\end{tabular}
\end{center}
\caption{
Panels show cumulative error distribution curves (the proportion of test images
that have average landmark localization error below a given threshold) on three
test sets: an occlusion rich subset of HELEN, COFW29 and COFW68.  The legend
indicates the training set (in parentheses), the success rate \% at a
localization threshold of $0.1$ and the average error [in brackets].  The HPM
shows good localization performance these difficult datasets with significant
occlusion.  In general regression models (dashed lines) have better performance
for a low localization threshold compare to part based models (solid lines).
However, the success rates for regression models increase more slowly and
eventually cross over those for part models (solid lines) as the allowable
localization error threshold increases.
}
\label{fig:curves}
\end{figure*}

\paragraph{Datasets}
We evaluate performance of our method and related baselines on three benchmark
datasets for landmark localization: the challenging portion of the IBUG dataset
which contains a range of poses and expressions ~\cite{300w}, a subset of the
HELEN dataset~\cite{le2012interactive} containing occlusions, and the Caltech
Occluded faces in the Wild (COFW)~\cite{burgos2013robust} dataset.  We evaluate
on IBUG to provide a baseline for localization in the absence of occlusion.
The latter two datasets were selected to evaluate the ability of our model in
the presence of substantial natural occlusion which is not well represented
in many benchmarks.  The authors of \cite{burgos2013robust} estimate that COFW
contains $23\%$ occluded landmarks.  Fig.~\ref{fig:results1} depicts selected
results of running our detector on example images from the HELEN and COFW test
datasets.

\paragraph{68 Landmark annotations for COFW}
We note there is a variety of annotation conventions across different face
landmark datasets. COFW is annotated with 29 landmarks while HELEN includes a
much denser set of 194 landmarks. The 300 Faces in-the-wild Challenge
(300-W)~\cite{300w} has produced several unified benchmarks in which HELEN
dataset have been re-annotated with a set of 68 standard landmarks. To 
allow for a greater range of comparisons and further this standardization, we
manually re-annotated the test images from the COFW dataset with 68 landmarks
and occlusion flags.  We also generated face bounding boxes (using a similar
detection method that used for the 300-W datasets \cite{asthana2013robust})
for evaluating pose regression methods that require initialization. We
bootstrapped our annotations from the 29-landmark annotations using a custom
annotation tool. The annotations and benchmarking code are publicly
available\footnote{\url{https://github.com/golnazghiasi/cofw68-benchmark}}.

\paragraph{Localization Evaluation Metrics}
To evaluate landmark localization independent of detection accuracy, we follow
a standard approach that assumes that detection has already been performed and
evaluates performance on cropped versions of test images.  While our model is
capable of both detecting and localizing landmarks, this protocol is necessary
to evaluate pose regression methods that require good initialization.
We thus follow the standard protocol (see e.g., \cite{300w}) of using the
bounding boxes provided for each dataset (usually generated from the output of a
face detector) by evaluating the localization accuracy for the highest scoring
detection that overlaps the given bounding box by at least 70\%.

We report the average landmark localization error across each test set as well
as the ``success rate'', the proportion of test images with average landmark
localization error below a given threshold.  Distances used in both quantities
are expressed as a proportion of the interpupillary distance (distance between
centers of eyes) specified by the ground-truth.  Computing the success rate
across a range thresholds yields a cumulative error distribution curve (CED)
(Fig. \ref{fig:curves}).  When a single summary number is desired, we report
the success rate at a standard threshold of $0.1$ interpupillary distance
(IPD).

\paragraph{Training and baselines} 
To train our model, we used training data from LFPW (811 images) and/or HELEN
(2000 images) annotated with 68 landmarks. The training set is specified in
parenthesis in figure legends. From each training image we generate 8
synthetically occluded ``virtual positives''.  To fit linear regression
coefficients for mapping from the HPM predicted landmark locations to 29
landmark datasets, we ran the trained model on the COFW training data set and
fit regression parameters $\beta$ that mapped from the 68 predicted points to
the 29 annotated.

For diagnostic purposes, we trained several baseline models including a version
of our model without occlusion mixtures (HPM-occ) and the (non-hierarchical)
deformable part model \footnote{The originally published DPM model
of~\cite{zhu2012face} was trained on the very constrained MultiPIE
dataset~\cite{gross2010multi}.  Retraining the model of Zhu et al. and
including in-plane rotation search at test time yielded significantly better
performance than reported elsewhere (c.f., \cite{burgos2013robust})} (DPM)
described by~\cite{zhu2012face}.  We also evaluated variants of the robust
cascaded pose regression (RCPR) described in~\cite{burgos2013robust} as well as
their implementation of explicit shape regression (ESR) ~\cite{cao2012face}
using both pre-trained models provided by the authors and models retrained to
predict 68 landmarks.  Unlike HPM which uses virtual occlusion, RCPR requires
training examples with actual occlusions and corresponding annotations.  For
training sets that featured no occlusion, we thus trained a variant that does
not model occlusion (RCPR-occ).

\begin{table}[t]
\begin{center}
{\footnotesize
\begin{tabular}{l|c|c}
Method                                   &average error\\
\hline\hline
DRMF~\cite{asthana2013robust}            &0.1979\\
CDM~\cite{yu2013pose}                    &0.1954\\
RCPR~\cite{burgos2013robust}             &0.1726\\
ESR~\cite{cao2012face}                   &0.1700\\
CFAN~\cite{zhang2014coarse}              &0.1678\\
SDM~\cite{xiong2013supervised}           &0.1540\\
CFSS~\cite{zhu2015face}        &0.1200$\ddagger$ / 0.0998\\
TCDCN~\cite{zhang2016learning}  &0.1121$\dagger$ / 0.0860\\
LBF~\cite{ren2014face}                   &0.1198\\
\hline
HPM                                      &0.1310\\
\hline
\end{tabular}
}
\end{center}
\caption{
Average errors as a fraction of IPD on IBUG68 ~\cite{300w} dataset.
Results with $\ddagger$/$\dagger$ are obtained by testing the method with the
standard detector bounding boxes provided by 300-W, using either the 
published model ($\dagger$) or retraining ($\ddagger$).
}
\label{table:ibug}
\end{table}

\begin{figure}[t]
\begin{center}
\begin{tabular}{cc}
\includegraphics[width=0.8\columnwidth]{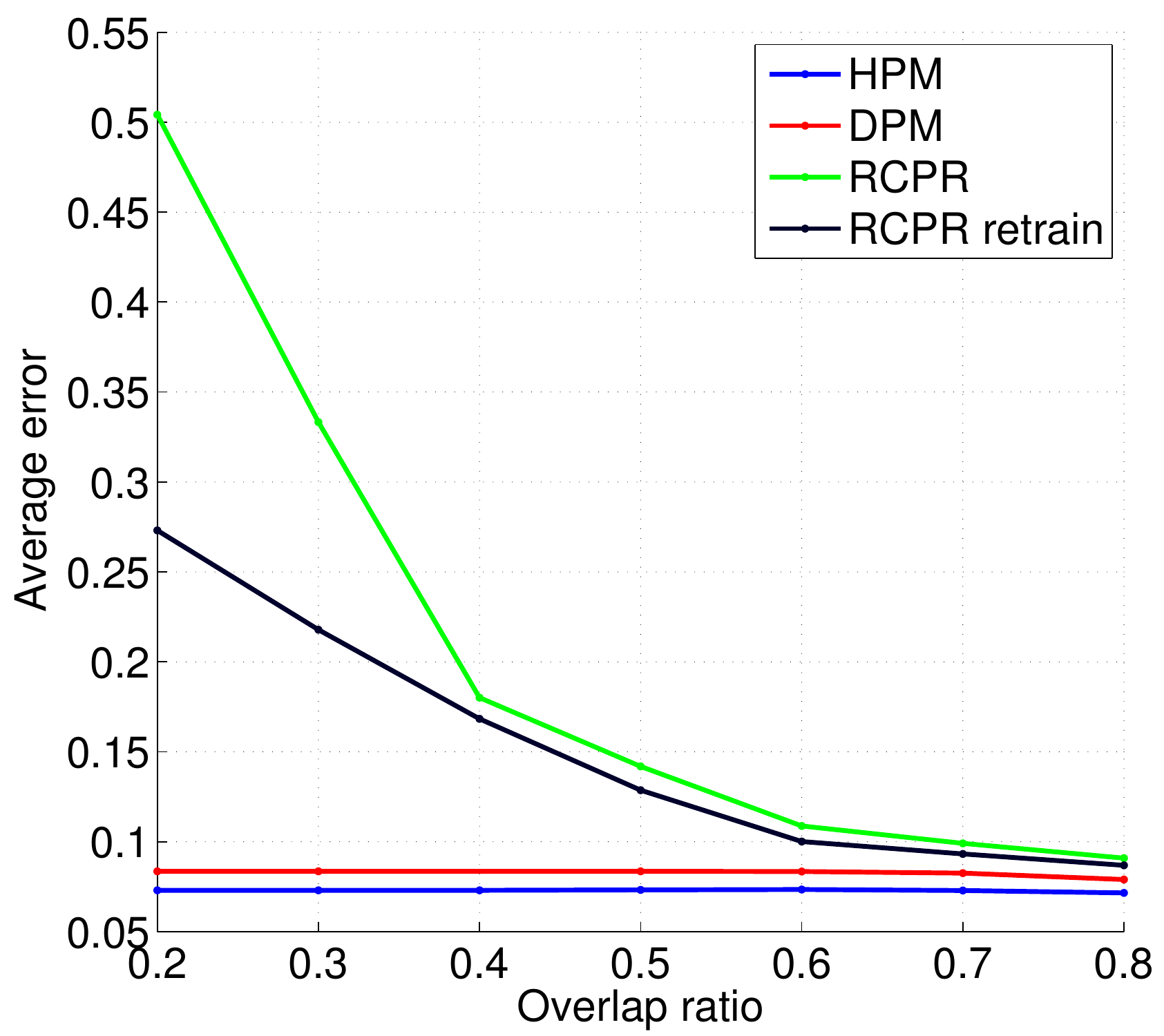}\\
\end{tabular}
\end{center}
\caption{We analyze the landmark localization average error of RCPR, HPM and
DPM for different overlap ratio with the ground-truth face boxes. For RCPR we change the minimum
overlap ratio of the initial bounding boxes and the ground-truth face boxes.
For HPM and DPM, we change the minimum overlap threshold of the returned detections and ground-truth
boxes. RCPR is very sensitive to the amount of overlap and its performance decreases rapidly 
as the overlap ratio decreases. But, HPM and DPM are robust to the overlap threshold and they 
can maintain the same performance over different thresholds.}
\label{fig:overlap}
\end{figure}

\begin{table}[t]
\begin{center}
{\footnotesize
\begin{tabular}{l|l||r|r||r|r||r|r||}
\multicolumn{2}{c||}{}      & \multicolumn{2}{|c||}{LFPW (29)} & \multicolumn{2}{|c||}{COFW (29)} \\ \hline
model &    training dataset    & SR &AE&  SR &AE\\ \hline \hline
RCPR-occ&LFPW29        &88.95 &0.073 &    63.44 &0.115 \\ \hline
RCPR-occ&LFPW29+       &98.95 &0.038 &    63.64 &0.096 \\ \hline
RCPR-occ&COFW29        &89.01 &0.071 &    76.28 &0.091 \\ \hline
RCPR &COFW29           &91.05 &0.064 &    79.25 &0.085 \\ \hline \hline
HPM&LFPW68,INR-        &97.37 &0.050 &    86.76 &0.075 \\ \hline
HPM&HELEN68,INR-       &98.42 &0.049 &    90.71 &0.072 \\ \hline
HPM&HELEN68,PAS-       &98.95 &0.048 &    92.09 &0.070 \\ \hline
\end{tabular}
}
\end{center}
\caption{ We find HPM generalizes well across datasets while pose regression
has a strong dependence on training data. Localization performance is 
measured by success rate (SR) and average error (AE).  The RCPR model trained
on COFW performs much better on COFW test data compared to RCPR-occ trained on
LFPW29+ (79\% SR vs 64\% SR) but has much worse performance on LFPW test data
compared to that model (91\% SR vs 99\% SR). Good performance on LFPW also
depends heavily on including additional warped positive instances (LFPW29+ vs
LFPW29).  The HPM trained on LFPW68 has high success rates on both COFW (87\%)
and LFPW (\%97) test data.  Last two rows of the table show the performance of
HPM when a different training data set (HELEN68) is used for training. This
dataset has more variation and more images (1758) compared to LFPW68 (682) and
improves performance of HPM on both test datasets.  Training on more negative
images (6000 images from PASCAL) decreases localization error of our model
compared to using only INRIA negatives.
}
\label{table:generalization}
\end{table}

\paragraph{Localization Results (Occluded HELEN 68)}
We evaluated on a subset of the HELEN dataset~\cite{le2012interactive}
consisting of 126 images which were selected on the basis having some
significant amount of occlusion
\footnote{\url{https://github.com/golnazghiasi/Occluded-HELEN-image-list}}.
We do not report results of the HPM (HELEN68)
model on this dataset as there was overlap between training and testing images.
Fig.~\ref{fig:curves}(a) shows the error distribution.  The HPM achieves an
average error of 0.0811, beating out the DPM baseline (0.0931) and RCPR-occ
(0.0903). Removing explicit occlusion from the model (HPM-occ) results in lower
success rates for a range of thresholds.

\paragraph{Localization Results (COFW29)}
To facilitate diagnostic comparison to previously published results, we
evaluated our model on the original COFW 29-landmark test set
~\cite{burgos2013robust} consisting of 507 internet photos depicting a wide
variety of more difficult poses and includes a significant amount of occlusion.
Since COFW training only contains 29 landmarks (we only performed additional
annotations on test data), we evaluated models trained on LFPW68 and HELEN68.
Fig.~\ref{fig:curves}(c) shows that HPM achieves a significantly lower average
error than RCPR and higher success rates for all but the smallest $(<0.06)$
localization success thresholds.

\paragraph{Localization Results (COFW68)}
We tested our model trained on LPFW68 and HELEN68 training data on this
benchmark and compared with CFSS, TCDCN and RCPR-occ (Fig. \ref{fig:curves}
(c)).  For CFSS and TCDCN we used the publicly available pre-trained models
which were trained on HELEN68, LFPW68 and AFW68 (TCDCN is also pretrained on
MAFL dataset). For RCPR-occ we used the authors' code to train a model on
HELEN68 and LFPW68 training sets. Note we that couldn't train the full RCPR
68-landmark model with occlusion since HELEN68 and LFPW68 do not have occlusion
and COFW train is only labeled with 29 landmarks.

\begin{figure*}[ht]
\begin{center}
\begin{tabular}{ccc}
\includegraphics[width=0.65\columnwidth]{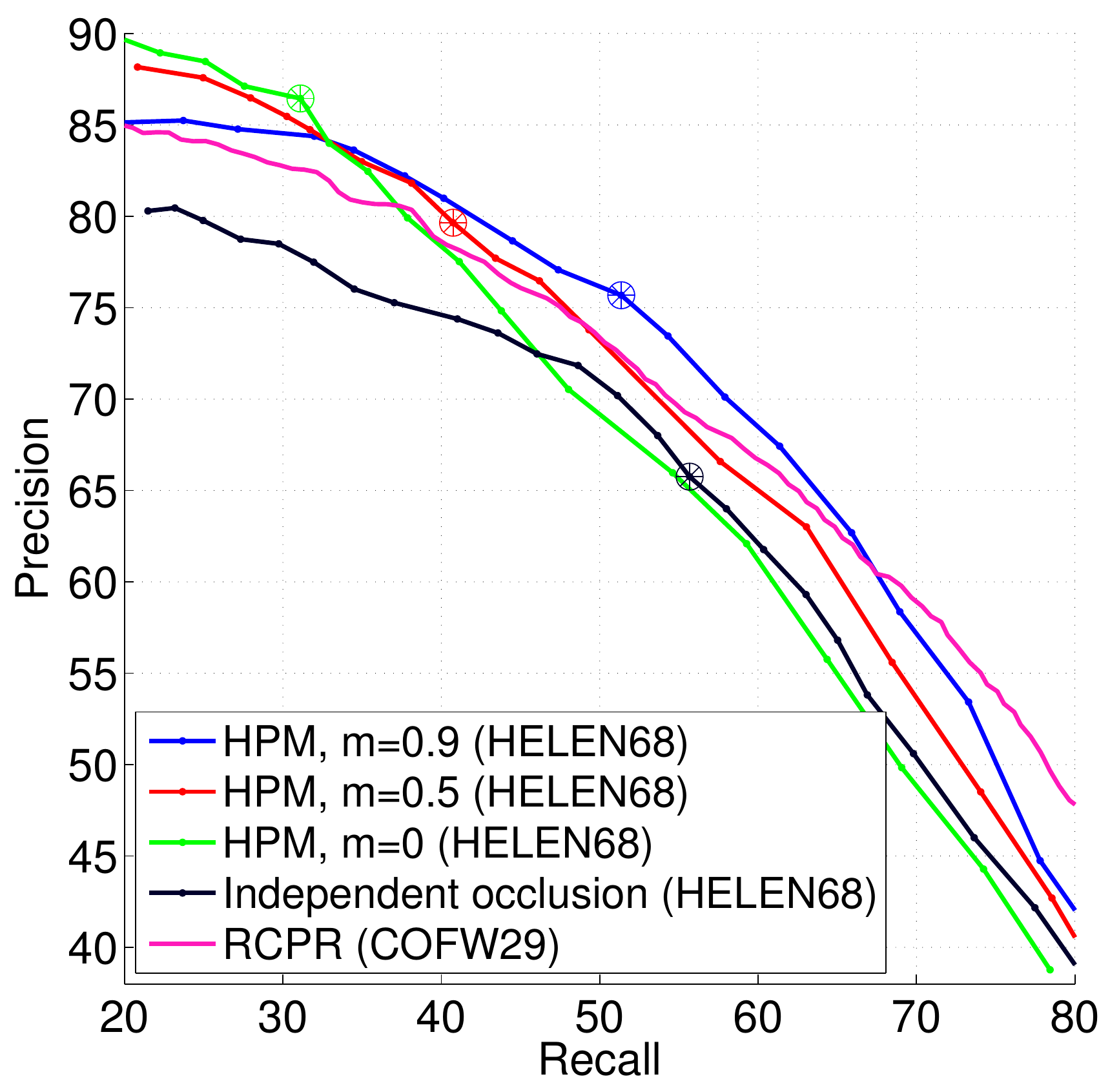}&
\includegraphics[width=0.65\columnwidth]{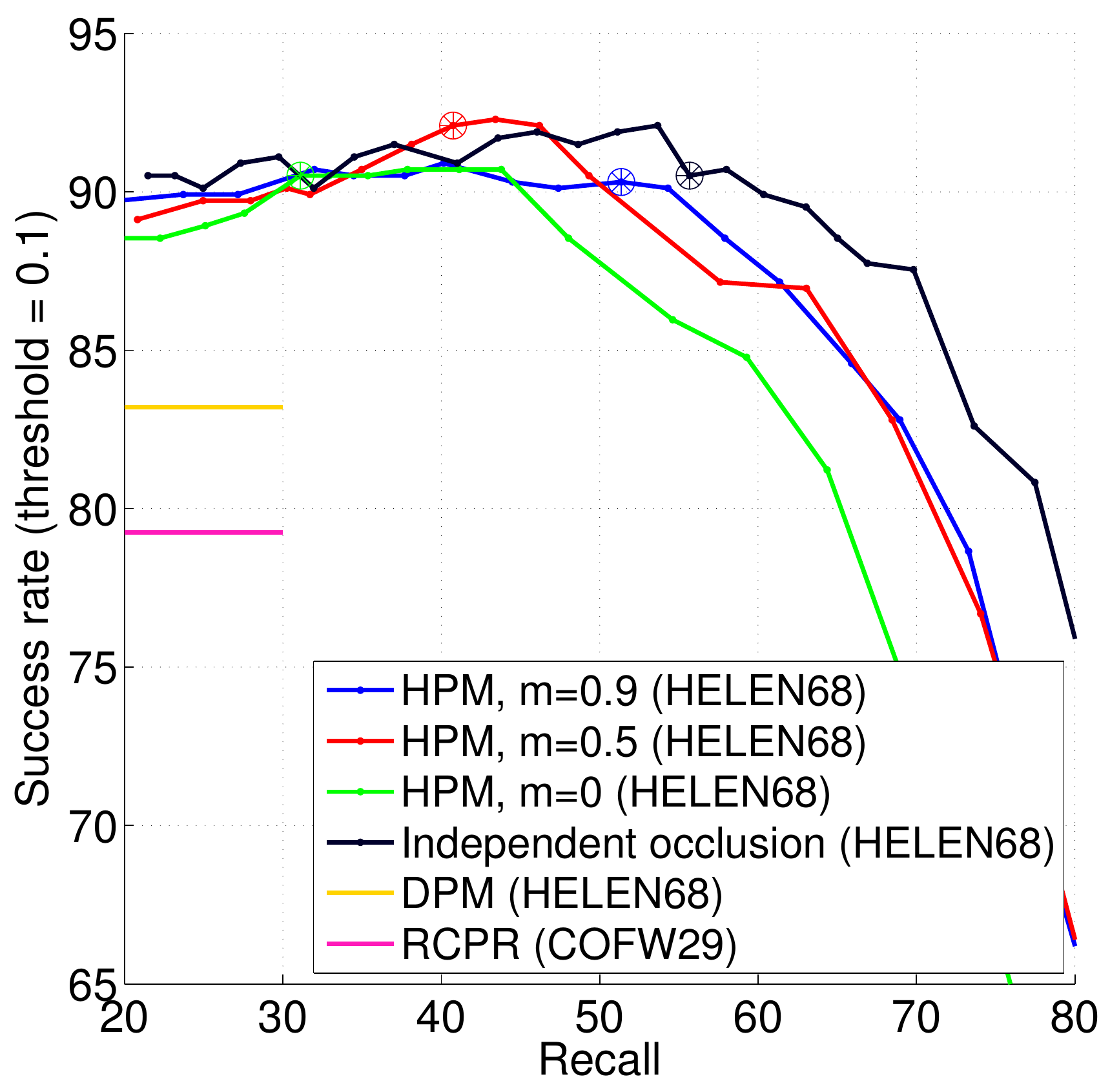}&
\includegraphics[width=0.65\columnwidth]{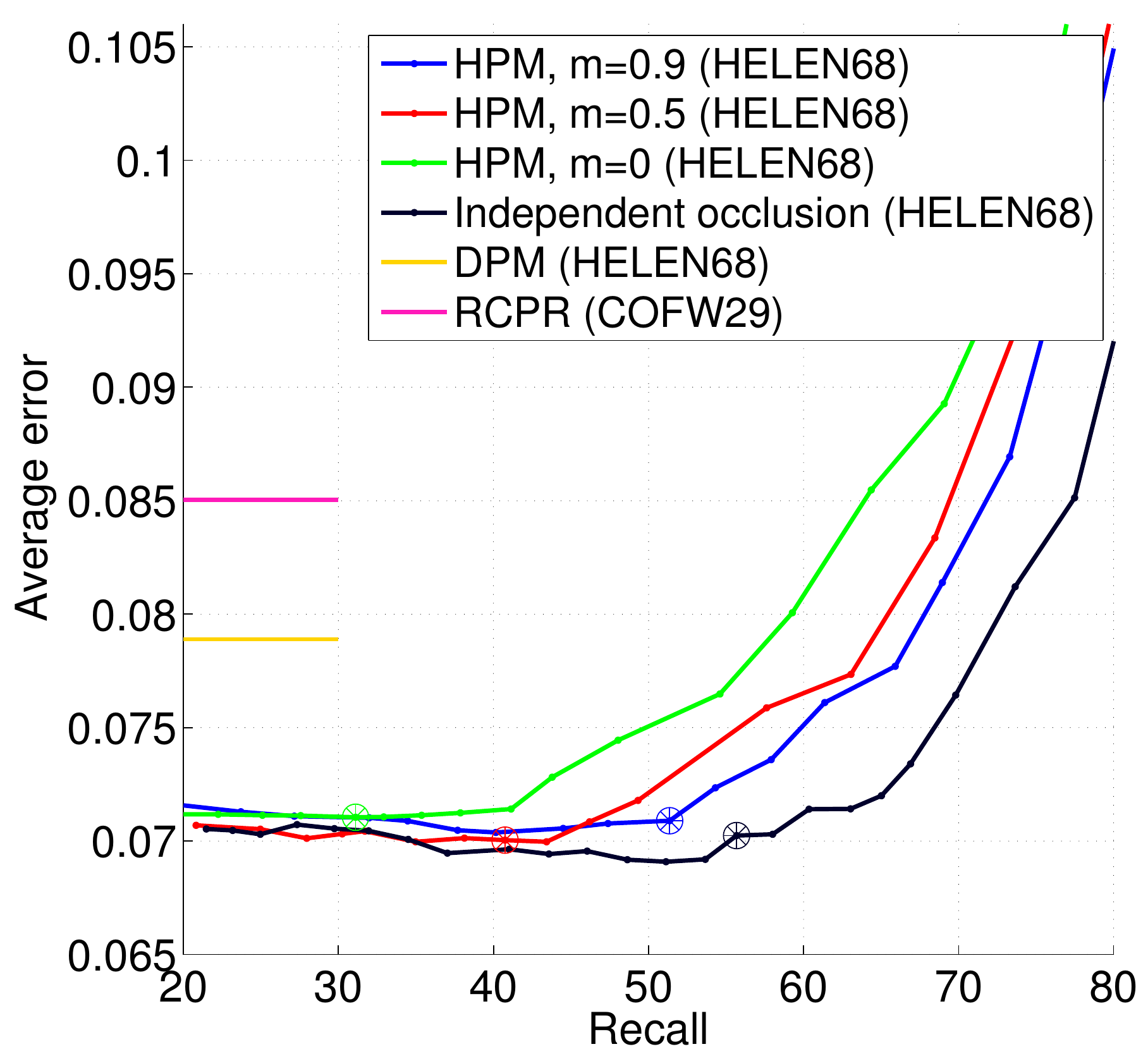}\\
(a) Occlusion prediction accuracy
& (b) Success rate vs. occlusion recall &
(c) Localization error vs. occlusions recall \\
\end{tabular}
\end{center}
\caption{Occlusion prediction accuracy on the COFW test dataset for variants of
our model.  Using a suitable margin scaling function (see Sec.
\ref{sec:param_learning}) allows for significantly better occlusion prediction
accuracy (a) over an independent occlusion model (a) with minimal loss in
localization performance (b,c).  Localization performance of DPM and RCPR are
included for reference.
}
\label{fig:margin_scaling}
\end{figure*}

\paragraph{Localization Results (IBUG 68)}
This dataset contains 68 landmark annotations for 135 faces in difficult poses
and expression \cite{300w}.  For testing our method on this dataset, we follow
previous work and trained our model on combined HELEN68
and LFPW68 training data provided by 300-W.  Since IBUG includes many side
view faces we trained a variant of our model with 7 viewpoints. We compare our
model with published performance of several state-of-the-art methods in Table
\ref{table:ibug} and achieve comparable performance.

In addition to reporting values from the published literature, we also
re-evaluated two recent top-performing models: TCDN~\cite{zhang2016learning}
and CFSS~\cite{zhu2015face}. Since these methods operate in the general
framework of pose regression, performing iterative refinement of predicted
landmark locations, they are sensitive to initial bounding box location. We
tested both models using the standardized detection bounding boxes provided by
the 300-W benchmark \cite{300w} rather than tight cropping images to the
ground-truth landmark locations.  We used the pre-trained TCDCN model available
online while for CFSS we retrained the model using the standard detector
bounding boxes.  In both cases, average error was significantly worse than
previously reported results, highlighting the sensitivity of these methods to
initialization.

\paragraph{Dependence of Localization on Detection} 
A key benefit of the HPM (and DPM~\cite{zhu2012face}) approach is that the same
model serves to both detect and localize the landmarks. In contrast, pose
regression methods such as RCPR, TCDN or CFSS require that the face already be
detected. This distinction becomes particularly important for occluded faces
since detection is significantly less accurate (see Detection experiments
below).

To highlight the dependence of landmark localization on accurate detection, we
benchmarked average localization error for varying degrees of overlap between
the hypothesized detection and ground-truth bounding box on the COFW test set.
As shown in Fig. \ref{fig:overlap}, decreasing the overlap ratio has no affect
HPM / DPM performance since there are never false positives in the vicinity of
the face that score higher than one with high overlap ratio.  In contrast, RCPR
performs significantly worse when initialized from bounding boxes that do not
have high overlap with the face. Since the area over which RCPR searches is
learned from training data, we also retrained a version of RCPR for each
degrees of overlap.  This yielded improved performance but still shows a
significant fall off in performance compared to the HPM. As noted above,
we encountered similar behavior when evaluating other methods such as
TCDNN and CFSS on realistic detector-generated bounding boxes.

\paragraph{Dependence of Localization on Training data}
One advantage of the HPM model is robustness to the choice of training data
set.  Table \ref{table:generalization}  highlights a comparison of HPM and RCPR in which the training set
is varied.  HPM performs well on LFPW and COFW regardless of training set
specifics.  In contrast, RCPR shows better performance on COFW when the
training data is also taken from COFW. Training data augmentation is also
important to achieve good performance with RCPR, while HPM works well even when
trained on the relatively smaller LFPW training set.

\subsection{Occlusion Prediction}
\label{sec:occlusionprediction}
To evaluate the ability of the model to correctly determine which landmarks are
occluded, we evaluate the accuracy of occlusion as a binary prediction task.
For a given test set, we compute precision and recall of occlusion predictions
relative to the ground-truth occlusion labels of the landmarks. 

For HPM, we trace out a precision-recall curve for occlusion prediction by adjusting the
model parameters to induce different predicted occlusions.  As described in
Section \ref{sec:model}, the bias parameter $b_{ij}(s_i, s_j, o_i, o_j)$ favors
particular co-occurrences of part types. By increasing (decreasing) the bias
for occluded configurations we can encourage (discourage) the model to use
those configurations on test.  Let $b_{ij}(s_i, s_j, o_i, o_j)$ be a learned
bias parameter between an occluded leaf and its parent. To make the model favor
occluded parts, we modify this parameter to $b_{ij}(s_i, s_j, o_i,
o_j)+abs(b_{ij}(s_i, s_j, o_i, o_j))\times \alpha$. 

Fig. \ref{fig:margin_scaling}(a) depicts occlusion precision-recall curves
generated by running the HPM model for different bias $\alpha$ offsets. The 
crosses mark the precision-recall for the default operating point when
$\alpha=0$.  We compare performance of the HPM model with different values of
the margin scaling hyper-parameter $m$ as well as RCPR and a baseline
independent occlusion model.  Fig. \ref{fig:margin_scaling} (b) and (c) show
the corresponding average errors and success rates for these models
parameterized by the recall of occlusion. For large values of $\alpha$, the
model predicts more occlusions, resulting in improved recall at the expense of
precision (a) and ultimately lower localization accuracy (b,c).

\paragraph{Margin scaling}
As described in section \ref{sec:param_learning}, we can change the learning
parameter $m$ to produce models with different recall of occlusions at the
trained operating point ($\alpha=0$).  When $m=0$ all the negative examples
including fully or partially occluded configurations are penalized equally.
Therefore, model learns small biases for occluded configurations, reducing the
total loss over occluded negative examples and decreasing default recall of
occlusion. When driven to predict more occlusion by increasing $\alpha$ the
model localization performance degrades rapidly. Training the model with larger
values of $m$ yields a model which naturally predicts occlusion more
frequently and degrades more gracefully for larger values of $\alpha$.  We
found that choosing a value of $m=0.5$ provided a good compromise, improving
both recall and localization accuracy.

\paragraph{Independent occlusion baseline}
We compared the results of HPM with a model that had the same architecture but
in which there are no occlusion mixtures at the part level and each landmark is
allowed to be independently set to visible or occluded depending on learned
biases.  We refer to this as ``independent occlusion'' since the model does not
capture any correlations between the occlusion of different landmarks.  We
found that this independent occlusion model has many of the same benefits as
the HPM model in terms of landmark localization accuracy (Fig.
\ref{fig:margin_scaling}).  However, occlusion prediction accuracy is
significantly worse in the independent model with precisions typically 5\%
lower than HPM$(m=0.9)$ over a range of recall values.

\begin{figure}[t]
\begin{center}
\includegraphics[width=1\columnwidth]{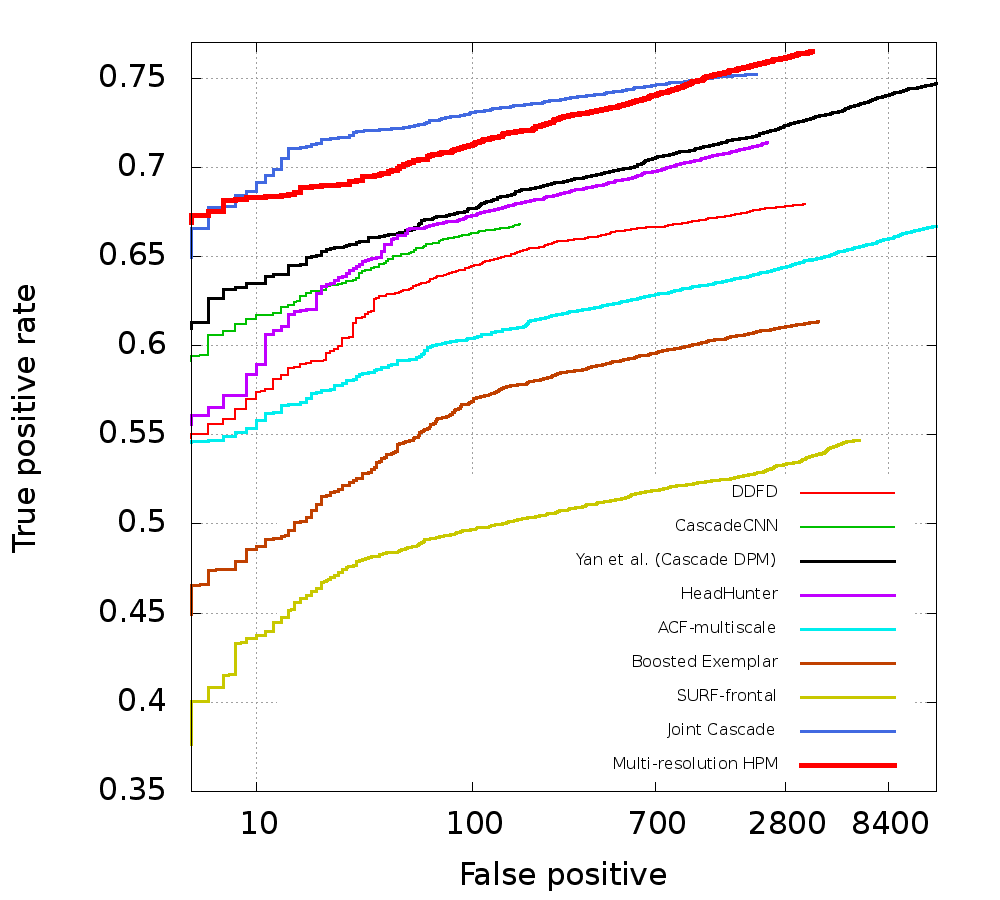}\\
\end{center}
\caption{Face detection performance of HPM and state-of-the-art
methods ~\cite{link:fddb_results} on the continuous-ROC 
FDDB benchmark~\cite{fddbTech}.
}
\label{fig:fddb_curves}
\end{figure}

\begin{figure*}[t]
\begin{center}
\begin{tabular}{ccc}
\includegraphics[width=0.65\columnwidth]{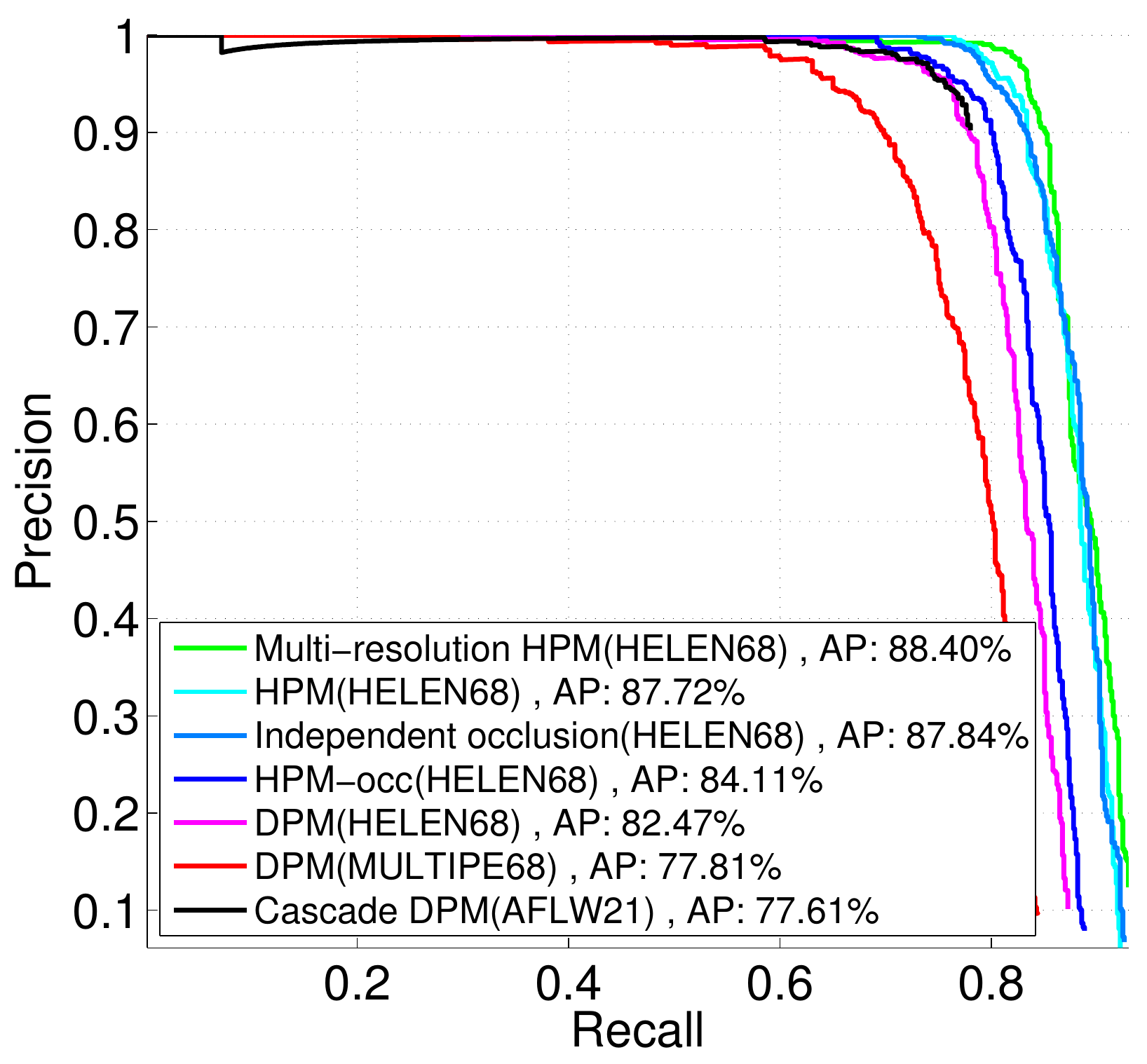}&
\includegraphics[width=0.65\columnwidth]{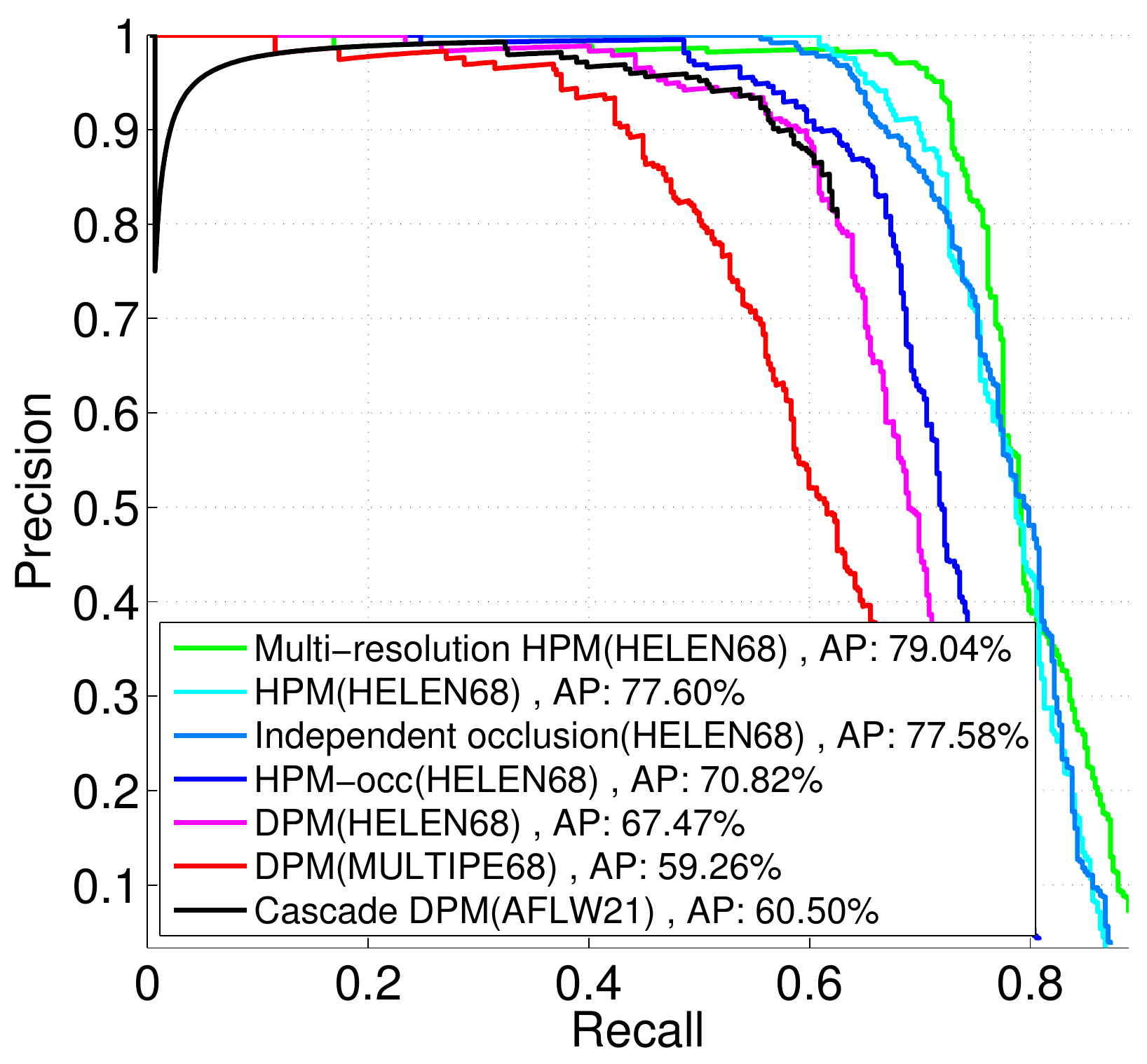}&
\includegraphics[width=0.65\columnwidth]{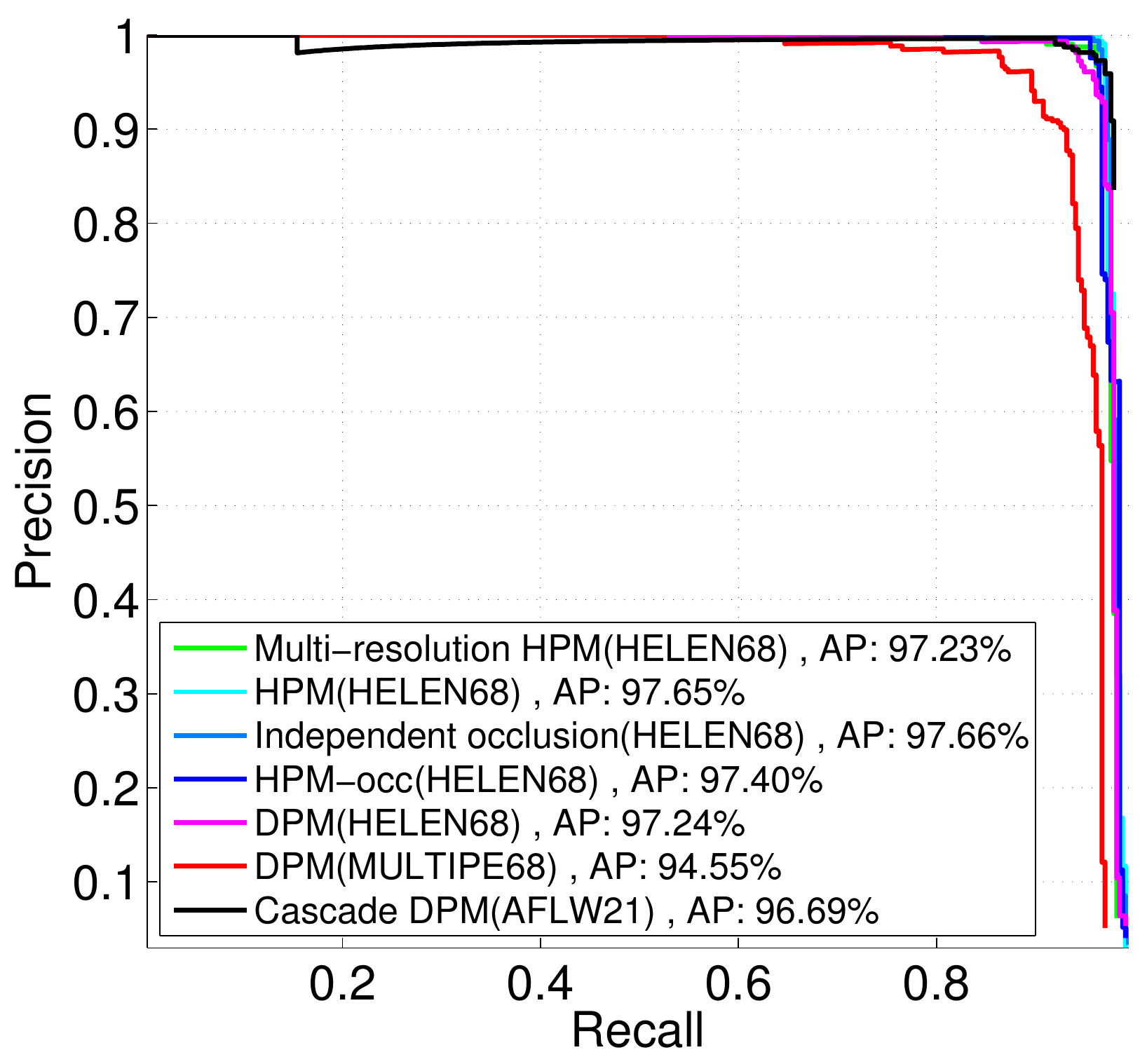}\\
(a) UCI-OFD & (b) UCI-OFD occluded & (c) UCI-OFD visible \\
\end{tabular}
\end{center}
\caption{
Precision-Recall curves of face detection on our UCI-OFD dataset (a) for all of
the faces, (b) occluded subset and (c) visible subset. On the visible subset
our model, DPM retrained on HELEN68 and Cascade DPM~\cite{yanfastest} have
almost similar performances, but our model significantly outperforms these
methods on the occluded subset and it has a better overall performance. Cascade
DPM uses many accelerate techniques, which may reject some of the faces. Its
maximum recall for the visible faces is near 100\%, while its maximum recall
for the occluded faces is only 60\%. The initial drop in the Precision-Recall
of this method for the occluded subset is because its returned bounding boxes
for some of the high scored occluded faces are not accurate and do not have the
minimum 0.5 overlap with the ground-truth bounding boxes.}
\label{fig:uci_ofd_curves}
\end{figure*}

\subsection{Detection}
\label{sec:facedetection}

Pose regression requires good initialization provided by a face detector 
to accurately locate landmarks.  In contrast, part-based models have the 
elegant advantage of performing detection and localization simultaneously.
In this section, we compare the detection performance of our approach and other
top methods on two datasets: FDDB~\cite{fddbTech} and our own Occluded Face 
Detection (UCI-OFD) dataset.

\paragraph{Multi-resolution HPM}
Since many face detection datasets such as FDDB contain many low-resolution faces,
we trained a multi-resolution variant of our model~\cite{park2010multiresolution}.  
This model has a high and a low-resolution model for each viewpoint. The high
resolution model has the same structure as our trained model for landmark
localization except that parts are represented as 3x3 HoG cells rather than
5x5. The low-resolution model has 7 parts (right eye, left eye, nose, mouth,
chin, left jaw and right jaw) each of which is represented by 7x7 HoG cells
with the spatial bin size of 4. Each part has one shape mixture and 2 occlusion
mixtures (visible or occluded).  The heights (eyebrow to chin) of the large
model and small model are about 100 and 60 pixels respectively. To detect even
smaller images, we upsample input images by a factor of 2 to allow for detection
of faces as small as 30 pixels.  We trained this model using the same 1758
positive examples from HELEN68 and generated 8 virtual positive examples per
example. For negative images we used 6000 images from the PASCAL VOC 2010
train-val set which do not contain people.

\paragraph{Detection on FDDB}
We evaluated our multi-resolution model on the widely used FDDB dataset.  This
dataset contains 5171 faces in a set of 2845 images.  Faces are annotated by
ellipses in this dataset and are as small as 20 pixels in height. To match
that, we map our predicted landmark locations to ellipses using a linear
regression model. FDDB has 10 folds and the ROC curves are the average over the
results of these folds. To compute ellipses for each fold, we learned the linear
regression coefficients using examples from the other 9 folds. 

We used the standard evaluation protocol for this dataset and compared our
method with the top published results available on the FDDB
website~\cite{link:fddb_results}. The continuous ROC curve for our
method and leading methods are shown in Fig. \ref{fig:fddb_curves} plotted
on a semi-log scale. Our result is highly competitive with the top results.
The model has better performance on the continuous ROC evaluation relative
to other methods because it can predict location of parts and compute
accurate bounding ellipses around the faces.  

\paragraph{UCI Occluded Face Detection Dataset (UCI-OFD)}
In order to better measure the ability of our model to handle detection of
occluded faces, we assembled a preliminary dataset for occluded face detection.
This dataset and benchmarking code are publicly available
\footnote{\url{https://github.com/golnazghiasi/hpm-detection-code/tree/master/UCI\_OFD}}.
It consists of 61 images from Flickr containing 766 labeled faces.  Of the
faces in these images, 430 include some amount of occlusion. Most of the faces
are near frontal and vertical.  Height (eyebrow to chin) of the smallest face
is about 40 pixels.

\begin{figure*}[t]
    \includegraphics[height=0.24\textwidth]{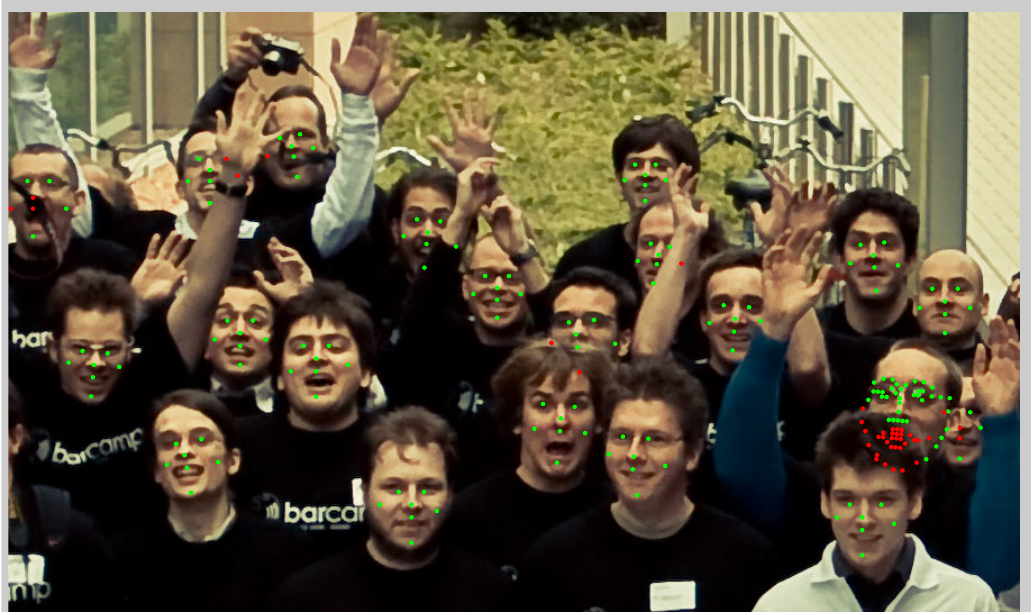}\hspace{-5pt}
    \includegraphics[height=0.24\textwidth]{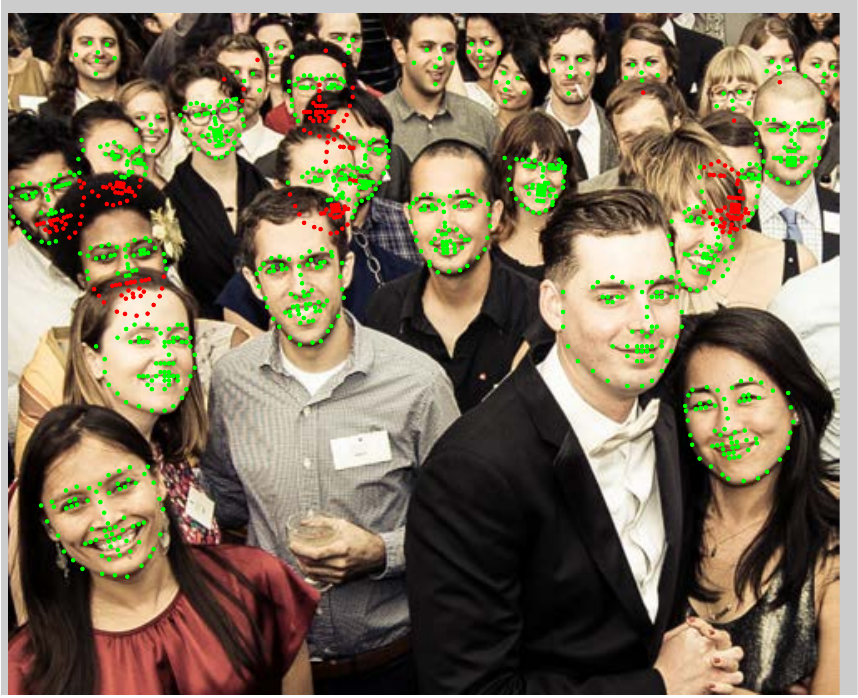}\hspace{-5pt}
    \includegraphics[height=0.24\textwidth]{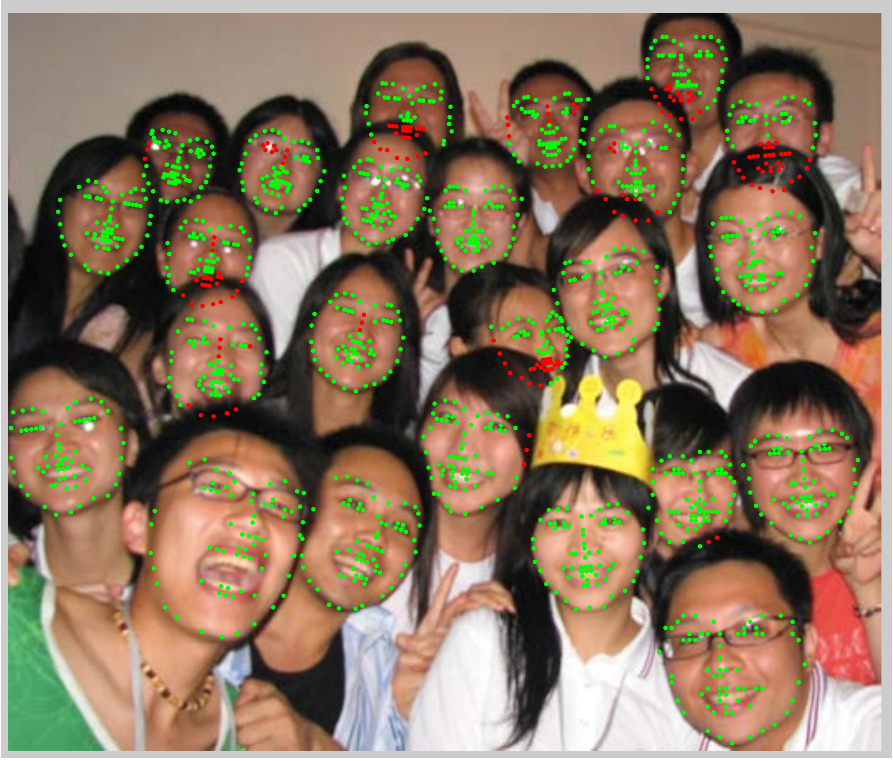}\\
    \includegraphics[height=0.208\textwidth]{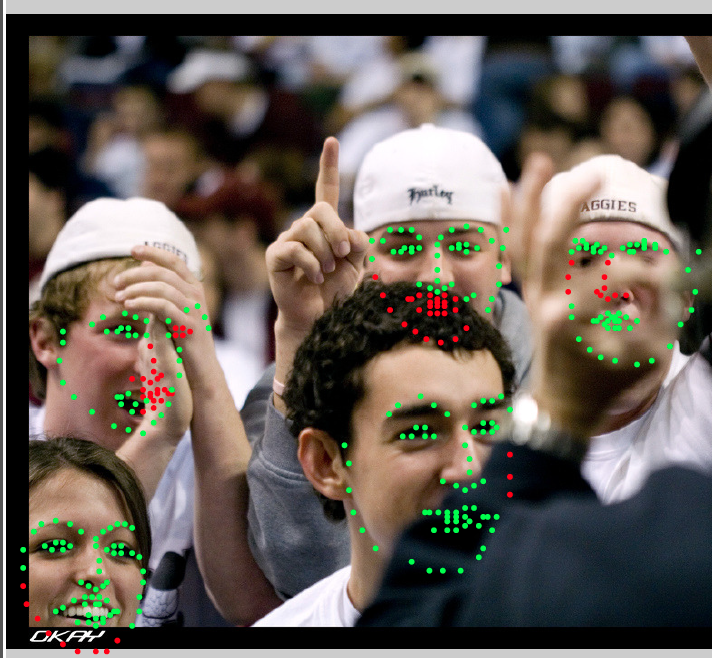}\hspace{-3pt}
    \includegraphics[height=0.208\textwidth]{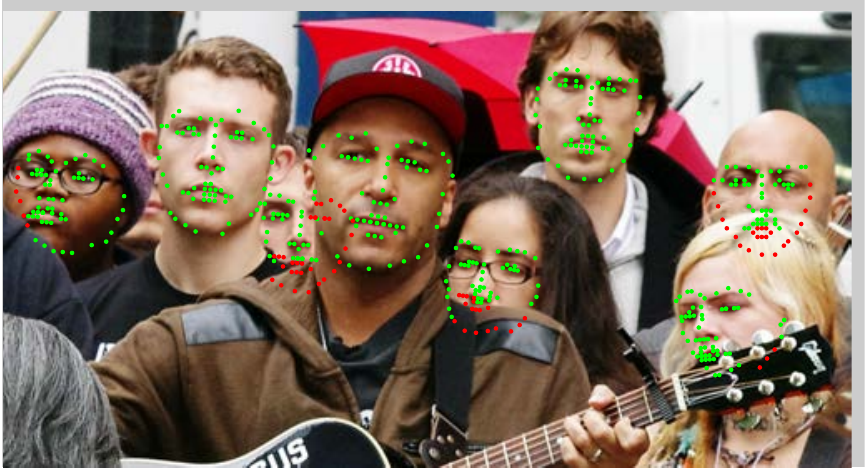}\hspace{-4pt}
    \includegraphics[height=0.208\textwidth]{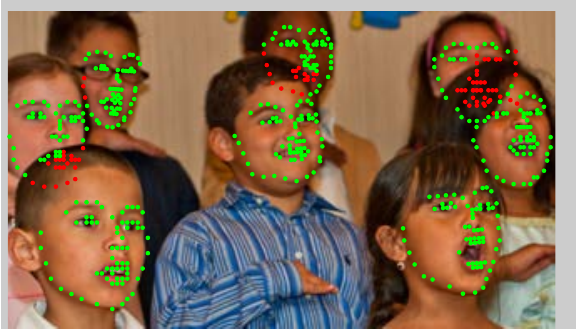}\\
	\includegraphics[height=0.215\textwidth]{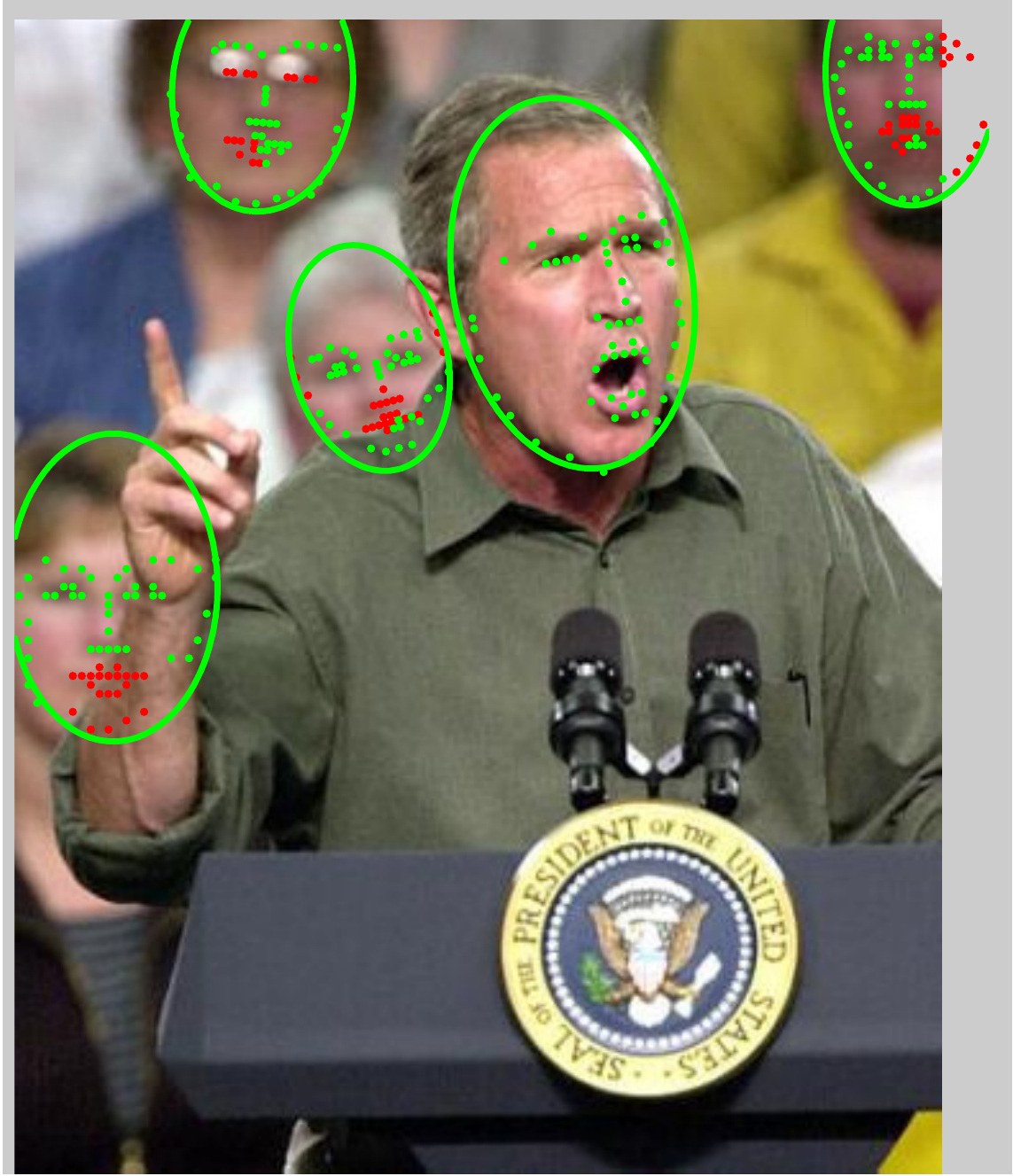} \hspace{-7pt}
	\includegraphics[height=0.215\textwidth]{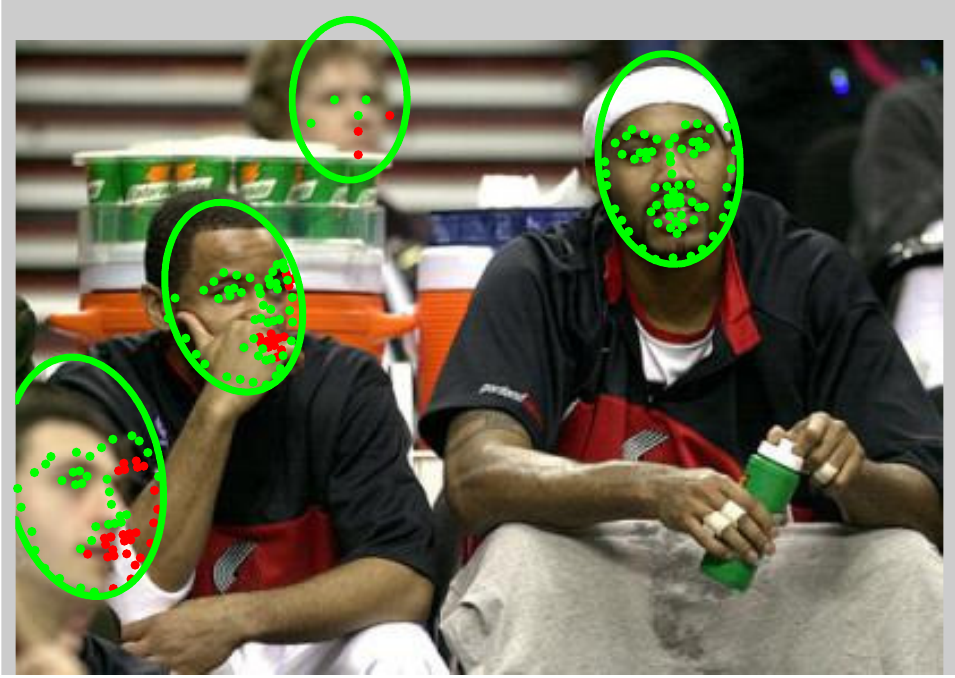} \hspace{-7pt}
	\includegraphics[height=0.215\textwidth]{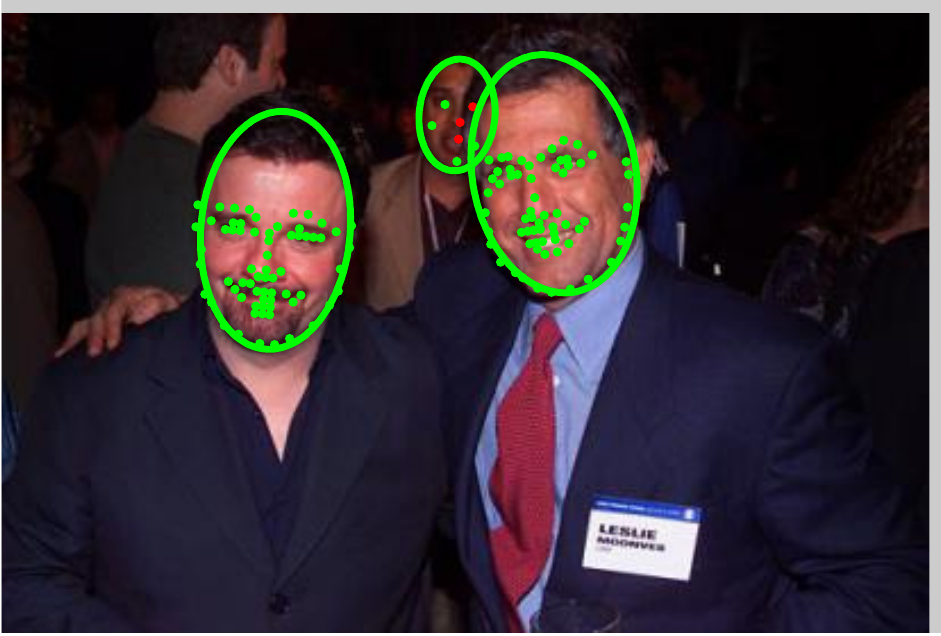} \hspace{-7pt}
	\includegraphics[height=0.215\textwidth]{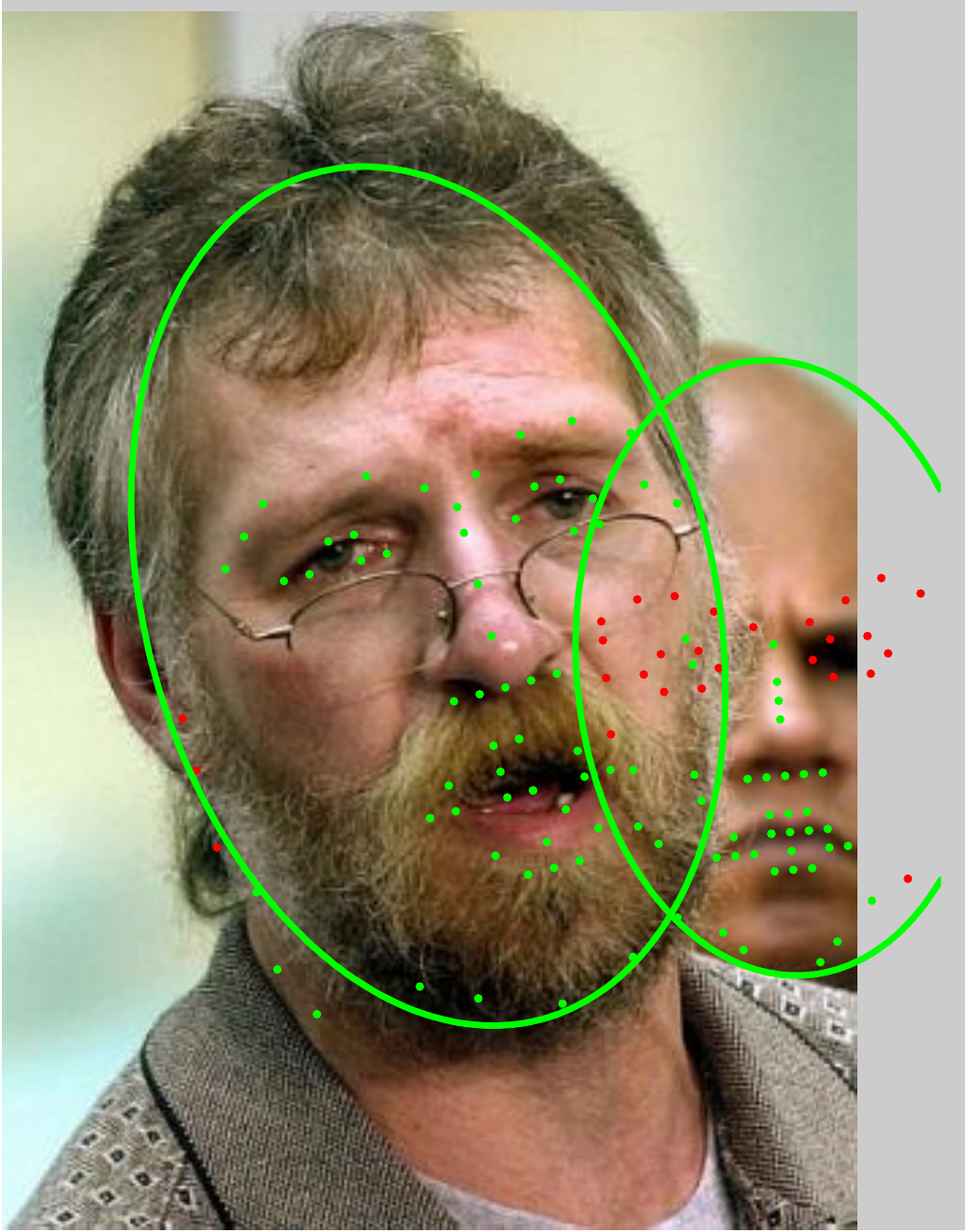} \\
	\includegraphics[height=0.221\textwidth]{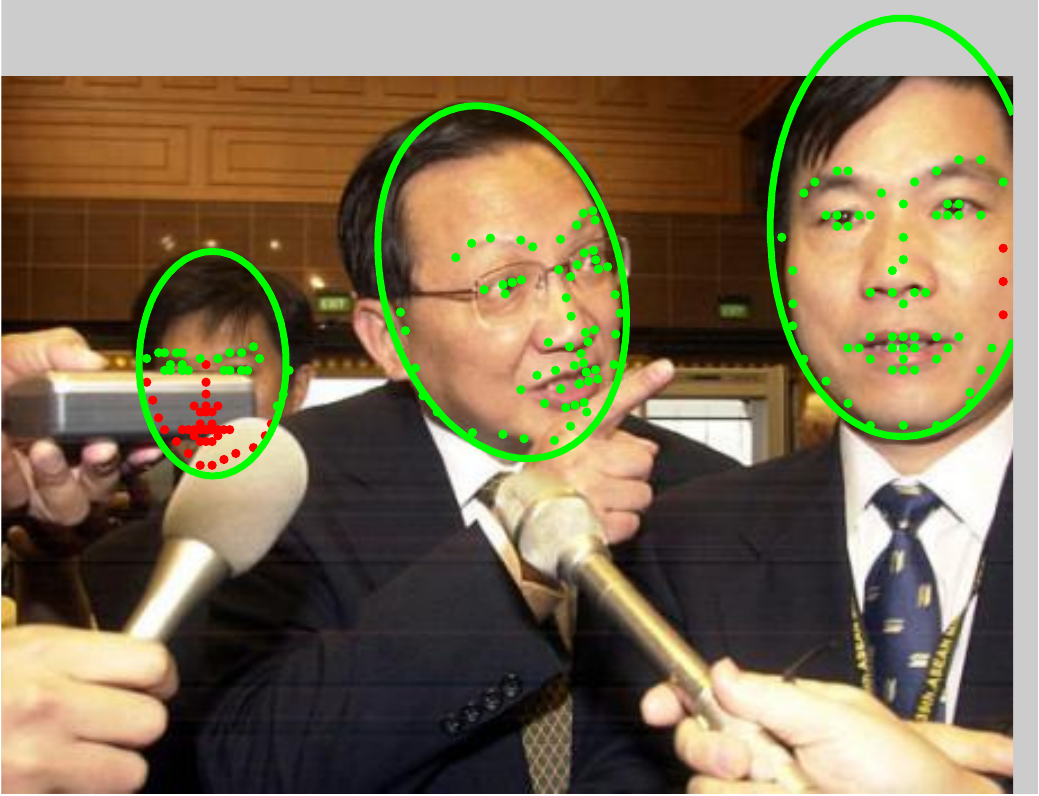} \hspace{-5pt}
	\includegraphics[height=0.221\textwidth]{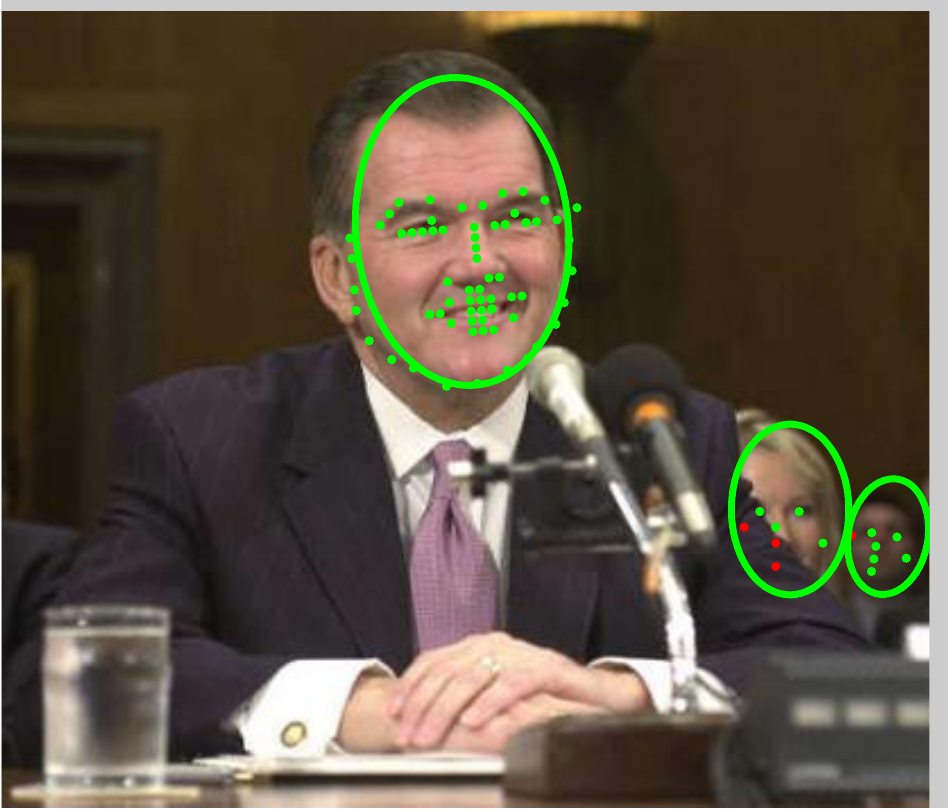}
	\includegraphics[height=0.221\textwidth]{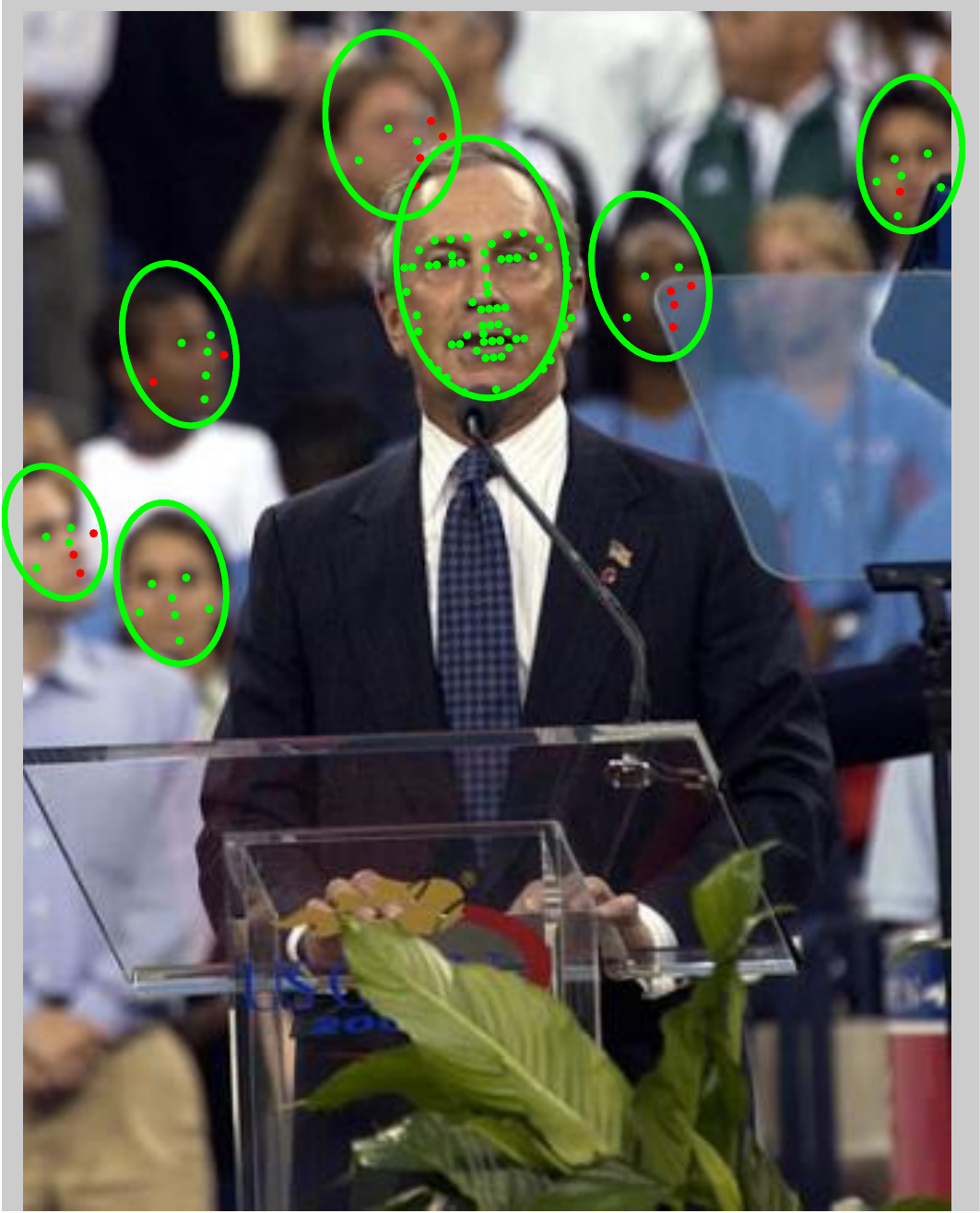}
	\includegraphics[height=0.221\textwidth]{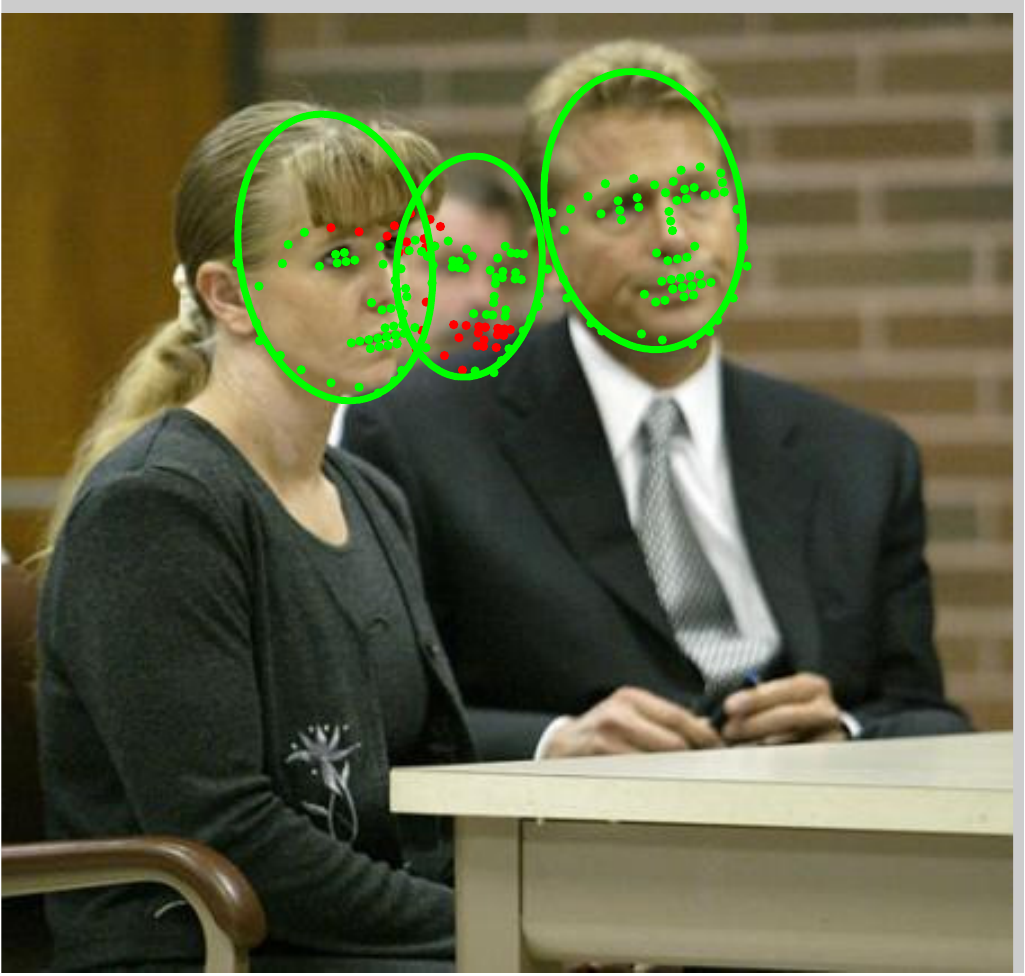}\\ 	
  \includegraphics[height=0.222\textwidth]{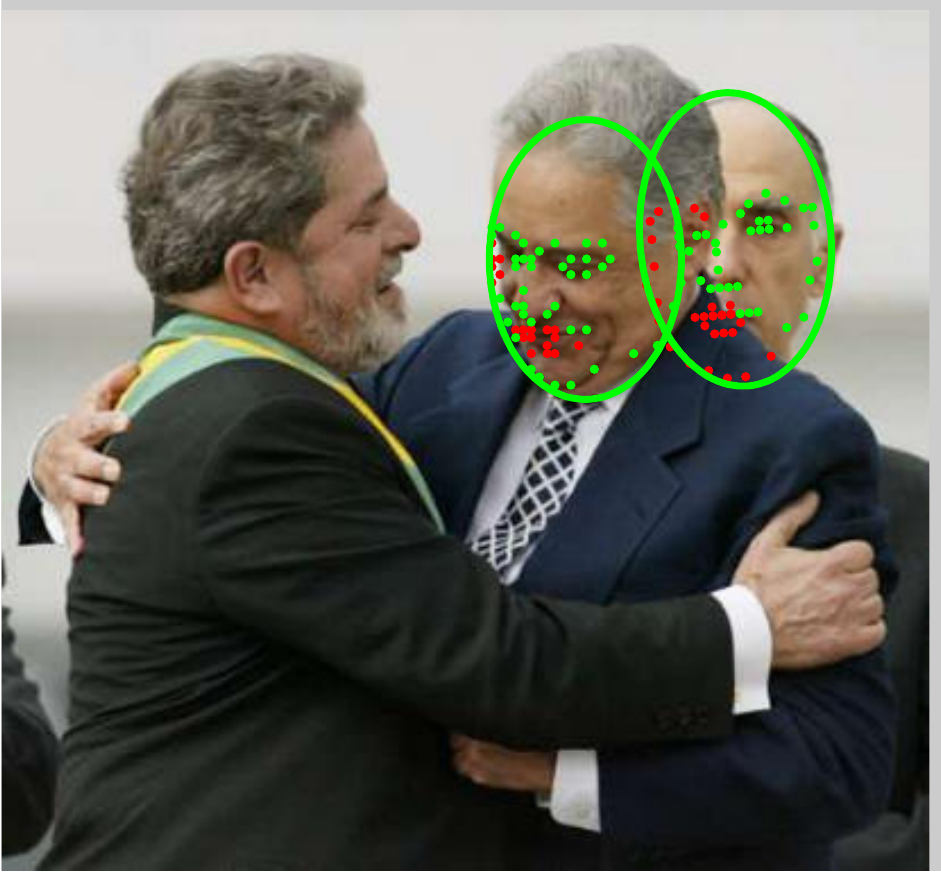} \hspace{-5pt}
	\includegraphics[height=0.222\textwidth]{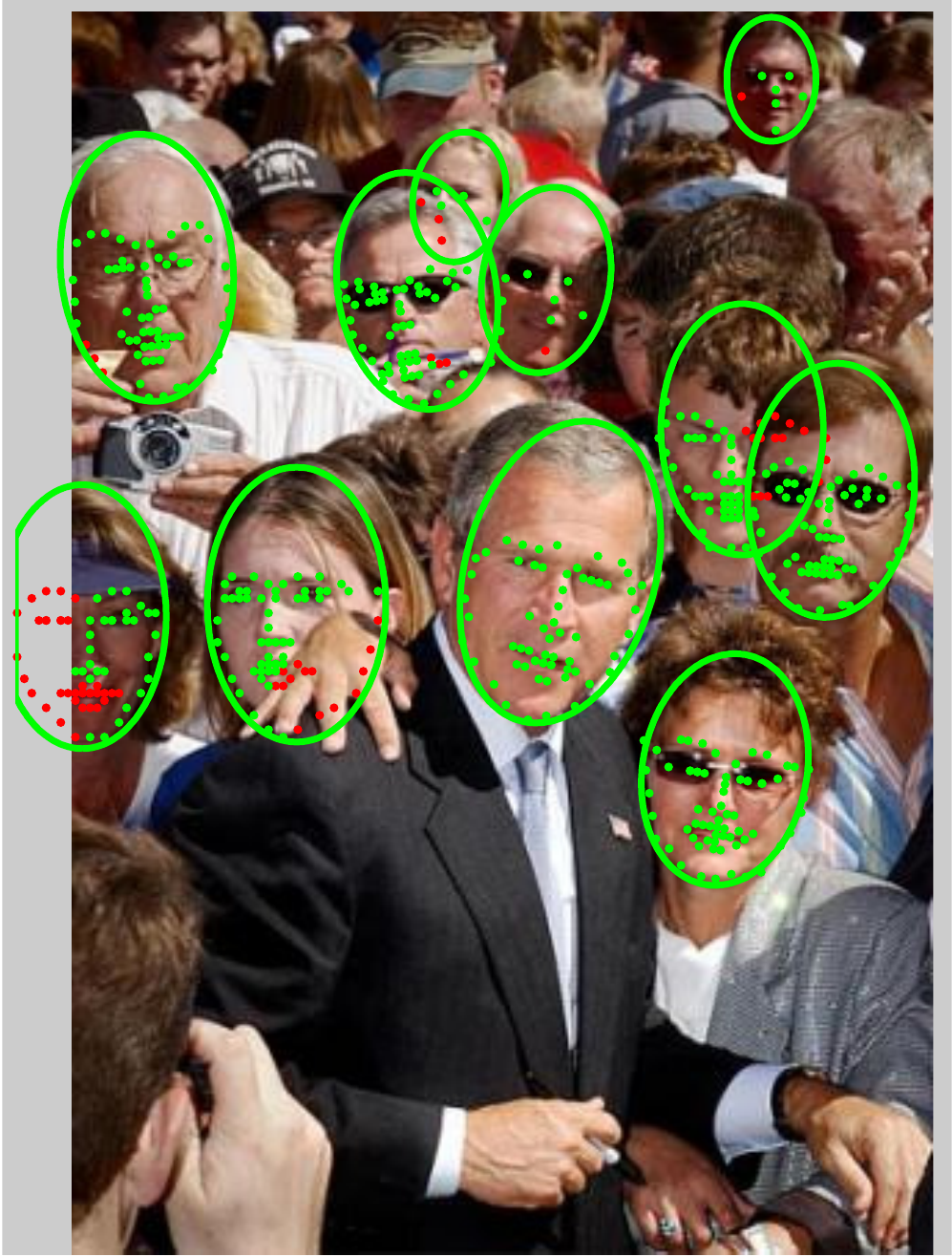} \hspace{-5pt}
	\includegraphics[height=0.222\textwidth]{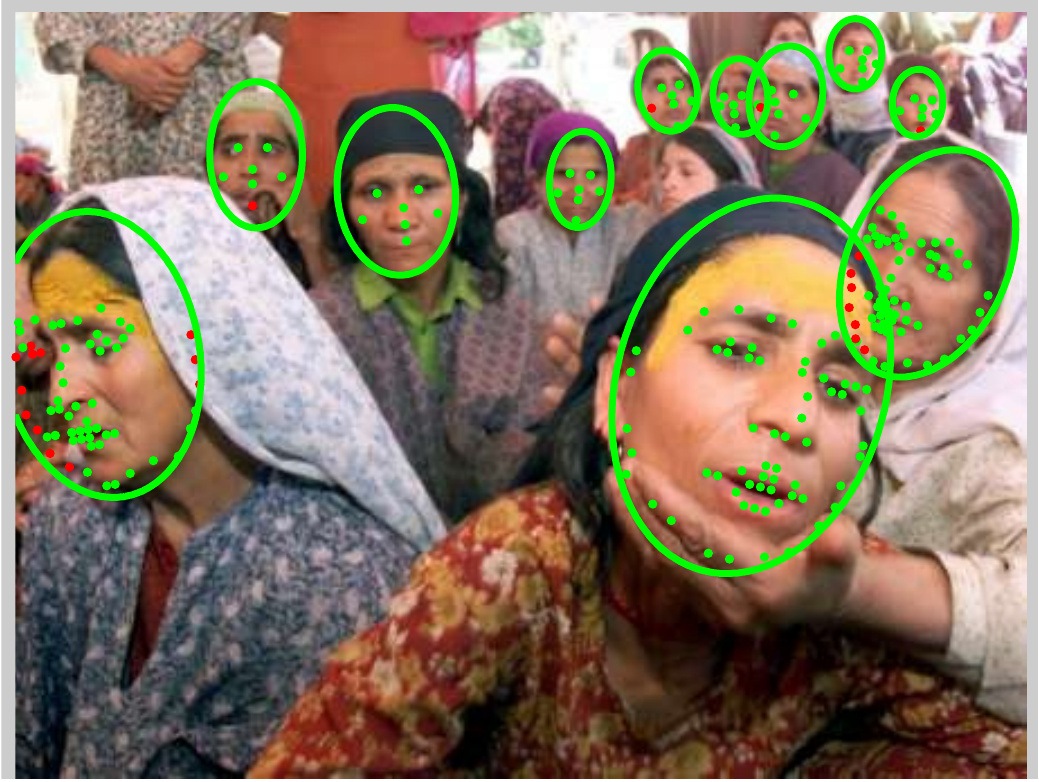}
	\includegraphics[height=0.222\textwidth]{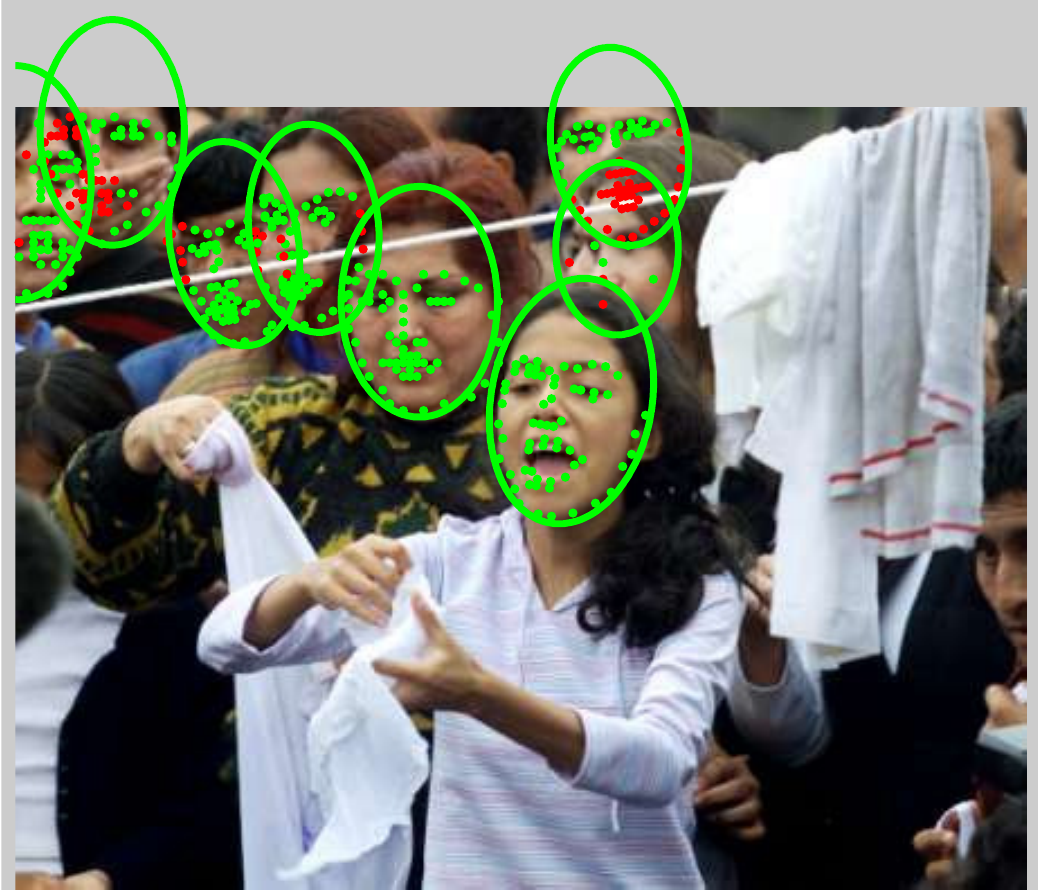}\\
\caption{Examples of detection and localization for images from our UCI-OFD
dataset (rows 1-2) and images containing occlusion from FDDB dataset (rows
3-4). Detections indicated with only 7 landmarks correspond to responses from
the low-resolution model component. Ellipses are predicted on FDDB images by
linear regression from landmark locations to ellipse parameters.}
\label{fig:results2}
\end{figure*}

Precision/Recall curves of face detection of multi-resolution HPM, HPM,
HPM-occ, DPM and Cascade DPM~\cite{yanfastest} are shown in Fig.
\ref{fig:uci_ofd_curves}(a).  We further break down performance, plotting
Precision/Recall curves for the subset of faces with some amount of occlusion
in (b) and fully visible in (c).  Precision and recall for occluded subset of
faces are calculated as below: 
\[
\text{Precision}_o = \frac{tp_o}{tp_o + fp}, \text{Recall}_{o} = \frac{tp_o}{tp_o + fn_o}
\]
where $tp_o$ and $fn_o$ show number of correct detection and miss detection of 
occluded faces, respectively.  Our method significantly outperforms other methods
on the occluded subset and the performance of all of the methods are almost equal
on the visible subset. Fig.~\ref{fig:results2} shows example detection results 
produced by the model on cluttered scenes containing many overlapping faces.

\section{Discussion and Conclusion}

Our experimental results demonstrate that adding coherent occlusion and
hierarchical structure allows for substantial gains in performance for landmark
localization and detection in part models.  In images with relatively little
occlusion, the HPM gives similar detection and localization performance to
other part-based approaches, e.g. DPM, but is significantly more robust to
occlusion. Our results also suggest that when it is useful to determine exactly
which parts are occluded (e.g., for later use in face identification),
independent occlusion makes weaker predictions than our part occlusion mixtures
which enforce coherence between neighboring landmarks.  While not specifically
trained for landmark estimation, the final HPM is competitive with pose
regression techniques in terms of landmark localization accuracy on unoccluded
faces (IBUG) and outperforms many such methods on occluded faces (Occluded
HELEN, COFW).

In comparing pose regression and part-based models, there seem to be several
interesting trade-offs.  In our experiments, we see a general trend in which
error distribution curves for pose regression and part-based models cross,
suggesting that pose regression yields very accurate localization for a subset
of images relative to the HPM but fails for some other proportion even at very
large error thresholds. Unlike pose regression, the part model performs
detection, eliminating the need for detection as a pre-process and improving
robustness.  In particular, we are able to detect many heavily occluded faces
which would not be detected by a standard cascade detector and hence inaccessible
to pose regression. We find that the HPM tends to generalize well across
datasets suggesting it avoids some overfitting problems present in pose
regression.  

This flexibility currently comes with a computational cost. The run-time of our
model implementation built on dynamic programming lags significantly behind
those of regression-based, feed-forward approaches. Our implementation takes
${\sim}10$s to run on a typical COFW image, roughly 100x slower than RCPR or
DCNN based approaches. However, the HPM is amenable to implementation on a GPU
which may address most of this runtime gap.

Finally, we note several avenues for future work.  Performance depends on the
graphical independence structure of the model which should ideally be learned
from data.  While our model implicitly represents the pattern of part
occlusions, it does not integrate local image evidence for the occluder itself.
A natural extension would be to add local filters that detect the presence of
an occluding contour between the occluded and non-occluded landmarks.  Such
filters could be shared across parts to avoid increasing too much the overall
computation cost while moving closer to our goal of explaining away missing
object parts using positive evidence of coherent occlusion.

\vspace{0.5cm}
\noindent{\bf Acknowledgements:} This work was supported in part by NSF grants
IIS-1253538 and DBI-1262547.

\bibliographystyle{elsarticle-num}
\section*{References}
\bibliography{main}

\end{document}